\documentclass[11pt,a4paper]{article}

\usepackage{graphicx}
\usepackage[a4paper, left=1in, right=1in, top=1in, bottom=1in, includehead, includefoot, foot=0pt, head=0pt]{geometry} % set margins
\parskip=\medskipamount % set spacing between paragraphs
\usepackage{amssymb} % math symbols
\usepackage{amsmath} % \eqref{label} AND split equations
\usepackage{amsbsy} % \boldsymbol{symbol}
\usepackage{booktabs} % nicer tables
\usepackage{multirow} %\multirow{number of rows}{width}{text}
\usepackage[scientific-notation=false, group-separator={,}]{siunitx} % format notation of numbers via \num{number}
\usepackage[small,bf,hang]{caption}	% set caption labels
\usepackage[round, colon, authoryear]{natbib}
\usepackage{hyperref}
\hypersetup{
	colorlinks=true,
	citecolor=[RGB]{17,110,138},
	linkcolor=[RGB]{17,110,138},
	urlcolor=[RGB]{17,110,138},
}
\usepackage{titling} % Modify title page settings
\usepackage{pifont}% http://ctan.org/pkg/pifont
\newcommand{\cmark}{\ding{51}}%
\newcommand{\xmark}{\ding{53}}%
\usepackage{authblk}
\usepackage{float}
\usepackage{subcaption}

\usepackage{eurosym}

\usepackage{bbm}

\usepackage{lipsum}

\usepackage{fancyhdr}
\usepackage{changepage}

\newcommand{\E}{\mathbb{E}}

\newcommand{\x}{\boldsymbol{x}}
\newcommand{\X}{\boldsymbol{X}}
\newcommand{\z}{\boldsymbol{z}}

\newcommand{\D}{\mathcal{D}}

\usepackage{mathrsfs}

\renewcommand{\l}{\left(}
\renewcommand{\r}{\right)}

\usepackage{outlines}

\usepackage{soul}
\newcommand{\ulcolor}[2][Red]{\setulcolor{#1}\ul{#2}}

\makeatletter
\newcommand*\bigcdot{\mathpalette\bigcdot@{.5}}
\newcommand*\bigcdot@[2]{\mathbin{\vcenter{\hbox{\scalebox{#2}{$\m@th#1\bullet$}}}}}
\makeatother

\usepackage[noend]{algorithm2e} % Algorithms
\SetAlCapFnt{\small} % Change the caption label font for the algorithms
\SetAlCapNameFnt{\small} % Change the caption text font for the algorithms
\usepackage{setspace} %more whitespace in between lines

\usepackage[dvipsnames,table]{xcolor} % Extra colors
\usepackage{tikz} % Tikz figures
\usetikzlibrary{matrix, fit, decorations.pathreplacing} % different Tikz libraries
\usepackage{listofitems} %the \readlist command in the Tikz
\tikzstyle{mynode}=[thick,draw=blue,fill=blue!20,circle,minimum size=22]

\usetikzlibrary{arrows.meta} % for arrow size
\usepackage[outline]{contour} % glow around text
\contourlength{1.4pt}

\tikzset{>=latex} % for LaTeX arrow head
\colorlet{myred}{red!80!black}
\colorlet{myblue}{blue!80!black}
\colorlet{mygreen}{green!60!black}
\colorlet{myorange}{orange!70!red!60!black}
\colorlet{mydarkred}{red!30!black}
\colorlet{mydarkblue}{blue!40!black}
\colorlet{mydarkgreen}{green!30!black}
\tikzstyle{node}=[thick,circle,draw=myblue,minimum size=22,inner sep=0.5,outer sep=0.6]
\tikzstyle{node in}=[node,green!20!black,draw=mygreen!30!black,fill=mygreen!25]
\tikzstyle{node hidden}=[node,blue!20!black,draw=myblue!30!black,fill=myblue!20]
\tikzstyle{node convol}=[node,orange!20!black,draw=myorange!30!black,fill=myorange!20]
\tikzstyle{node out}=[node,red!20!black,draw=myred!30!black,fill=myred!20]
\tikzstyle{connect}=[thick,mydarkblue] %,line cap=round
\tikzstyle{connect arrow}=[-{Latex[length=4,width=3.5]},thick,mydarkblue,shorten <=0.5,shorten >=1]
\tikzset{ % node styles, numbered for easy mapping with \nstyle
  node 1/.style={node in},
  node 2/.style={node hidden},
  node 3/.style={node out},
}
\usepackage{adjustbox}

\usepackage{nicematrix}

\usepackage{tablefootnote}

\usepackage{diagbox}
\newcolumntype{P}[1]{>{\centering\arraybackslash} m{#1}} % centered horizontally and vertically

\usepackage{tabularx}
\makeatletter
\def\hlinewd#1{%
\noalign{\ifnum0=`}\fi\hrule \@height #1 %
\futurelet\reserved@a\@xhline}
\makeatother
\newcolumntype{?}[1]{!{\vrule width #1}}

\usepackage{array}

\usepackage{arydshln}

\usepackage{makecell}

\usepackage{colortbl}

\definecolor{myRed}{RGB}{228,15,15}
\definecolor{myGreen}{RGB}{40,150,40}
\definecolor{myOrange}{RGB}{255, 150, 0}
\definecolor{myPink}{RGB}{255, 20, 147}
\definecolor{myPurple}{RGB}{204,20, 167}
\definecolor{kul-secblue}	{RGB}{220,231,240}	
\xdefinecolor{UUxblue}{cmyk}{0.806,0.139,0,0.576}
\definecolor{myRed}{RGB}{255, 40, 40}

\definecolor{kul-blue}		{RGB}{29,141,176}		%KUL blue (primair blauw)
\definecolor{kul-blue2}{RGB}{17,110,138} 			% Old KU Leuven blue
\definecolor{kul-secblue}	{RGB}{220,231,240}		%KUL light gray-blue (secundair blauw)
\definecolor{kul-lightblue} {RGB}{82,189,236}
\definecolor{kul-dark}		{RGB}{47,77,93}			%KUL dark blue
\definecolor{gray}			{gray}{.5}				%gray
\definecolor{lgray}			{gray}{.9}				%light gray for theorem background

\definecolor{tabwhite}{HTML}{FFFFFF}
\definecolor{tabgray}{HTML}{FAFAFF}

\usepackage{calc}
\graphicspath{{graphics/}}

\usepackage[many]{tcolorbox}
\newtcolorbox{rightbrace}[1]{%
    enhanced jigsaw, 
    frame hidden, % hide the default frame
    overlay={%
        \draw [
            fill=none,
            gray,
        ];
        \node at ([shift={(2mm,0mm)}]frame.east) [anchor=center, gray, rotate=270] {#1}; 
    },
    parbox=false,
}

\newcommand{\absdiv}[1]{%
  \par\addvspace{.5\baselineskip}% adjust to suit
  \noindent\textbf{#1}\quad\ignorespaces
}

\usepackage[defaultcolor=black, authormarkup=none]{changes}
\definechangesauthor[name=roundtwo, color=black]{roundtwo}
\newcommand\addedtwo{\added[id=roundtwo]}

\begin{document}

% Make the title page
\title{\huge Neural networks for insurance pricing with frequency and severity data: \\ \vspace{6pt} \Large a benchmark study from data preprocessing to technical tariff}
\author[a,c]{Freek Holvoet}
\author[a,b,c]{Katrien Antonio}
\author[a]{Roel Henckaerts}
\affil[a]{Faculty of Economics and Business, KU Leuven, Belgium.}
\affil[b]{Faculty of Economics and Business, University of Amsterdam, The Netherlands.}
\affil[c]{LRisk, Leuven Research Center on Insurance and Financial Risk Analysis, KU Leuven, Belgium.}
%\author{Anonymous}
\date{\today}

%\predate{}
%\postdate{}

{\centering \maketitle}
\thispagestyle{empty}

\begin{abstract}
\noindent Insurers usually turn to generalized linear models for modeling claim frequency and severity data. Due to their success in other fields, machine learning techniques are gaining popularity within the actuarial toolbox. Our paper contributes to the literature on frequency-severity insurance pricing with machine learning via deep learning structures. We present a benchmark study on four insurance data sets with frequency and severity targets in the presence of multiple types of input features. We compare in detail the performance of: a generalized linear model on binned input data, a gradient-boosted tree model, a feed-forward neural network (FFNN), and the combined actuarial neural network (CANN). The CANNs combine a baseline prediction established with a GLM and GBM, respectively, with a neural network correction. We explain the data preprocessing steps with specific focus on the multiple types of input features typically present in tabular insurance data sets, such as postal codes, numeric and categorical covariates. Autoencoders are used to embed the categorical variables into the neural network, and we explore their potential advantages in a frequency-severity setting. \added{Model performance is evaluated not only on out-of-sample deviance but also using statistical and calibration performance criteria and managerial tools to get more nuanced insights.} Finally, we construct global surrogate models for the neural nets' frequency and severity models. These surrogates enable the translation of the essential insights captured by the FFNNs or CANNs to GLMs. As such, a technical tariff table results that can easily be deployed in practice.
\absdiv{Practical applications summary:} This paper explores how insights captured with deep learning models can enhance the insurance pricing practice. Hereto, we discuss the required data preprocessing and calibration steps, and we present a workflow to construct GLMs for frequency and severity data by leveraging the insights obtained with a carefully designed neural network.
\\[.5\baselineskip]
\\[.5\baselineskip]
\end{abstract}

\setlength{\extrarowheight}{3pt} % a bit of extra whitespaces between lines in tables

\section{Introduction}\label{intro}

One of the central problems in actuarial science is the technical pricing of insurance contracts. Premiums are determined at the time of underwriting, while the actual cost of the contract is only known when claims are processed. The technical premium is defined as the expected loss on a contract. In property and casualty insurance (P\&C), expected losses are often estimated by independently modeling the frequency and severity of claims in function of policy and policyholder information. Hence, the use of historical data sets, with policyholder characteristics and the observed claim frequency and severity, is key in the design of predictive models. These historical data sets are of tabular structure, containing numerical, categorical and spatial variables.

Industry-standard is the use of generalized linear models (GLM), introduced by \citet{Nelder1972}, as a predictive modeling tool for claim frequency and severity. For instance, \citet{HabermanSteven1996GLMa}, \citet{dejong2008generalized}, \citet{ohlsson2010non} and \citet{trufinbook1} apply GLMs for non-life insurance pricing. \citet{frees2008hierarchical} and \citet{antonio2010multilevel} convert the numerical inputs to categorical format for use in a frequency GLM. \citet{Henckaerts2018} present a data-driven method for constructing both a frequency and severity GLM on categorized input data, by combining evolutionary trees and generalized additive models to convert the numerical inputs to categorical variables. 

In recent years, machine learning techniques for actuarial purposes have been rising in popularity because of their strong predictive powers\addedtwo{, see, for example, \citet{Blier-Wong2021-hu} for an overview in insurance pricing and reserving.} Both \citet{wuthrich2021data} and \citet{trufinbook2} detail the use of tree-based models in an actuarial context. \citet{liu2014using} use Adaboost for claim frequency modeling. \cite{Henckaerts2021} compare the performance of decision trees, random forests, and gradient boosted trees for modeling claim frequency and severity. Moreover, their paper studies a range of interpretational tools to look under the hood of these predictive models and compares the resulting technical tariffs with managerial tools. Instead of modeling the claim frequency and severity independently, the total loss random variable can be modeled directly via a gradient boosting model with Tweedie distributional assumption, see \citet{yang2016} and \citet{hainauttweedieboost}. \citet{henckaerts2022added} combine tabular policy and policyholder specific information with telematics data in a gradient boosting model for usage-based pricing. \citet{Henckaerts2022} construct a global surrogate for a gradient boosting model (GBM) to translate the insights captured by the GBM into a tariff table. A benchmark study on six data sets then examines the robustness of the proposed strategy. 

Deep learning methods have been popular in the field of machine learning for many years. An early study of deep learning in an actuarial context is \citet{Dugas2003}, comparing the performance of a GLM, decision tree, neural network and a support vector machine for the construction of a technical insurance tariff. \citet{ferrario2020insights} use neural networks for frequency modeling and discuss various preprocessing steps. \citet{Wuthrich2019} compares the performance of neural networks and GLMs on a frequency case study. Both \citet{Wuthrich2019} and \citet{Schelldorfer2019} propose a combined actuarial neural network (CANN) for claim frequency modeling. The CANN starts with a GLM and builds a neural network adjustment on top of the GLM predictions, via a skip connection between input and output layer. \addedtwo{\citet{Shi2024-sparse} use deep learning for dependent frequency-severity modeling,  including a demonstration of regularization to select variables in the construction of the network. \citet{Shi2024-weather} incorporate dynamic weather information into a deep learning model to enhance pricing accuracy within a frequency-severity framework for a property damage portfolio.}

Categorical or factor data must be transformed into numerical representations in order to be utilized by neural networks \citep{guo2016entity}. This transformation is known in the literature as embedding, which maps categorical variables into numerical vectors. The choice of embedding technique can significantly impact the neural network's performance; see, for example, the claim severity study by \citet{kuo2021embeddings} on both feed-forward neural networks and transformer networks to demonstrate the enhanced performance when using embedding layers. Embedding layers are a type of supervised learning that allows a neural network to learn meaningful representations from the categorical inputs during the training of the neural network. \addedtwo{\citet{Shi2022-embedding} detail the use of embedding layers in neural network structures for insurance pricing. Their study demonstrates that models with embedding layers outperform those using one-hot encoding in terms of Gini index. Additionally, they show that embedding layers trained through supervised learning can be effectively transferred to models with limited data.} \citet{delong2021} suggest using autoencoders as an alternative method for categorical embedding using unsupervised learning. An autoencoder is a type of neural network that learns to compress and to reconstruct data in an unsupervised manner. Using an autoencoder, a compact, numerical representation of the factor input data results that can then be used in both frequency as well as severity modeling. \citet{delong2021} compare different setups of the autoencoder for claim frequency modeling and highlight the importance of normalization of the resulting numerical representation before using it in a feed-forward neural network. \citet{Meng2022} use the same technique in a claim frequency case study with telematic input data and extend the autoencoder with convolutional layers to process input data in image format.

Table \ref{tab_litoverview} gives an overview of the discussed literature on deep learning for insurance pricing. We list the treatment techniques applied to categorical input data, the model architectures used and the extent of the case studies covered by these papers. Lastly, we summarize the interpretation tools used by the authors to extract insights from the model architectures. 

\begin{table}[ht!]
\centering
\scriptsize
\begin{adjustwidth}{-1cm}{-1cm}
\setlength{\extrarowheight}{3pt} % a bit of extra whitespaces between lines in tables

\begin{NiceTabular}{r|P{3cm}P{2.8cm}P{1.2cm}P{1.5cm}P{2.5cm}}[
code-before = \rowcolor[HTML]{FFFFFF}{1,3,5,7,9,11,13,15,17}
              \rowcolor[HTML]{FAFAFF}{2,4,6,8,10,12,14,16}
]
\toprule
\textbf{Contribution} & \textbf{Categorical treatment} & \textbf{Model architecture} & \textbf{\# Data sets} & \textbf{Case study} & \textbf{Interpretation tools} \\[0.7ex]
\noalign{\hrule height 1.0pt}
\citet{Dugas2003} & $-$ & LR, GLM, DT, NN, SVM & $1$ & Tech. tariff & $-$ \\[0.7ex]
\noalign{\hrule height 0.3pt}
\citet{yang2016} & $-$ & TDBoost & $1$ & Tweedie compound & PDP, VIP \\[0.7ex]
\noalign{\hrule height 0.3pt}
\citet{Henckaerts2018} & $-$ & GLM & $1$ & Freq, sev & $-$ \\[0.7ex]
\noalign{\hrule height 0.3pt}
\citet{Wuthrich2019} &  Dummy encoding, embedding layers & GLM, NN, CANN & $1$ & Freq & Avg. neuron activation \\[0.7ex]
\noalign{\hrule height 0.3pt}
\citet{Schelldorfer2019} &  Embedding layers & CANN & $1$ & Freq & $-$ \\[0.7ex]
\noalign{\hrule height 0.3pt}
\citet{ferrario2020insights} & One-hot encoding & Boosted trees, NN & $1$ & Freq & $-$ \\[0.7ex]
\noalign{\hrule height 0.3pt}
\citet{noll2020} & Dummy encoding, empirical means, one-hot encoding & GLM, DT, Boosted trees, NN & $1$ & Freq & Loss-per-label \\[0.7ex]
\noalign{\hrule height 0.3pt}
\citet{Henckaerts2021} & $-$ & DT, RF, GBM & $1$ & Freq, sev, tech. tariff & PDP, VIP, ICE \\[0.7ex]
\noalign{\hrule height 0.3pt}
\citet{kuo2021embeddings} & One-hot encoding, embedding layers, attention layers & GLM, NN, Transformer, TabNET & $1$ & Sev & $-$ \\[0.7ex]
\noalign{\hrule height 0.3pt}
\citet{Meng2022} & Convolutional autoencoder & GLM & $1$ & Freq & $-$ \\[0.7ex]
\noalign{\hrule height 0.3pt}
\citet{Henckaerts2022} & $-$ & GBM & $6$ & Freq & PDP, SHAP, Surrogates \\[0.7ex]
\noalign{\hrule height 0.3pt}
\citet{Shi2022-embedding} & \addedtwo{One-hot encoding, embedding layers} & \addedtwo{NN} & \addedtwo{$1$} & \addedtwo{Freq} & \addedtwo{$-$} \\[0.7ex]
\noalign{\hrule height 0.3pt}
\citet{delong2021} &  Autoencoder & NN & $1$ & Freq & $-$ \\[0.7ex]
\noalign{\hrule height 0.3pt}
\citet{Shi2024-sparse} & \addedtwo{One-hot encoding, embedding layers} & \addedtwo{GLM, NN, deep-GLM}  & \addedtwo{$1$} & \addedtwo{Freq, sev, tech. tariff}  & \addedtwo{$-$} \\[0.7ex]
\noalign{\hrule height 0.3pt}
\citet{Shi2024-weather} & \addedtwo{Embedding layers}  & \addedtwo{NN} & \addedtwo{$1$} & \addedtwo{Freq, sev, tech. tariff} & \addedtwo{$-$} \\[0.7ex]
\noalign{\hrule height 0.3pt}
This paper & Autoencoder & GLM, GBM, NN, CANN & $4$ & Freq, sev, tech. tariff & PDP, VIP, surrogates, Shapley \\[0.7ex]
\bottomrule
\end{NiceTabular}
\caption{Overview of the literature on deep learning for insurance pricing. \emph{Categorical treatment} describes the preprocessing steps taken for categorical input variables. \emph{Model architecture} lists the different models compared in the paper. We give the number of benchmark data sets in \emph{\# Data sets} and the focus of the case study in \emph{Case study}. Lastly, \emph{Interpretational tools} list the tools used to look under the hood of the fitted models. The used abbreviations are: linear regression (LR), generalized linear model (GLM), decision tree (DT), neural network (NN), support vector machine (SVM), Tweedie boosted tree model (TDBoost), combined actuarial neural network (CANN), random forest (RF), gradient boosting model (GBM), deep generalized linear model (deep-GLM), frequency case study (freq), severity case study (sev), technical tariff structure (tech. tariff), partial dependency plot (PDP), variable importance plot (VIP), individual conditional expectation (ICE) and Shapley additive explanations (SHAP).}
\label{tab_litoverview}
\end{adjustwidth}
\end{table}

Historical claim data sets are often of tabular structure, meaning they can be represented in matrix notation, with each column representing an input variable and each row representing a vector of policy and policyholder information. Several papers recently questioned the performance of neural networks on tabular data. \citet{Borisov2021} compare 23 deep learning models on five tabular data sets and show how different tree-based ensemble methods outperform them. They highlight the predictive powers of combinations of gradient boosting models and neural network models, such as DeepGBM \citep{ke2019deepgbm}, which combines a GBM and a neural network for, respectively, numerical and categorical input features and TabNN \citep{ke2018tabnn}, which bins the input features based on a GBM and uses the resulting bins in a neural network. \citet{Shwartz-Ziv2021} analyze eight tabular data sets and compare five ensemble methods with four deep learning methods, concluding that the best performer combines gradient-boosted trees and a neural network. \citet{grinsztajn2022} compare the performance of gradient-boosted trees, a random forest and different neural network structures on 45 tabular data sets, highlighting the importance of data normalization and categorical treatment for deep learning models.

In light of these recent papers questioning the performance of deep learning architectures on tabular data, this paper aims to explore the added value of deep learning for non-life insurance pricing using tabular frequency and severity data. For this, we extend the analyses performed in \cite{Henckaerts2021} to deep learning models. Our study is an extension of the existing literature in five directions. First, we extend the CANN model architecture from \cite{Schelldorfer2019} by combining a GBM baseline with neural network adjustments. Moreover, we study both trainable and non-trainable adjustments. Second, we compare a neural network, the proposed CANN structures, and two benchmark models, a GLM and a GBM, by considering out-of-sample deviance, statistical performance and calibration criteria, and interpretation and managerial tools. The GLM is constructed on categorized input data, following the approach outlined in \citet{Henckaerts2018}; the GBM follows the setup from \citet{Henckaerts2021}. \added{We compare models not only based on out-of-sample deviance but look at the underlying structure and calibration of the predictions and use Murphy diagrams \citep{ehm2016quantiles} and Diebold-Mariano tests \citep{dieboldmariano} to see whether a model with lower deviance is really a statistical improvement. This provides a more nuanced understanding of model performance, moving beyond simple deviance comparison, towards a more robust and reliable model evaluation framework.} Third, we study the autoencoder embedding technique from \citet{delong2021} and highlight its importance in frequency-severity modeling. Because the autoencoder is trained in an unsupervised setting, the embedding can be learned on the frequency data and transferred to the severity setting, where we typically have fewer data points. Fourth, our case study is not limited to frequency modeling only but studies both frequency and severity modeling. We use four different insurance data sets to study the impact of sample size and the composition of the input data. Lastly, we use a set of interpretation techniques to capture insights from the constructed frequency and severity models and construct a GLM as a global surrogate for the neural networks, along the ideas of \citet{Henckaerts2022}. We compare the resulting technical tariffs based on their \added{(ordered) Lorenz curves and Gini indices} and look at the balance achieved by each model at portfolio level. This allows us to get a robust look at the possibilities of neural networks for frequency-severity pricing, from preprocessing steps to technical tariff.

\section{Technical insurance pricing: notation and set-up}

This paper assumes access to an insurance data set with tabular structure, meaning the data can be written in matrix notation, with each column representing a variable and each row representing a data point. We denote a data set as $\D = \left(\x_i,y_i\right)^{n}_{i=1}$, where each $\x_i$ is a $p$-dimensional data point with response $y_i$. Each data point $\x_i$ can be written as a vector $\left(x_{i,1},\ldots,x_{i,p}\right)$, where each entry $x_{i,j}$ represents the value of input variable $j$ for data point $i$. When not referencing a specific observation $i$, we often omit the subscript $i$ and write $(\x,y)$, with $\x=(x_1,\ldots,x_p)$, each $x_j$ representing a variable in our data set $\D$.

The variables in our data sets can be either numerical or categorical. Assuming $c$ categorical variables, we order the variables in $\D$ as follows:
\[\D = \big(\underbrace{x_1,\ldots,x_{p-c}\strut}_\text{numerical variables},\underbrace{x_{p-c+1},\ldots,x_p\strut}_\text{categorical variables},\underbrace{y\strut}_\text{response variable}\big).\]
Insurance data sets can also contain spatial information. A spatial variable is either numerical, i.e., latitude and longitude coordinates, or categorical, i.e., postal code of residence. We do not denote spatial variables separately, but count them as a numerical or categorical variable. When introducing a data set, we specify how the spatial information is encoded.

For frequency-severity modeling, we work with a frequency data set $\mathcal{D}^{\text{freq}}$ and a severity data set $\mathcal{D}^{\text{sev}}$, where a data point $\x_i$ represents information about policyholder $i$. In $\mathcal{D}^{\text{freq}}$, the response $y_i$ is the number of claims reported by policyholder $i$. The severity data set $\mathcal{D}^{\text{sev}}$ consists of the policyholders from $\mathcal{D}^{\text{freq}}$ who had at least one claim. In $\mathcal{D}^{\text{sev}}$ we use the average claim size over the corresponding reported claims as the response $y$. Because $\mathcal{D}^{\text{sev}}\subseteq\mathcal{D}^{\text{freq}}$, but with a different response, we often omit the superscript $\text{freq}$ and use $\mathcal{D}$ and $\mathcal{D}^{\text{sev}}$. We denote the number of observations as $n_f$ for $\D$ and $n_s$ for $\D^{\text{sev}}$, with $n_s\leq n_f$. Note that for both $\mathcal{D}$ and $\mathcal{D}^{\text{sev}}$, we have the same variables $x_1,\ldots,x_p$, except that we add the extra variable \emph{exposure-to-risk} $e$ to the frequency data set. Exposure is the fraction of the year the insurance covered the policyholder. This is only relevant for frequency modeling, hence we do not add this variable to the severity data set. In $\mathcal{D}^{\text{sev}}$ we do take into account the observed number of claims for each data point, to be used as a weight in the loss function. 

For a regression model $f(\cdot)$ with input covariates $\left(x_1,\ldots,x_p\right)$ and target response $y$, we write the model prediction for data point $i$ as $f\left(x_{i,1},\ldots,x_{i,p}\right) = \hat{y}_i$. We train $f(\cdot)$ on a training set, denoted as $\D^{\text{train}}\subset\D$, by choosing the model-specific parameters that minimize a chosen loss function $\sum_{i:\x_i\in\D^{\text{train}}}  \mathscr{L}\left(\hat{y}_i,y_i\right)$. The out-of-sample performance of a trained model is calculated on the test set $\D^{\text{test}} = \D\backslash\D^{\text{train}}$ as $\sum_{i:\x_i\in\D^{\text{test}}} \mathscr{L}\left(\hat{y}_i,y_i\right)$. We follow the loss functions proposed by \citet{wuthrich2021data} and \citet{Henckaerts2021} for modeling claim frequency and severity. For claim frequency modeling, where the claim count is typically assumed to be Poisson distributed, we use the Poisson deviance:

\begin{equation}
D_{\text{Poisson}}(f(\boldsymbol{x}),y) = \frac{2}{n_f} \sum_{i=1}^{n_f}\l y_i\ln\frac{y_i}{f(\boldsymbol{x}_i)}-(y_i-f(\boldsymbol{x}_i)) \r.
\label{eq_devpoisson}
\end{equation}

Note that when using the exposure-to-risk $e$ in the frequency model, we replace each prediction $f(\boldsymbol{x}_i)$ with $e_i\cdot f(\boldsymbol{x}_i)$ in the Poisson loss function. Claim severity data are often assumed to be long-tailed and right-skewed, so we use the gamma deviance given by

\begin{equation}
D_{\text{gamma}}(f(\boldsymbol{x}),y) = \frac{2}{n_s}\sum_{i=1}^{n_s} \alpha_i\l\frac{y_i-f(\boldsymbol{x}_i)}{f(\boldsymbol{x}_i)}-\ln\frac{y_i}{f(\boldsymbol{x}_i)} \r,
\label{eq_devgamma}
\end{equation}
where the weight $\alpha_i$ is the observed number of claims for data point $i$.

\section{Deep learning architectures and preprocessing steps} \label{sec_NNbased}

\subsection{Neural network architectures} \label{sec_NNarch}

\paragraph{Feed-forward neural network}
A feed-forward neural network (FFNN) is a type of machine learning model that utilizes interconnected layers, represented by $\boldsymbol{z}^{(m)}$ with $m=0,\ldots,M+1$. The input layer $\boldsymbol{z}^{(0)}$ provides the network with input data, while the output layer $\boldsymbol{z}^{(M+1)}$ gives the network's prediction. Between the input and output layers, there can be one or more hidden layers $\boldsymbol{z}^{(1)}, \ldots,\boldsymbol{z}^{(M)}$. When there are two or more hidden layers, we call the neural network a deep learning model. Each layer $\boldsymbol{z}^{(m)}$ consists of $q_m$ nodes and can be written as a vector $\boldsymbol{z}^{(m)} = \left(z^{(m)}_1,\ldots,z^{(m)}_{q_m}\right)$.

Each node in a layer, excluding the input layer, is connected to all nodes in the previous layer through weights $W_m\in\mathbb{R}^{q_m\times q_{m-1}}$, and a bias term $\boldsymbol{b}_m\in\mathbb{R}^{q_{m}}$. An activation function $\sigma^{(m)}(\cdot)$, $m=1,\ldots,M+1$, adds non-linearity to the network and allows it to learn complex relationships between inputs and outputs. The activation function is applied to the weighted sum of inputs to a node, along with its bias. Each layer $\boldsymbol{z}^{(m)}$ can be written in function of the previous layer as follows:
\begin{equation}
  \boldsymbol{z}^{(m)} = \sigma^{(m)}\left(W_m\cdot\boldsymbol{z}^{(m-1)} + \boldsymbol{b}_m\right). 
  \label{eq_nnlayer}
\end{equation}
Calculating the output of the FFNN in function of the input consists of performing a matrix multiplication for each layer and applying the activation function. The value of a layer $\boldsymbol{z}^{(m)}$ for input $\x_i$ is denoted as $\boldsymbol{z}^{(m)}_i$ and the value of a specific node $j$ as $z^{(m)}_{ij}$. When referencing a node without a specific input, we omit the subscript $i$ and write $z^{(m)}_j$.

\begin{figure}[ht]
\centering
  \includegraphics[scale=0.45]{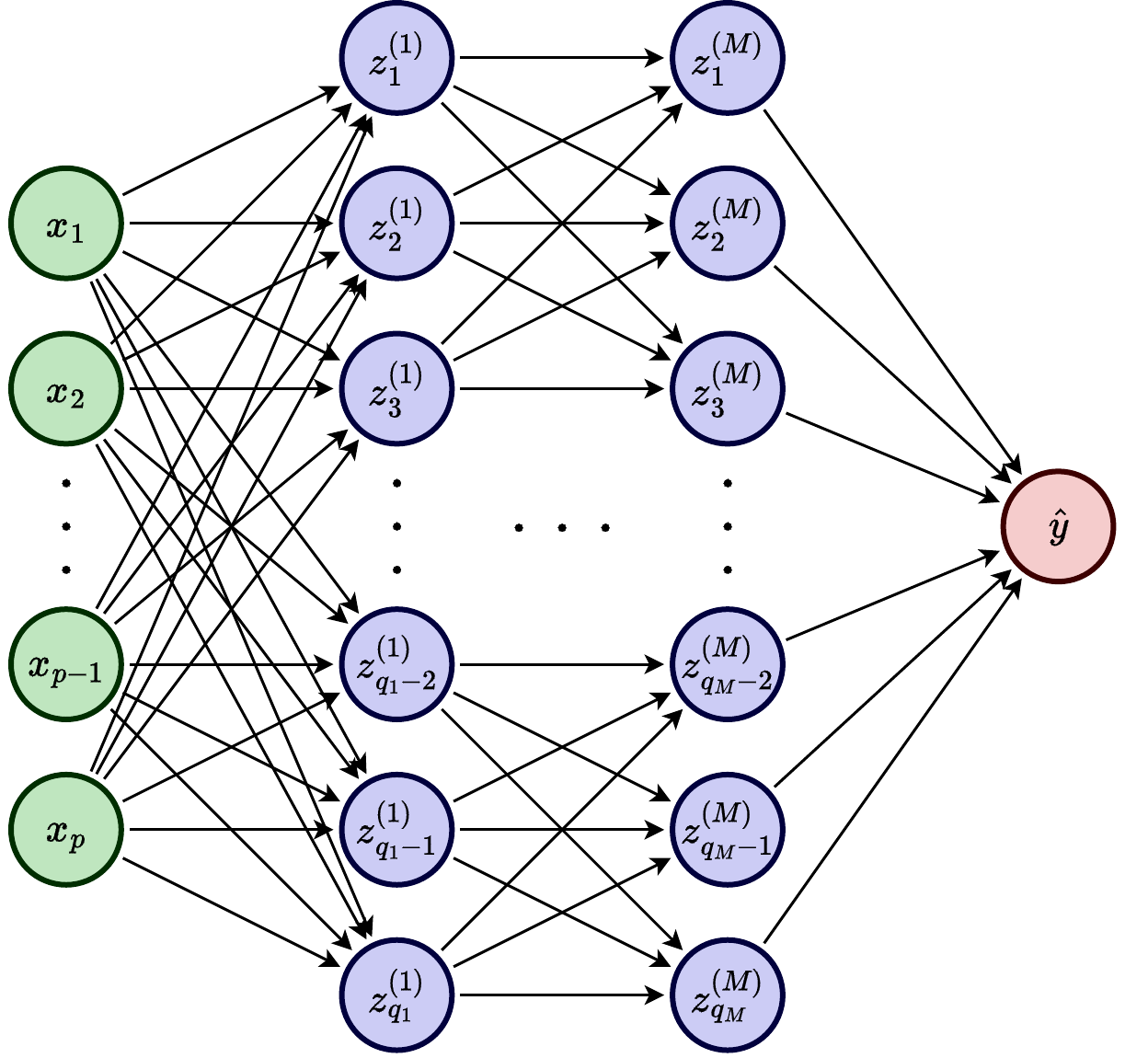}
\caption{Structure of a feed-forward neural network with $p$-dimensional input layer, hidden layers $\boldsymbol{z}^{(1)}, \ldots,\boldsymbol{z}^{(M)}$, with $q_1,\ldots,q_M$ nodes, respectively. The network has a single output node $\hat{y}$.}
\label{fig_ff_nn}
\end{figure}

The inputs of the neural network are the data points in a data set $\D$, the dimension $q_0$ of the input layer is equal to the number of variables $p$ in the data set.\footnote{The dimension of the input layer can be larger than $p$ when using an encoding technique, such as one-hot encoding.} We write the input layer as $\left(x_1,\ldots,x_p\right)$ to indicate that each node in the input layer represents an input variable from the data set. The target variable $y$ in our insurance data sets is one-dimensional, so the output layer $z^{(M+1)}$ has only one node and $q_{M+1} = 1$. We write the output node as $\hat{y}$. Figure \ref{fig_ff_nn} gives a schematic overview of a feed-forward neural network.

When modeling claim frequency and severity data with GLMs, an actuary typically relies on a Poisson GLM with a log-link function for frequency and a gamma GLM with a log-link function for severity modeling. To mimic this log-link relationship between covariates and output in the FFNN, we use an exponential activation function for the output layer in both the frequency and the severity model. As such, we obtain strictly positive predictions for claim counts and claim amounts.

\begin{figure}[ht]
\centering
  \includegraphics[scale=0.45]{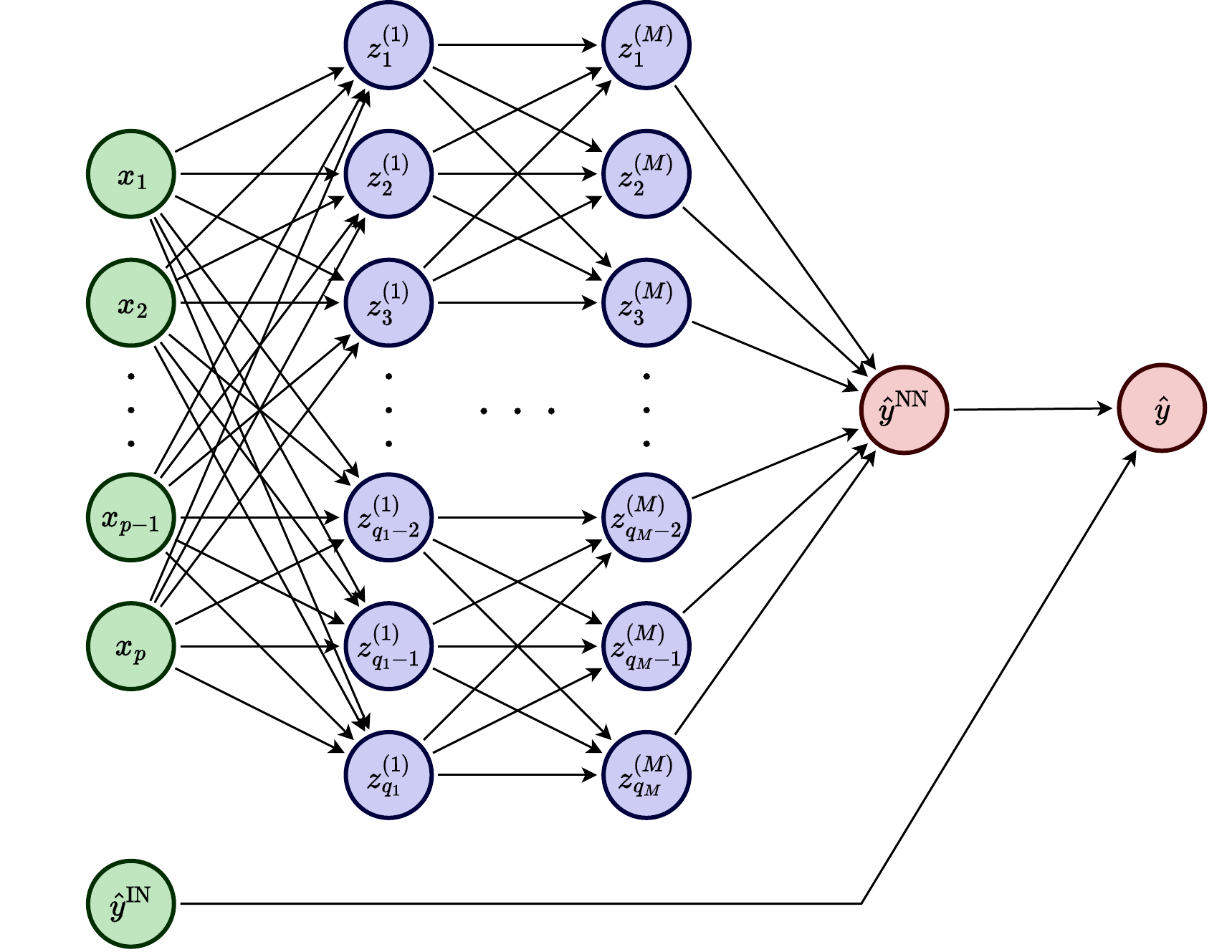}
\caption{Structure of a Combined Actuarial Neural Network (CANN). The initial model prediction $\hat{y}^{\text{IN}}$ is connected via a skip-connection to the output node of the FFNN.}
\label{fig_CANN}
\end{figure}

\paragraph{Combined actuarial neural networks}
\citet{Wuthrich2019} and \cite{Schelldorfer2019} propose a combination of a GLM with a FFNN, called the Combined Actuarial Neural Network (CANN). A CANN model calibrates a neural network adjustment on top of the GLM prediction. We refer to the GLM prediction as the \emph{initial model prediction}, denoted as $\hat{y}^{\text{IN}}$. We use $\hat{y}^{\text{IN}}$ as an input node in a FFNN but do not connect this node to the hidden layers. Instead, $\hat{y}^{\text{IN}}$ directly connects to the output node of the FFNN via a so-called skip connection. The adjustment made by the neural network on the initial model prediction is called the \emph{adjustment model prediction} and denoted as $\hat{y}^{\text{NN}}$. The combination of the initial model prediction and the adjustment calibrated by the neural net is the resulting CANN model prediction, denoted as $\hat{y}$. Figure \ref{fig_CANN} shows the structure of the CANN model. 

The output node of the CANN model, $\hat{y}$, is only connected to the initial model input $\hat{y}^{\text{IN}}$ and the neural network adjustment $\hat{y}^{\text{NN}}$. We use the exponential activation function in the output layer to ensure the log-link relationship between inputs and the predicted output. Because $\hat{y}^{\text{IN}}$ is a prediction at the level of the response, we apply a log transform on the initial model predictions. The output of the CANN model is then calculated as:

\begin{equation}
   \hat{y} = \exp\left(w_{\text{NN}}\cdot\hat{y}^{\text{NN}} + w_{\text{IN}}\cdot\ln\left(\hat{y}^{\text{IN}}\right) + b\right). 
   \label{eq_cannoutput}
\end{equation}

The case study in \cite{Schelldorfer2019} fixes the weights and bias in the output of the CANN as follows
\[w_{\text{NN}} = 1,\,w_{\text{IN}}=1\,\text{and}\,b=0.\]
Following \citet{gielis}, we call this the \emph{fixed} CANN, as the output weights are fixed and not trainable. In our case study, we also run experiments with trainable weights in the output layer and refer to this model as the \emph{flexible} CANN. This flexibility allows the training of the neural network to put more, or less, weight on the initial model prediction. This can potentially improve the predictive accuracy of the flexible CANN compared to the fixed CANN. Moreover, the initial model input is not restricted to GLM predictions and we will also run experiments in Section \ref{sec_oos} with an input prediction established with a carefully trained GBM. According to \citet{Henckaerts2021} the GBMs are capable of achieving a higher predictive accuracy compared to a GLM. Using the GBM predictions as initial model input can therefore potentially increase the performance of the CANN model, compared to a CANN using the GLM predictions. 

\subsection{Preprocessing steps} \label{sec_Prep}

\paragraph{Continuous variables}

We normalize the continuous input variables to ensure that each variable in the input data has a similar scale. This is important because most neural network training algorithms use gradient-based optimization, which can be sensitive to the scale of the input data \citep{sola_normalize}. For a continuous variable $x_j$ in the input data $\mathcal{D}$, we use normalization around zero as a scaling technique. Hereto, we replace each value $x_{i,j}$ as follows:

\begin{equation}
    x_{i,j}\mapsto\tilde{x}_{i,j}=\frac{x_{i,j}-\mu_{x_j}}{\sigma_{x_j}},
    \label{eq_normalization}
\end{equation}

where $\mu_{x_j}$ and $\sigma_{x_j}$ are the mean and standard deviation of the variable $x_j$ in the data set $\mathcal{D}$. When using a subset $\D^{\text{train}}\subset\D$ to train the model, we calculate the $\mu_{x_j}$ and $\sigma_{x_j}$ only on the data set $\D^{\text{train}}$ to avoid data leakage.

\paragraph{Categorical variables}
The FFNN and CANN models generate output by performing matrix multiplications and applying activation functions. Therefore, all inputs must be in numerical format. So-called embedding techniques convert categorical input variables to a numerical format. In this study, we utilize the \emph{autoencoder embedding} proposed by \citet{delong2021}. Autoencoders are neural networks commonly used for dimensionality reduction \citep{Goodfellow-et-al-2016}. They consist of two components: an encoder and a decoder. The encoder maps a numerical input vector to a lower-dimensional representation, while the decoder reconstructs the original input from this representation. During training, the autoencoder minimizes the difference between the original and reconstructed inputs, resulting in an encoder that captures the most important characteristics of the data.

\begin{figure}[ht]
\begin{adjustwidth}{-1.2cm}{-1.2cm}
\centering
\begin{subfigure}[t]{0.33\linewidth}
  \hspace{0.6em}
  \includegraphics[height=8cm, keepaspectratio]{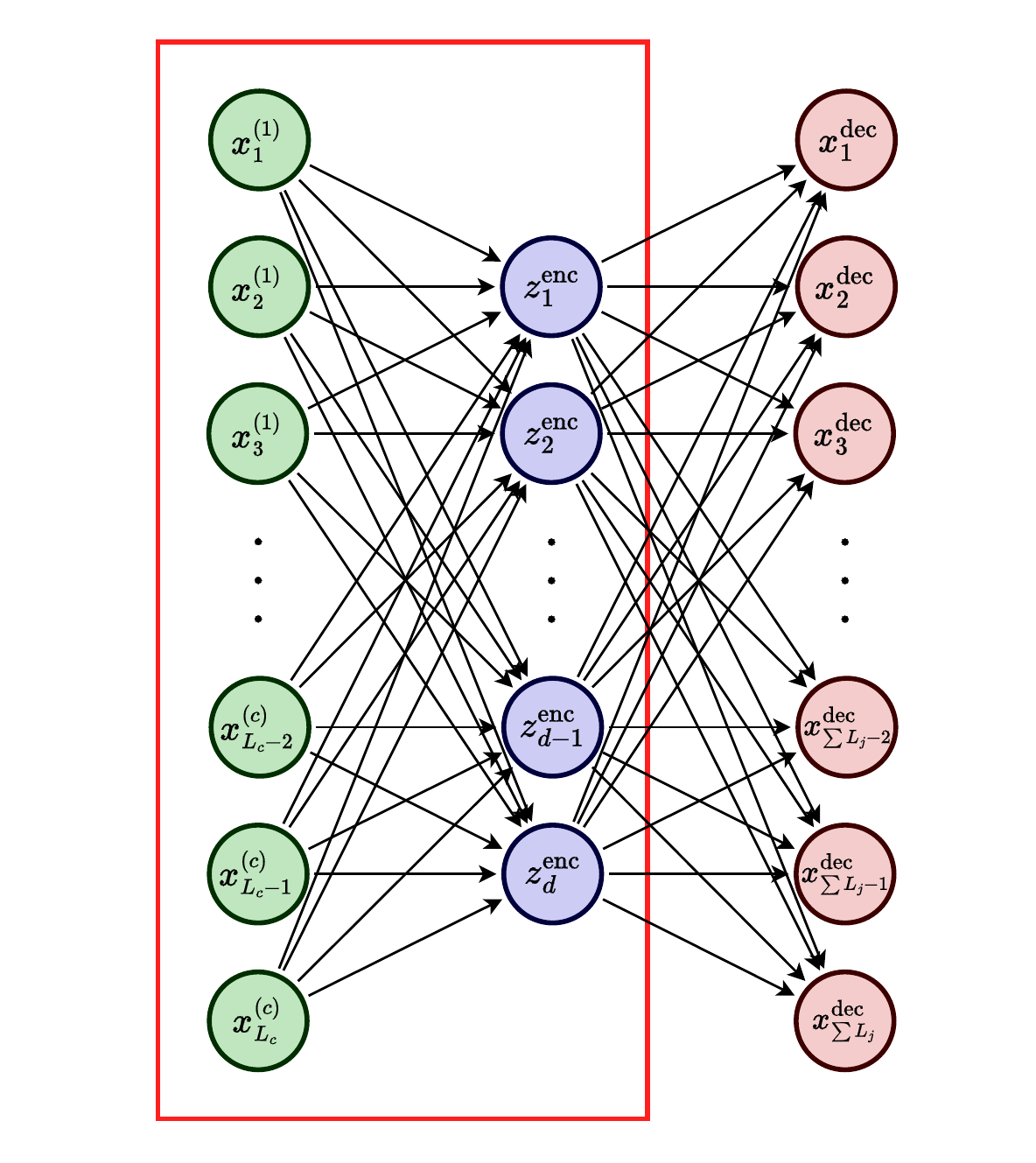}
  \caption{General structure of an autoencoder neural network. The connections between the input layer and the encoded layer are called the \emph{encoder}, highlighted in red, the connections between encoded layer and decoded output layer are called the \emph{decoder}. After training, the encoded vector is a $d$-dimensional representation of the input.}
  \label{fig_cann_with_ae_A}
\end{subfigure}%
\hspace{1em}
\begin{subfigure}[t]{0.64\linewidth}
  %\hspace{1.4em}
  \includegraphics[height=8cm, keepaspectratio]{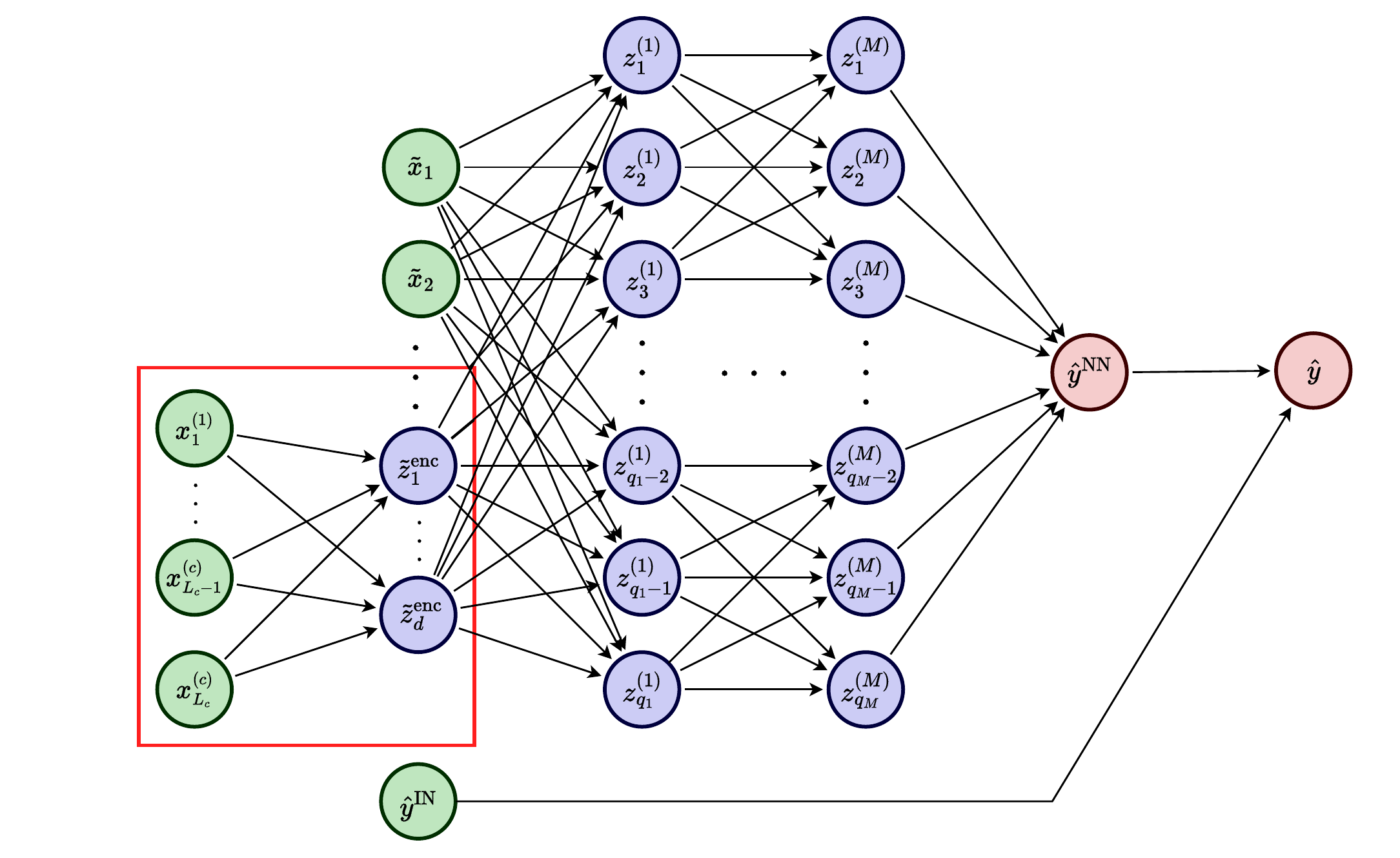}
  \caption{The CANN model where the last input nodes represent the one-hot encoding of the categorical variables in the data set. Using the encoder part of the pretrained autoencoder, shown with the red box, we embed the categorical variables into a $d$-dimensional numerical representation, and connect the embeddings to the CANN model structure.}
  \label{fig_cann_with_ae_B}
\end{subfigure}
\end{adjustwidth}\caption{Our proposed network structure combines the autoencoder embedding technique from \cite{delong2021} and the CANN structure from 
  \cite{Schelldorfer2019}.}
\label{fig_cann_with_ae}
\end{figure}

Figure \ref{fig_cann_with_ae_A} shows the general structure of such an autoencoder. It consists of an input layer, one hidden layer $\boldsymbol{z}^{\text{enc}}$ of dimension $d$, and an output layer of the same dimension as the input layer. The encoding layer is defined by the activation function $\sigma^{(\text{enc})}(\cdot)$, weight matrix $W_{\text{enc}}$ and bias vector $\boldsymbol{b}_{\text{enc}}$. Similarly, the output layer is defined by activation function  $\sigma^{(\text{dec})}(\cdot)$, weight matrix $W_{\text{dec}}$ and bias vector $\boldsymbol{b}_{\text{dec}}$. For an input $\boldsymbol{x}_i$, the encoded and decoded representations are calculated as

\begin{equation}
\begin{aligned}
     & \boldsymbol{z}_i^{\text{enc}} = \sigma^{(\text{enc})}\left(W_{\text{enc}}\cdot\boldsymbol{x}_i+\boldsymbol{b}_{\text{enc}}\right), \\
     & \boldsymbol{x}_i^{\text{dec}} = \sigma^{(\text{dec})}\left(W_{\text{dec}}\cdot\boldsymbol{z}_i^{\text{enc}}+\boldsymbol{b}_{\text{dec}}\right). \\
\end{aligned}   
\label{eq_ae_calculation}
\end{equation}

\iffalse The autoencoder is trained on all data points $\boldsymbol{x}_i$ in a data set $\mathcal{D}$ by adjusting the weight matrices and bias vectors in order to minimize a chosen loss function $\sum_{\boldsymbol{x}_i\in\mathcal{D}}^{}\mathscr{L} \left( \boldsymbol{x}^{\text{dec}}_i, \boldsymbol{x}_i \right)$. \fi

To set-up an autoencoder for embedding multiple categorical input variables, we first construct the one-hot encoded representation of each categorical variable \citep{ferrario2020insights}. One-hot encoding maps a categorical variable $x_j$ with $L_j$ levels to a binary vector $x_j^{\text{OH}} = \left(x^{(j)}_1,\ldots,x^{(j)}_{L_j}\right)$ in the space $\{0,1\}^{L_j}$. If we have $c$ categorical variables, the dimension of all one-hot representations together equals $\sum_{j=1}^{c}L_j$.

Second, we train an autoencoder using the one-hot representations of the categorical variables as input nodes. As such, the input layer has a dimension of $\sum_{j=1}^{c}L_j$. The input layer is connected to an encoded layer of dimension $d$, which is then connected back to the output layer of dimension $\sum_{j=1}^{c}L_j$. We use the identity function as activation function for both $\sigma^{(\text{enc})}(\cdot)$ and $\sigma^{(\text{dec})}(\cdot)$ in Equation \eqref{eq_ae_calculation}.

\iffalse Following the construction in \citet{delong2021}, we apply a softmax transformation on the output layer of the autoencoder after the activation function $\sigma^{(\text{dec})}(\cdot)$. For each categorical variable $x_j$, exactly one value in the input nodes $\left(x^{(j)}_1,\ldots,x^{(j)}_{L_j}\right)$ is one and the rest of the input nodes takes the value zero. \fi

Following \citet{delong2021}, we apply a softmax activation function to the output layer $\boldsymbol{x}_i^{\text{dec}}$ of the autoencoder as calculated by Equation \eqref{eq_ae_calculation}. For each categorical variable $x_j^{\text{OH}} = \left(x^{(j)}_1,\ldots,x^{(j)}_{L_j}\right)$ and for each $h\in\{1,\ldots,L_j\}$, the softmax transformation of the output node $x^{(j,\text{dec})}_h$ is defined as:

\begin{equation}
   x^{(j,\text{dec})}_h\mapsto\tilde{x}^{(j,\text{dec})}_h = \frac{\exp\left(x^{(j,\text{dec})}_h\right)}{\exp\left(x^{(j,\text{dec})}_1+\ldots+x^{(j,\text{dec})}_{L_j}\right)},\qquad h=1,\ldots,L_j.
   \label{eq_softmaxtranform}
\end{equation}

The use of the softmax activation function ensures that the values of the decoded vectors $\left(x^{(j,\text{dec})}_1,\ldots,x^{(j,\text{dec})}_{L_j}\right)$ sum up to one for each variable $x_j$. 

To train the autoencoder, we use the cross-entropy loss function, which is suitable because of the $0/1$ values in its input data. With $\x_i^{\text{OH}}$ the one-hot encoding of all categorical variables for policyholder $i$ and $\tilde{\boldsymbol{x}}^{\text{dec}}_i$ the values of the autoencoder's output layer for policyholder $i$, the cross-entropy loss function is defined as:

\begin{equation} \sum_{i:\boldsymbol{x}_i\in\mathcal{D}} \mathscr{L}^{\text{CE}} \left( \tilde{\boldsymbol{x}}^{\text{dec}}_i, \boldsymbol{x}_i^{\text{OH}} \right) = \sum_{i:\boldsymbol{x}_i\in\mathcal{D}} \left(- \sum_{j=1}^{c} \sum_{h=1}^{L_j} x_{ih}^{(j)}\cdot\log\left(\tilde{x}^{(j,\text{dec})}_{ih}\right)\right) .
\label{eq_bincrossentropy}
\end{equation}

When  training the autoencoder until an acceptable low loss between $\boldsymbol{x}_i^{\text{OH}}$ and $\tilde{\boldsymbol{x}}^{\text{dec}}_i$, the vector of categorical inputs $(x_{i,p-c+1},\ldots,x_{i,p})$ is represented in the vector $(z^{\text{enc}}_{i1},\ldots,z^{\text{enc}}_{id})$ in an accurate, compact and numerical way. We call the vector $\z^{\text{enc}}_i$ the embedding of the categorical inputs of $\x_i$. To use the embedding together with the numerical features of $\x_i$, we normalize the values in the nodes $z^{\text{enc}}_1,\ldots,z^{\text{enc}}_d$ by scaling the weight matrix $W_{\text{enc}}$ and bias vector $\boldsymbol{b}_{\text{enc}}$ of the trained encoder. With $\mu_1,\ldots,\mu_d$ the means, and $\sigma_1,\ldots,\sigma_d$ the standard deviations, of the values  $z^{\text{enc}}_{i1},\ldots,z^{\text{enc}}_{id}$ for all $\x_i\in\D^{\text{train}}$, we scale the weight matrix $W_{\text{enc}}$ and bias vector $\boldsymbol{b}_{\text{enc}}$ of the pre-trained encoder as follows:

\begin{equation}
  W_{\text{enc}}\mapsto \tilde{W}_{\text{enc}} = \begin{pmatrix}
\frac{w_{11}}{\sigma_1} & \frac{w_{12}}{\sigma_1} & \ldots & \frac{w_{1\sum L_j}}{\sigma_1}\\
\frac{w_{21}}{\sigma_2} & \frac{w_{12}}{\sigma_2} & \ldots & \frac{w_{1\sum L_j}}{\sigma_2}\\
\vdots & \vdots & \ddots & \vdots \\
\frac{w_{d-1,1}}{\sigma_{d-1}} & \frac{w_{d-1,2}}{\sigma_{d-1}} & \ldots & \frac{w_{d-1,\sum L_j}}{\sigma_{d-1}}\\
\frac{w_{d1}}{\sigma_d} & \frac{w_{d2}}{\sigma_d} & \ldots & \frac{w_{d\sum L_j}}{\sigma_d}\\
\end{pmatrix} , \boldsymbol{b}_{\text{enc}}\mapsto \tilde{\boldsymbol{b}}_{\text{enc}} = \begin{pmatrix}
\frac{b_1 - \mu_1}{\sigma_{1}}\\
\frac{b_2 - \mu_2}{\sigma_{2}}\\
\vdots \\
\frac{b_{d-1} - \mu_{d-1}}{\sigma_{d-1}}\\
\frac{b_d - \mu_d}{\sigma_{d}}\\
\end{pmatrix}.  
\label{eq_normmatrix}
\end{equation}

The normalised embedding $\tilde{\boldsymbol{z}}_i^{\text{enc}}$ is then calculated as

\begin{equation}
\begin{aligned}
     & \tilde{\boldsymbol{z}}_i^{\text{enc}} = \sigma^{(\text{enc})}\left(\tilde{W}_{\text{enc}}\cdot\boldsymbol{x}_i+\tilde{\boldsymbol{b}}_{\text{enc}}\right). \\
\end{aligned}   
\label{eq_emb_normalised}
\end{equation}

Having access to the trained and scaled autoencoder, we now add the encoder part to the FFNN and the CANN structures by replacing the input nodes of the categorical variables in Figure \ref{fig_ff_nn} and \ref{fig_CANN} with the encoding part of the trained autoencoder, as shown in Figure \ref{fig_cann_with_ae_B} for the CANN. For clarity, we omit the one-hot encoding notation of each variable in Figure \ref{fig_cann_with_ae}. We say the autoencoder is \emph{pre-trained} because we perform a first training and scaling of the autoencoder before training the neural network architectures equipped with the added encoder. Adding the encoder to the network allows the network to finetune the weights and biases of the pre-trained encoder with respect to the considered regression task and its applicable loss function as in Equation \eqref{eq_devpoisson} or Equation \eqref{eq_devgamma}.

Autoencoders used to embed categorical variables provide several advantages over one-hot encoding \citep{delong2021}. Firstly, they allow for a significantly smaller dimension of the encoding compared to the dimension resulting from one-hot encoding. Secondly, autoencoders enable the encoding of all categorical variables together, capturing interactions between variables more effectively than variable specific encoding does. Lastly, autoencoders prove advantageous in multi-task scenarios such as frequency-severity modeling. Learning to encode categorical variables solely on the severity dataset can be problematic due to its smaller size. Since autoencoders are unsupervised learning methods, we can train the autoencoder using all data available, and add the resulting pre-trained encoder to both frequency and severity models. 

\subsection{Training and tuning neural networks} \label{sec_Train}

%\paragraph{Training and tuning the deep learning structures}
We train the FFNN and CANN models using the Adam optimization algorithm. Adam, introduced by \citet{kingma2014adam}, is a stochastic gradient descent algorithm with an adaptive learning rate. Iteratively, the Adam algorithm changes the weights and biases in the network to minimize the loss between predictions and the observed responses. We use batches of training data for each training iteration to speed up optimization; see \citet{keskar2016}. The size of the batches is a parameter that needs to be tuned. The network size is also tuned; the number of hidden layers $M$, and the number of nodes in each layer $q_1,\ldots,q_M$ are tuning parameters. We use a drop-out rate \citep{dropout2014} to avoid overfitting, and consider this rate to be a tuning parameter as well. The drop-out rate is the percentage of nodes in each layer that are disconnected from the next and previous layer during each iteration of the Adam algorithm. Finally, the choice of the activation functions $\sigma^{(1)}(\cdot),\ldots,\sigma^{(M)}(\cdot)$ is a tuning parameter. To simplify the tuning process, we use layers of equal sizes, $q_1=\ldots=q_M=q$, and apply the same activation function for all hidden layers, $\sigma^{(1)}(\cdot)=\ldots=\sigma^{(M)}(\cdot)=\sigma(\cdot)$. Hence, only the value for $q$ and the choice of the activation function $\sigma(\cdot)$ are tuned and applied to each hidden layer. 

\begin{figure}[ht!]
    \centering
    \includegraphics[width = 0.5\textwidth]{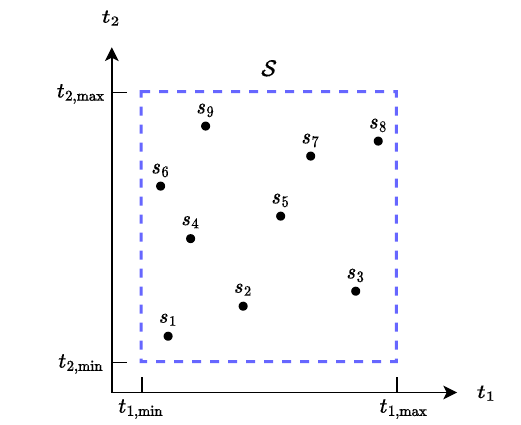}
    \caption{Example of random grid search with two tuning parameters $t_1$ and $t_2$. The search space $\mathcal{S} = \left[t_{1,\text{min}},t_{1,\text{max}}\right]\times\left[t_{2,\text{min}},t_{2,\text{max}}\right]$ is shown in the figure by the dotted square. The random grid $\mathcal{R}$ consists of nine randomly drawn $s_1,\ldots,s_9$ from $\mathcal{S}$. The optimal $s^{*}\in\mathcal{R}$ is then selected via a cross-validation scheme.}
    \label{fig_randomgrid_example}
\end{figure}

%\paragraph{Random grid search}
We deploy a random grid search, introduced by \citet{Bergstra2012}, to determine the optimal tuning parameters. For each tuning parameter $t_k$, with $k=1,\ldots,K$, and $K$ the total number of tuning parameters under consideration, we define a range of possible values $\left[t_{k,\text{min}},t_{k,\text{max}}\right]$. The search space $\mathcal{S}$ is the space consisting of all possible values for all tuning parameters:
$$\mathcal{S} = \left[t_{1,\text{min}},t_{1,\text{max}}\right] \times\ldots\times\left[t_{K,\text{min}},t_{K,\text{max}}\right].$$ 
The \emph{random grid} $\mathcal{R}\subset\mathcal{S}$ consists of randomly drawn points in the search space $\mathcal{S}$. Each $s\in\mathcal{R}$ represents a set of candidate tuning parameter values. Out of the random grid $\mathcal{R}$, we select the optimal $s^{*}$ via a cross-validation scheme. In Figure \ref{fig_randomgrid_example}, we give an example of a search space defined by two tuning parameters and a random grid of size nine sampled in the search space.

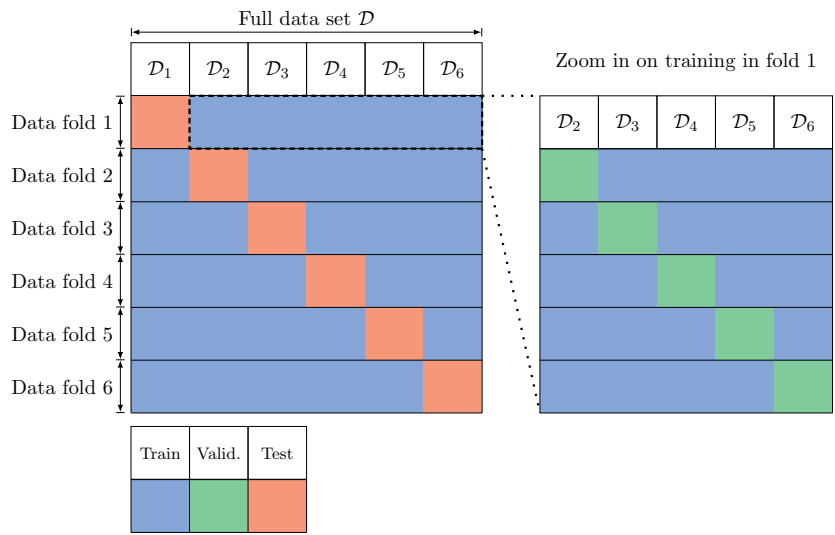
\begin{figure}[ht!]
	\centering
    \scalebox{.7}{	\begin{tikzpicture}
	\matrix (M) [matrix of nodes,
	nodes={minimum height = 1cm, minimum width = 1.1cm, outer sep=0, anchor=center, draw},
	%column 1/.style={nodes={draw=none}, minimum width = 4cm},
	%row sep=-\pgflinewidth, 
	column sep=-\pgflinewidth, %nodes in empty cells,
	g/.style={fill=Green!50, draw=none},
	b/.style={fill=NavyBlue!50, draw=none},
	r/.style={fill=Red!50, draw=none},
	w/.style={fill=White!50, draw=none}
	]
	{
		{$\mathcal{D}_1$} & {$\mathcal{D}_2$} & {$\mathcal{D}_3$} & {$\mathcal{D}_4$} & {$\mathcal{D}_5$} & {$\mathcal{D}_6$} &  & |[w]|  & |[w]| & |[w]| & |[w]| & |[w]| \\[-0.05ex]
		|[r]| & |[b]| & |[b]| & |[b]| & |[b]| & |[b]| & |[w]| & {$\mathcal{D}_2$} & {$\mathcal{D}_3$} & {$\mathcal{D}_4$} & {$\mathcal{D}_5$} & {$\mathcal{D}_6$} \\
		|[b]| & |[r]| & |[b]| & |[b]| & |[b]| & |[b]| & |[w]| & |[g]| & |[b]| & |[b]| & |[b]| & |[b]| \\
		|[b]| & |[b]| & |[r]| & |[b]| & |[b]| & |[b]| & |[w]| & |[b]| & |[g]| & |[b]| & |[b]| & |[b]| \\
		|[b]| & |[b]| & |[b]| & |[r]| & |[b]| & |[b]| & |[w]| & |[b]| & |[b]| & |[g]| & |[b]| & |[b]| \\
		|[b]| & |[b]| & |[b]| & |[b]| & |[r]| & |[b]| & |[w]| & |[b]| & |[b]| & |[b]| & |[g]| & |[b]| \\
		|[b]| & |[b]| & |[b]| & |[b]| & |[b]| & |[r]| & |[w]| & |[b]| & |[b]| & |[b]| & |[b]| & |[g]|  \\[1.5ex]
		{\footnotesize Train}&{\footnotesize Valid.}&{\footnotesize Test}&&&&&&&&\\
		|[b]|&|[g]|&|[r]|&&&\\
	};
	
	% Annotations on data folds
	\draw (M-1-1.north west) ++(0,2mm) coordinate (LT) edge[|<->|, >= latex] node[above]{Full data set $\mathcal{D}$} (LT-|M-1-6.north east);
	\draw[|<->|, >= latex, transform canvas={xshift=-2mm}] (M-2-1.north west) -- (M-2-1.south west) node[midway,left]{Data fold 1};
	\draw[<->|, >= latex, transform canvas={xshift=-2mm}] (M-3-1.north west) -- (M-3-1.south west) node[midway,left]{Data fold 2};
	\draw[<->|, >= latex, transform canvas={xshift=-2mm}] (M-4-1.north west) -- (M-4-1.south west) node[midway,left]{Data fold 3};
	\draw[<->|, >= latex, transform canvas={xshift=-2mm}] (M-5-1.north west) -- (M-5-1.south west) node[midway,left]{Data fold 4};
	\draw[<->|, >= latex, transform canvas={xshift=-2mm}] (M-6-1.north west) -- (M-6-1.south west) node[midway,left]{Data fold 5};
	\draw[<->|, >= latex, transform canvas={xshift=-2mm}] (M-7-1.north west) -- (M-7-1.south west) node[midway,left]{Data fold 6};
	
	% Annotations on tuning
	%\draw (M-2-8.north west) ++(0,2mm) coordinate (LT) edge[] node[above]{Zoom in on training in data fold 1} (LT-|M-2-12.north east);
	\node[fit=(M-1-8)(M-1-12)]{Zoom in on training in fold 1};
	\draw[very thick, densely dashed, black] (M-2-2.north west) -- (M-2-6.north east) -- (M-2-6.south east) -- (M-2-2.south west) -- (M-2-2.north west)  ;
	\draw[loosely dotted, very thick] (M-2-6.north east) -- (M-2-8.north west);
	\draw[loosely dotted, very thick] (M-2-6.south east) -- (M-7-8.south west);
	
	% Borders
	% data folds
	\draw[] (M-2-1.north west) -- (M-7-1.south west);
	\draw[] (M-2-6.north east) -- (M-7-6.south east);
	%\draw[] (M-2-1.north west) -- (M-2-6.north east);
	\draw[] (M-7-1.south west) -- (M-7-6.south east);
	% tuning
	\draw[] (M-3-8.north west) -- (M-7-8.south west);
	\draw[] (M-3-12.north east) -- (M-7-12.south east);
	\draw[] (M-3-8.north west) -- (M-3-12.north east);
	\draw[] (M-7-8.south west) -- (M-7-12.south east);
	% legend
	\draw[] (M-9-1.north west) -- (M-9-1.south west);
	\draw[] (M-9-3.north east) -- (M-9-3.south east);
	\draw[] (M-9-1.north west) -- (M-9-3.north east);
	\draw[] (M-9-1.south west) -- (M-9-3.south east);
	\draw[] (M-9-1.north east) -- (M-9-1.south east);
	\draw[] (M-9-2.north east) -- (M-9-2.south east);
	
	% Horizontal lines
	% data folds
	\draw[] (M-2-1.south west) -- (M-2-6.south east);
	\draw[] (M-3-1.south west) -- (M-3-6.south east);
	\draw[] (M-4-1.south west) -- (M-4-6.south east);
	\draw[] (M-5-1.south west) -- (M-5-6.south east);
	\draw[] (M-6-1.south west) -- (M-6-6.south east);
	% tuning
	\draw[] (M-3-8.south west) -- (M-3-12.south east);
	\draw[] (M-4-8.south west) -- (M-4-12.south east);
	\draw[] (M-5-8.south west) -- (M-5-12.south east);
	\draw[] (M-6-8.south west) -- (M-6-12.south east);
	
	\end{tikzpicture}}
	\caption{Representation of the $6$ times $5$-fold cross-validation scheme; figure from \citet{Henckaerts2021}.}
	\label{fig_crossval}
\end{figure}

%\paragraph{Cross-validation scheme}
We use the extensive cross-validation scheme proposed by \citet{Henckaerts2021}, as sketched in Figure \ref{fig_crossval}. We divide the data set $\mathcal{D}$ in six disjoint and stratified subsets $\mathcal{D}_1,\ldots,\mathcal{D}_6$. We define six data folds; in data fold $\ell$, for $\ell=1,\ldots,6$, we select a hold-out test set $\mathcal{D}_{\ell}$ and use five-fold cross-validation \citep{Freidmanetal2001} on the data set $\mathcal{D}\backslash\mathcal{D}_{\ell}$. Each cross-validation loop uses four out of the five data subsets in $\mathcal{D}\backslash\mathcal{D}_{\ell}$ to train the neural network. The fifth subset is used both for early stopping and to calculate the validation error. The cross-validation error is the average validation error over the five validation sets. We then determine the optimal $s^{*}_{\ell}\in\mathcal{R}$ which minimizes the cross-validation error for data fold $\ell$. We use the six optimal tuning parameter sets $s_{1}^{*},\ldots,s_{6}^{*}$ to determine the out-of-sample performance on the test set $\mathcal{D}_1,\ldots,\mathcal{D}_6$ of each data fold $\ell=1,\ldots,6$. As such, we obtain an out-of-sample prediction for every data point in the data set.

\section{Performative comparison between benchmark models and deep learning architectures}\label{sec_Bench}

Section \ref{sec_datasets} introduces four data sets that are used in our benchmark study. In this study we compare the performance of the deep learning architectures against two benchmark models introduced in Section \ref{sec_benchmodel}. Section \ref{sec_nnsetup} covers the tuning parameter grid used for both the autoencoder and the deep learning architectures. Section \ref{sec_onehotcomparison} compares the autoencoder embedding against the one-hot encoding when used in the deep learning models under consideration. Lastly, \added{Section \ref{sec_oos} sketches our proposed model evaluation framework and compares the models under consideration.}

\subsection{Data sets} \label{sec_datasets}

The used data sets are an Australian, Belgian, French\footnote{The French data set in the \citet{Schelldorfer2019} package contains $35\,560$ claims, but only $24\,000$ claims have a claim amount. We exclude the policies with claims but without claim amount from our study.} and Norwegian motor-third-party-liability (MTPL) data set. \addedtwo{The French dataset is available through \citet{Schelldorfer2019}, the Australian, Belgian, and Norwegian data sets through the R packages \texttt{CASdatasets} \citep{casdatasets} and \texttt{maidrr} \citep{henckaerts2022added,maidrrdatasets}.} Table \ref{tab_datasets} gives an overview of the number of data points for each data set and the number of continuous, categorical and spatial variables. \added{These four data sets were chosen for the benchmark study for three reasons. First, the data sets have been used in many other studies, facilitating a comparison of the results obtained with the existing literature. Second, the four data sets provide a mix of large and small data sets, which is relevant to explore the link between data set size and performance of neural network structures; see \citet[Section 5.2]{Goodfellow-et-al-2016}. Third, the data sets have different amounts of continuous, categorical, and spatial variables allowing us to investigate the effect of the data set structure on model performance.}

\setlength{\extrarowheight}{3pt} % a bit of extra whitespaces between lines in tables
\begin{table}[!h]
%\small
\centering
\begin{NiceTabular}{p{0.2cm}rrrrr}[
code-before = \rowcolor[HTML]{FFFFFF}{1,2,4,5,7,9}
              \rowcolor[HTML]{FAFAFF}{3,6,8}
]
\toprule
& & Australian MTPL & Belgian MTPL & French MTPL & Norwegian MTPL \\
\noalign{\hrule height 0.3pt}
\noalign{\medskip}
\multicolumn{6}{l}{\textbf{Number of observations}} \\
\noalign{\hrule height 0.3pt}
& Frequency & $67\,856$  & $163\,212$ & $668\,897$  & $183\,999$ \\
& Severity & $4\,624$  & $18\,276$ & $24\,944$ & $8\,444$ \\
\noalign{\hrule height 0.3pt}
\noalign{\medskip}
\multicolumn{6}{l}{\textbf{Covariates: number and type}} \\
\noalign{\hrule height 0.3pt}
& Continuous & 1 & 4 & 2 & 0 \\
& Categorical & 4 & 5 & 5 & 3\\
& Spatial & 0 & 1 & 2 & 1\\
\bottomrule
\end{NiceTabular}
\caption{Overview of the structure of the data sets used in the benchmark study. The number of data points for both frequency and severity modeling is given, as well as the number of different input variables per type. \added{The acronym MTPL stands for motor-third-party-liability.} }
\label{tab_datasets}
\end{table}

The spatial variables are listed separately in Table \ref{tab_datasets}. The Belgian spatial variable is the postal code, which is converted to two continuous variables, the latitude and longitude coordinates of the center of that postal code. The French data set includes two spatial variables: the French district, which is categorical, and the logarithm of the population density of the place of residence of the policyholder, which is a continuous variable. The Norwegian data set has one spatial variable denoting the population density of the region of residence as a categorical variable. \added{Appendix \ref{app_datasets} gives an overview of all variables in each data set.}

\subsection{Benchmark models} \label{sec_benchmodel}

To enable an assessment of the predictive performance of the neural network and CANN structures we construct two benchmark models: a generalized linear model (GLM) and a gradient boosting model (GBM), for both frequency as well as severity. Predictions from these benchmark models are then also used as the initial model inputs in the CANN models. For the Belgian data set we use the GLM constructed in \cite{Henckaerts2018} and the GBM from \cite{Henckaerts2021}.

For the construction of the GLM, we follow the strategy proposed in \citet{Henckaerts2018} and start from a generalized additive model (GAM), including interaction effects between continuous variables. Based on the insights from the GAM, we bin the continuous variables using a regression tree. On the binned input data, we fit a GLM. We repeat the construction of the GLM six times, each time withholding a subset $\mathcal{D}_{\ell}$, $\ell=1,\ldots,6$. This way, we obtain GLM-based out-of-sample predictions for all observations in the data set $\mathcal{D}$. 

GBM is an ensemble method combining multiple decision trees \citep{friedman2001greedy}. A GBM has two tuning parameters: the number of trees and the depth of each tree. We use a tuning grid with the following values:

$$ \text{Number of trees:}\, \{100,300, 500,\ldots,5\,000\},$$
$$ \text{Depth of each tree:}\, \{1,2,3,4,5,6,7,8,9,10\}. $$

We use three hyperparameters of which the values are not tuned: shrinkage $=0.01$, bagging fraction $=0.75$, and minimum observations per node is $0.75\%$ of the number of data points in the training data. The loss functions in Equation \eqref{eq_devpoisson} and Equation \eqref{eq_devgamma} are used for, respectively, frequency and severity modeling. We follow the repeated $5$-fold cross-validation scheme as described in Section \ref{sec_Train}. With optimal tuning parameters, we fit a GBM for each data fold and obtain the predictions for the observations in the corresponding test set. As such, we obtain an out-of-sample prediction for every data point in the portfolio.

\subsection{Neural network models} \label{sec_nnsetup}

For each test set $\mathcal{D}_{\ell}$, the pre-training of the autoencoder uses the Adam optimizer algorithm, a batch size of $1\,000$ and a randomly selected validation set of $20\%$ of the training set $\mathcal{D}\setminus\mathcal{D}_{\ell}$ for early stopping. The number of nodes $d$ in the encoding layer is tuned, testing across the values $\{5,10,15\}$, selecting the lowest value of $d$ while the cross-entropy loss $\mathscr{L}^{\text{CE}}(\cdot,\cdot) < 0.001$, as calculated with Equation \eqref{eq_bincrossentropy}. After the autoencoder is trained and scaled, the encoder is used in each FFNN and CANN structure, for both frequency and severity modeling. 

For both the FFNN and the CANN models, a random grid $\mathcal{R}$ of size $40$ is sampled from the search space $\mathcal{S}$ defined by the tuning parameters and their respective ranges as shown in Table \ref{tab_randomgrid}.

\setlength{\extrarowheight}{3pt} % a bit of extra white space between lines in tables
\begin{table}[H]
\centering
%\scriptsize
\begin{NiceTabular}{wl{7cm}wl{4cm}wr{1.5cm}}[
code-before = \rowcolor[HTML]{FFFFFF}{1,2,5}
              \rowcolor[HTML]{FAFAFF}{3,4,6}
]
\toprule
\textbf{Tuning parameter} & \textbf{Range} & \\
\noalign{\hrule height 0.3pt}
Activation function for hidden layers & ReLU, sigmoid, softmax\footnotemark & \\
Batch size & $[10\,000,50\,000]$ & \scriptsize{Frequency} \\
& $[200,10\,000]$ & \scriptsize{Severity} \\
Number of hidden layers & $[\num{1},\num{4}]$ & \\
Nodes per hidden layer & $[\num{10},\num{50}]$ & \\
Dropout rate & $[\num{0},\num{0.1}]$ & \\
\bottomrule
\end{NiceTabular}
\caption{Collection of tuning parameters and their respective ranges for the random grid search tuning strategy. This range is used for both the FFNN and CANN structures.}
\label{tab_randomgrid}
\end{table}

\footnotetext{The softmax activation function is applied on the collection of all nodes in a hidden layer, rather than individually at each node within the layer.}

The cross-validation scheme is shown in Algorithm \ref{cv_algo}, starting with the pre-processing steps, and resulting in out-of-sample performances for each holdout test set $\mathcal{D}_1,\ldots,\mathcal{D}_6$. For $\ell=1,\ldots,6$, we train a network on the data $\mathcal{D}\backslash\mathcal{D}_{\ell}$, choosing a random validation set consisting of $20\%$ of the training data for early stopping. With this model, we construct out-of-sample predictions on the test set $\mathcal{D}_{\ell}$ and calculate the out-of-sample loss using the loss functions in Equation \eqref{eq_devpoisson} and \eqref{eq_devgamma}. Because optimization in a neural network is dependent on the random initialization of the weights, we train each model three times and use the average out-of-sample loss over the three training runs. This ensures an objective out-of-sample loss evaluation without the risk of accidentally getting well or badly initialized weights.

%\begin{adjustwidth}{-1.6cm}{-1.6cm}
\begin{algorithm}[!ht]
    \setstretch{1}
	\SetKwInput{Input}{Input}
	\SetKwInput{Output}{Output}
	\noindent\rule{1.035\linewidth}{0.4pt} \\
	\Input{model class (\texttt{mclass}) and corresponding tuning grid $\mathcal{R}$}
	data $\mathcal{D}$ with $6$ disjoint stratified subsets $\mathcal{D}_1,\ldots,\mathcal{D}_6$\;
	\For{$\ell=1,\ldots, 6$}{
		leave out $\mathcal{D}_{\ell}$ as test set\;
        \begin{rightbrace}{pre-processing steps}
        \ForEach{ \emph{continuous variable} $x_j\in\mathcal{D}$}{
         calculate mean $\mu_{x_j}$ and standard deviation $\sigma_{x_j}$ on the data $\mathcal{D} \setminus \mathcal{D}_\ell$\;
         normalize the variable $x_j$ in the data set $\mathcal{D}$ along Equation \eqref{eq_normalization}\;
        }
        \ForEach{ \emph{categorical variable} $x_j\in\mathcal{D}$}{
         construct one-hot encoding $\left(x^{(j)}_1,\ldots,x^{(j)}_{L_j}\right)$ of variable $x_j$\;
        }
        \For{$d\in\{5,10,15\}$}{
         train an autoencoder $f^{\text{AE}}_{d}$ using the one-hot encoding of all categorical variables in $\mathcal{D} \setminus \mathcal{D}_\ell$ as input and $d$ nodes in the encoding layer, using a randomly selected validation set for early stopping\;
         evaluate the model performance on $\mathcal{D} \setminus \mathcal{D}_\ell$ using loss function $\mathscr{L}^{\text{CE}}(\cdot,\cdot)$\;
        }
        select $f^{\text{AE}}_{d}$ with the lowest $d$ while $\mathscr{L}^{\text{CE}}(\cdot,\cdot) < 0.001$\;
        calculate the scaled $\tilde{W}_{\text{enc}}$ and bias vector $\tilde{\boldsymbol{b}}_{\text{enc}}$ for the encoder part of $f^{\text{AE}}_{d}$\;
        add the encoder part of $f^{\text{AE}}_{d}$, with $\tilde{W}_{\text{enc}}$ and $\tilde{\boldsymbol{b}}_{\text{enc}}$, to the model class \texttt{mclass}\;
        \end{rightbrace}
        \vspace{-4mm}
        \begin{rightbrace}{cross-validation steps}
		\ForEach{ \emph{tuning parameter point} $s\in\mathcal{R}$}{
			\For{$k \in \left\lbrace 1,\ldots,6\right\rbrace  \setminus \ell$}{
				train a model $f_{\ell k}$ of \texttt{mclass} on $\mathcal{D} \setminus \left\lbrace \mathcal{D}_{\ell},\mathcal{D}_k \right\rbrace $\;
				evaluate the model performance on $\mathcal{D}_k$ using loss function $\mathscr{L}(\cdot,\cdot)$\;
                $\text{valid\_error}_{\ell k} \leftarrow \frac{1}{|\mathcal{D}_k|}\underset{i:\boldsymbol{x}_i \in \mathcal{D}_k}{\sum} \mathscr{L}\{y_i,f_{\ell k}(\boldsymbol{x}_i)\}$\;
			}
			$\text{valid\_error}_\ell \leftarrow \frac{1}{5}\sum_{k \in \left\lbrace 1,\ldots,6\right\rbrace  \setminus \ell}\text{valid\_error}_{\ell k}$\;
		}
		optimal parameter point $s^{*}_{\ell}\in\mathcal{R}$ minimizes $\text{valid\_error}_\ell$\;
        \end{rightbrace}
        \vspace{-4mm}
        \begin{rightbrace}{out-of-sample performance}
        \For{\emph{rep} $\in\{1,2,3\}$}{
            select a validation set $\mathcal{D}_{\text{val}}$ containing $20\%$ of the records in $\mathcal{D} \setminus \mathcal{D}_\ell$\;
            train a model $f_{\ell,\text{rep}}$ of \texttt{mclass} on $\mathcal{D} \setminus \{ \mathcal{D}_\ell, \mathcal{D}_{\text{val}} \}$ using the optimal parameter point $s^{*}_{\ell}$ and using $\mathcal{D}_{\text{val}}$ for early stopping\;
    		evaluate the model performance on $\mathcal{D}_\ell$ using loss function $\mathscr{L}(\cdot,\cdot)$\;
    		$\text{test\_error}_{\ell,\text{rep}} \leftarrow \frac{1}{|\mathcal{D}_\ell|}\underset{i:\boldsymbol{x}_i \in \mathcal{D}_\ell}{\sum} \mathscr{L}\{y_i,f_{\ell,\text{rep}}(\boldsymbol{x}_i)\}$\;
        }
        $\text{test\_error}_{\ell} \leftarrow \frac{1}{3}\sum_{\text{rep}} \text{test\_error}_{\ell,\text{rep}}$\;
        \end{rightbrace}
        \vspace{-2mm}
	} 
    \vspace{2mm}
	\Output{optimal tuning parameters + performance measure for each of the six folds.}
	\vspace{-2mm} \noindent\rule{1.035\linewidth}{0.4pt}
  \caption{Pseudocode to sketch the pipeline for calculating out-of-sample performances with the neural network structures from Section \ref{sec_NNarch}, including the data pre-processing steps, the cross-validation scheme as outlined in \citet{Henckaerts2021} with the random grid search methodology and the repeated out-of-sample loss calculation to avoid local minima solutions.}
\label{cv_algo}
\end{algorithm}
%\end{adjustwidth}

\subsection{Comparison of categorical embedding methods} \label{sec_onehotcomparison}

We investigate the impact of the autoencoder embedding compared to directly utilizing one-hot encoded categorical variables. For each data set, we train a FFNN and a CANN model using the one-hot encoding of each categorical variable and a FFNN and a CANN model using the autoencoder embedding. For this investigation, we do not tune the models but choose a set of tuning parameters that we apply to all the models shown. This means the deviance of each model is not relevant here, only the difference in deviance between the model with one-hot encoding and the model with autoencoder embedding. This approach allows to isolate the embedding technique's effect on a model's predictive performance. Each model is trained on $\mathcal{D}\setminus\mathcal{D}_1$, and the out-of-sample performance is calculated on the out-of-sample test set $\mathcal{D}_1$. \added{We provide a comparison with limited tuning of parameters, but refer to \citet{delong2021} for a more in-depth comparison between the embedding methodologies, including the effect of tuning the model architectures.}

\begin{figure}[ht!]
\centering
  \includegraphics[width = \linewidth]{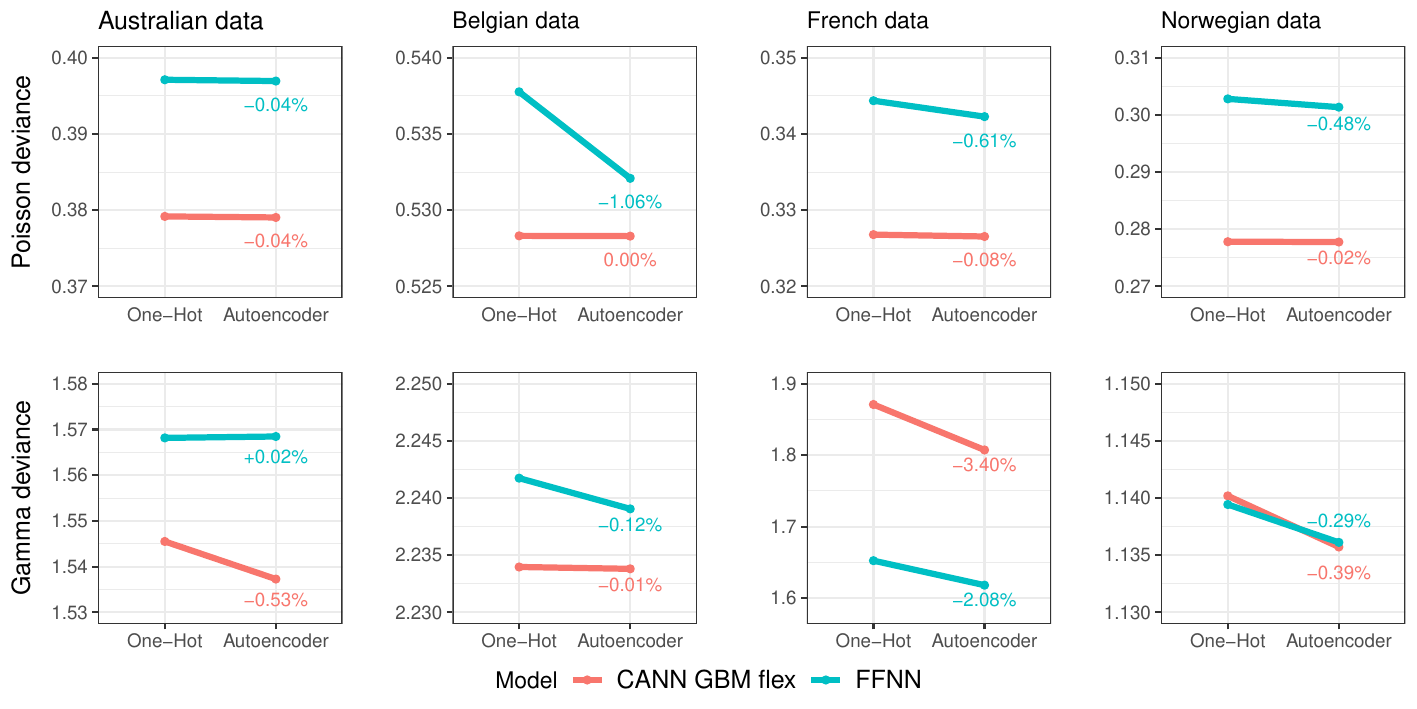}
\caption{Comparison of one-hot encoding and autoencoder embedding on the out-of-sample performance of both the FFNN and the CANN model. Top row shows the effect on frequency modeling and bottom row on severity modeling. The numerical relative difference between the model using the autoencoder and the model using one-hot encoding is given in the graph.}
\label{fig_OHAE_comp}
\end{figure}

Figure \ref{fig_OHAE_comp} displays the predictive accuracy, as captured with the deviance, of each model under consideration, with the frequency models in the top and the severity models in the bottom row. \added{The numerical relative difference between the autoencoder models and the one-hot encoding models is given in the graph.} For frequency modeling, the autoencoder embedding has the most pronounced effect on the performance of the FFNNs, leading to lower deviance compared to models utilizing one-hot encoding. However, the impact on CANN models appears to be negligible. In the case of severity modeling, both FFNNs and CANNs demonstrate an improved predictive performance when using the autoencoder embedding. Only the FFNN on the Australian data set and the CANN model on the Belgian data set perform similarly when comparing the one-hot encoding and the autoencoder embedding for claim severity modeling. The reduced deviance in most severity models highlights the benefits of unsupervised learning through the autoencoder approach.

\subsection{Model evaluation framework}\label{sec_evalframe}

The two benchmark models enable a comparison with the proposed neural network architectures. Moreover, they serve as initial model input for the CANN models. We investigate the predictive performance of seven models for each data set: GLM, GBM, neural network, the CANN with GLM input, with fixed and flexible output layer, and the CANN with GBM input, with both fixed and flexible output layer.

\added{Inspired by \citet{FisslerTobias2023MCaC} and \citet{GneitingTilmann2023Rdmf}, we evaluate and compare the performance of these models along four dimensions. We compare the out-of-sample performances of the models using the Poisson and gamma deviance. We employ the statistical Diebold-Mariano test to compare the predictive accuracy of the models to determine whether a lower out-of-sample deviance corresponds to a statistically more accurate model. We examine the prediction structure, which provides deeper insights into the model's behavior and potential biases. This includes assessing the calibration of predictions to ensure that predicted values align well with observed outcomes. Lastly, we use Murphy diagrams to verify if the model performance rankings based on the Poisson and gamma deviance hold across various scoring functions, ranking the models for predictive dominance. This set of model evaluation tools provides a nuanced view of model performance. We refer to \citet{FisslerTobias2023MCaC} for a more complete overview of different model evaluation techniques. With the exception of the comparison of out-of-sample deviances, we restrict this section to the frequency models. The comparison between the severity models can be found in Appendix \ref{app_sev_evalframe}.}

\paragraph{Comparison on out-of-sample deviances} \label{sec_oos}

\begin{figure}[ht!]
\centering
  \includegraphics[width = \linewidth]{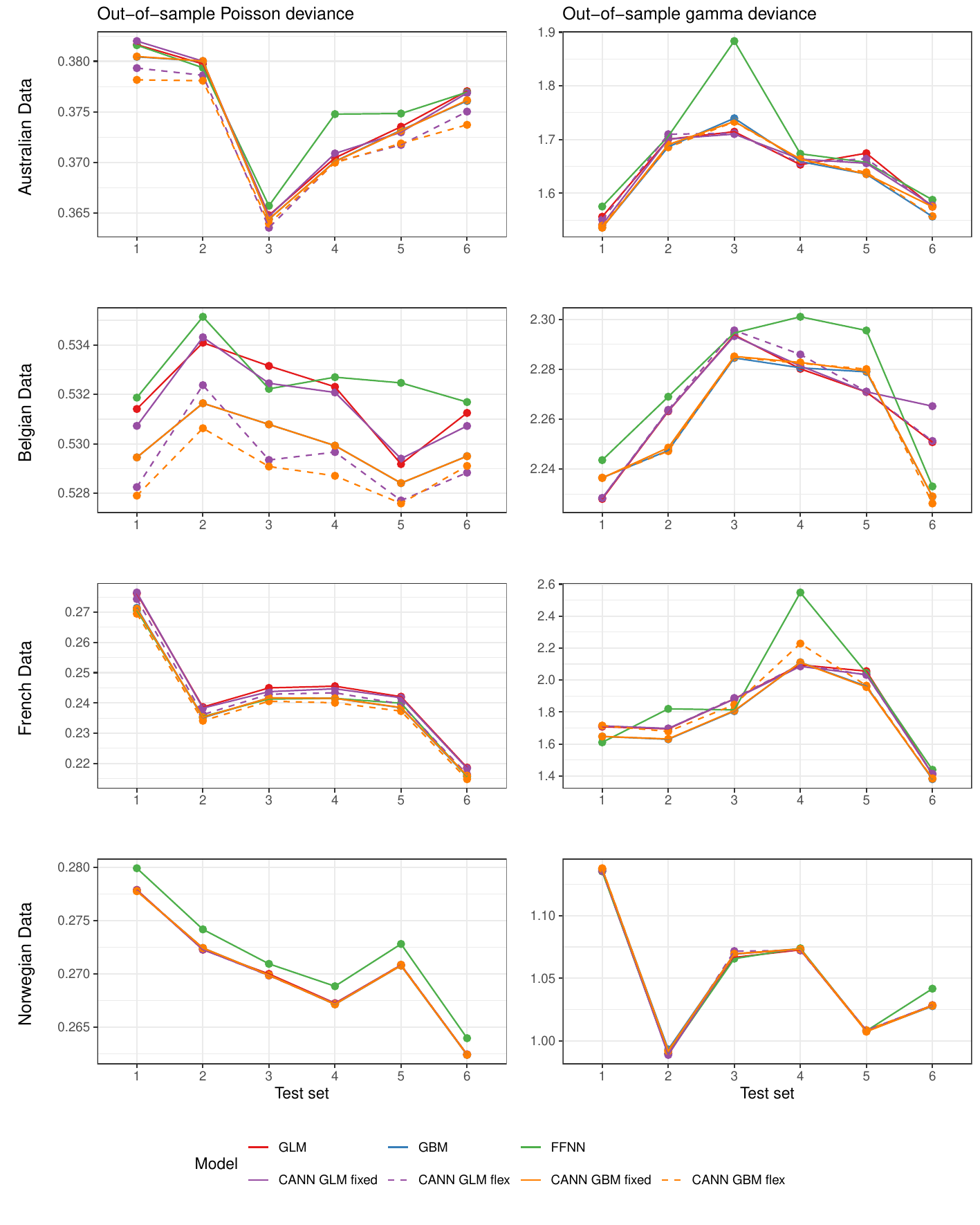}
\caption{Out-of-sample performance comparison between the different models for each data set. The left-hand side shows the performance of the frequency models and the right-hand side for the severity models. From top to bottom, we show the results on the Australian, Belgian, French and Norwegian data sets. The deviances for the GLM and GBM on the Belgian data correspond to the results reported in, respectively, \citet{Henckaerts2018} and \cite{Henckaerts2021}.}
\label{fig_oos_freq}
\end{figure}

We first compare out-of-sample performances of the benchmark models and the neural network structures by looking at the Poisson deviance \eqref{eq_devpoisson} or gamma deviance \eqref{eq_devgamma} for each withheld test set. Figure \ref{fig_oos_freq} shows the deviance for the frequency models on the left-hand side and the severity models on the right-hand side. \added{An accompanying table with the numerical values of these deviances is in Appendix \ref{app_oostable}.}

Among the four data sets analyzed, the combination of a neural network and a gradient boosting model (CANN GBM flex) consistently yields the lowest deviance when modeling claim frequency. This aligns with recent research highlighting the predictive performance of combining a gradient boosting model with a neural network \citep{Borisov2021,ke2019deepgbm,Shwartz-Ziv2021}. However, for the Norwegian data set, which has few input variables, the impact was less pronounced, with similar performance observed across all models except for the feed-forward neural network, which exhibits slightly higher deviance. Regarding claim severity modeling, no single model consistently achieves the lowest deviance across all data sets. For the Australian and Norwegian data sets, all models perform comparably in terms of deviance. The CANN models with GBM input demonstrate the lowest deviance for the Belgian data set, while for the French data set, the CANN model with GLM input achieves the best results. Notably, the CANN models with a flexible output layer structure outperform those with a fixed output layer in most cases, for both frequency and severity modeling. This suggests that the more adaptable combination of the initial model input and the neural network adjustment leads to reduced deviance.

\paragraph{Diebold-Mariano statistical tests for predictive accuracy}

\added{In addition to examining the out-of-sample deviances, we use the statistical Diebold-Mariano test to determine whether a model with a lower deviance demonstrates greater predictive accuracy. We say two models $f_A$ and $f_B$ have the same predictive accuracy under loss function $\mathscr{L}(\cdot,\cdot)$ if \newline $\E\left[\mathscr{L}(f_A(\X),Y) - \mathscr{L}(f_B(\X),Y)\right] = 0$. We test this hypothesis using a t-test, known as the Diebold–Mariano test \citep{dieboldmariano, diebold2015comparing}. We follow the approach in \citet{FisslerTobias2023MCaC}. The equal predictive null hypothesis of the Diebold-Mariano test is $$H_0: \E\left[\mathscr{L}(f_A(\X),Y) - \mathscr{L}(f_B(\X),Y)\right] = 0.$$ As alternative hypothesis, we use $$H_1: \E\left[\mathscr{L}(f_A(\X),Y) - \mathscr{L}(f_B(\X),Y)\right] > 0$$ to test for the predictive superiority of model B over model A. We reject the null hypothesis if our test statistic has a $p$-value $<0.05$. Note that failing to reject the null hypothesis does not exclude the possibility that model A is more accurate than model B, as we only used the alternative hypothesis $H_1$ that model B is statistically more accurate than model A.}

\added{Table \ref{tab_DB} shows the result of the Diebold-Mariano test for the frequency models on each data set under consideration. We look at the results on the test set $\mathcal{D}_1$ to isolate the effects of the models themselves and to avoid influences from variations across different test sets. Each row represents model A, and each column represents model B. A checkmark (\cmark) is used if the null hypothesis is not rejected and an \xmark\, if $H_0$ is rejected in favor of the alternative hypothesis. A column full of \xmark\,marks indicates that we always reject the null hypothesis when using this model as model B, meaning this model is consistently seen as statistically more accurate compared to the others. Conversely, a row full of \xmark\,marks means that when comparing to this model, the null hypothesis is always rejected in favor of the other models. Rows or columns with exclusively \xmark\,marks are highlighted. In the first row of Table \ref{tab_DB_BE} for the Belgian data, we see that according to the Diebold-Mariano test, the GLM has equal predictive performance as the neural network and the CANN GLM fixed, although the out-of-sample deviance of the CANN GLM fixed is lower on $\mathcal{D}_1$ as seen in Figure \ref{fig_oos_freq}. Looking at the CANN GBM flexible as model A, we can never reject $H_0$ in favor of another model; using the other models as model A, the test rejects $H_0$ in favor of the CANN GBM flexible each time, except for the CANN GLM flexible. This shows that the CANN GBM flexible, which has the lowest deviance on the test set, is statistically more accurate than the other model types, except for the CANN GLM flexible, where we cannot reject $H_0$. For the Australian data, Table \ref{tab_DB_AUS}, and the French data, Table \ref{tab_DB_FR}, the CANN GBM flexible is always seen as more statistically accurate when compared to the other models. Table \ref{tab_DB_NOR} shows the results on the Norwegian data, where the test only rejects $H_0$ when looking at the neural network as model A. This is in line with the out-of-sample performance in Figure \ref{fig_oos_freq}, where we can see that each frequency model on the Norwegian data has very similar deviance, except for the neural network.
}

\setlength{\extrarowheight}{3pt} % a bit of extra white space between lines in tables
\begin{table}[ht!]
 \begin{adjustwidth}{-0.6cm}{-0.6cm}
  \centering
  \scriptsize
  \begin{subtable}[t]{.49\linewidth}
    \begin{NiceTabular}{cr*{7}{r}}[
    code-before = \rowcolor[HTML]{FFFFFF}{1,2,4,6,8}
              \rowcolor[HTML]{FAFAFF}{3,5,7,9}
    ]
    \toprule
    & & \Block{1-7}{\footnotesize \textbf{Model B}} \\[1mm]
    \RowStyle{\rotate}
    & & GLM & GBM & FFNN & CANN GLM fixed & CANN GLM flex & CANN GBM fixed & CANN GBM flex\\
    \noalign{\hrule height 0.3pt}
    \parbox[t]{2mm}{\multirow{7}{*}{\rotatebox[origin=c]{90}{\footnotesize \textbf{Model A}}}}
    & GLM &  & \xmark & \cmark & \cmark & \xmark & \xmark & \cellcolor{blue!25}\xmark \\ 
    & GBM & \cmark & & \cmark & \cmark & \cmark & \cmark &  \cellcolor{blue!25}\xmark \\ 
    & FFNN & \cmark & \xmark &  & \cmark & \xmark & \xmark & \cellcolor{blue!25}\xmark \\ 
    & CANN GLM fixed & \cmark & \xmark & \cmark & & \xmark & \xmark & \cellcolor{blue!25}\xmark \\ 
    & CANN GLM flex & \cmark & \cmark & \cmark & \cmark &  & \cmark & \cellcolor{blue!25}\xmark \\ 
    & CANN GBM fixed & \cmark & \cmark & \cmark & \cmark & \cmark &  & \cellcolor{blue!25}\xmark \\ 
    & CANN GBM flex & \cmark & \cmark & \cmark & \cmark & \cmark & \cmark & \\ 
    \bottomrule
    \end{NiceTabular}
  \caption{Australian data set}
  \label{tab_DB_AUS}
  \end{subtable} 
  \hfill
  \begin{subtable}[t]{.49\linewidth}
    \begin{NiceTabular}{cr*{7}{r}}[
    code-before = \rowcolor[HTML]{FFFFFF}{1,2,4,6,8}
              \rowcolor[HTML]{FAFAFF}{3,5,7,9}
    ]
    \toprule
    & & \Block{1-7}{\footnotesize \textbf{Model B}} \\[1mm]
    \RowStyle{\rotate}
    & & GLM & GBM & FFNN & CANN GLM fixed & CANN GLM flex & CANN GBM fixed & CANN GBM flex\\
    \noalign{\hrule height 0.3pt}
    \parbox[t]{2mm}{\multirow{7}{*}{\rotatebox[origin=c]{90}{\footnotesize \textbf{Model A}}}}
    & GLM &  & \xmark & \cmark & \cmark & \xmark & \xmark & \xmark \\ 
    & GBM & \cmark & & \cmark & \cmark & \xmark & \cmark &  \xmark \\ 
    & FFNN & \cmark & \xmark &  & \cmark & \xmark & \xmark & \xmark \\ 
    & CANN GLM fixed & \cmark & \xmark & \cmark & & \xmark & \xmark & \xmark \\ 
    & CANN GLM flex & \cmark & \cmark & \cmark & \cmark &  & \cmark & \cmark \\ 
    & CANN GBM fixed & \cmark & \cmark & \cmark & \cmark & \xmark &  & \xmark \\ 
    & CANN GBM flex & \cmark & \cmark & \cmark & \cmark & \cmark & \cmark & \\ 
    \bottomrule
    \end{NiceTabular}
  \caption{Belgian data set}
  \label{tab_DB_BE}
  \end{subtable} 
\newline
\vspace*{1 cm}
\newline
  \begin{subtable}[t]{.49\linewidth}
    \begin{NiceTabular}{cr*{7}{r}}[
    code-before = \rowcolor[HTML]{FFFFFF}{1,2,4,6,8}
              \rowcolor[HTML]{FAFAFF}{3,5,7,9}
    ]
    \toprule
    & & \Block{1-7}{\footnotesize \textbf{Model B}} \\[1mm]
    \RowStyle{\rotate}
    & & GLM & GBM & FFNN & CANN GLM fixed & CANN GLM flex & CANN GBM fixed & CANN GBM flex\\
    \noalign{\hrule height 0.3pt}
    \parbox[t]{2mm}{\multirow{7}{*}{\rotatebox[origin=c]{90}{\footnotesize \textbf{Model A}}}}
    & GLM &  & \xmark & \xmark & \cmark & \xmark & \xmark & \cellcolor{blue!25}\xmark \\ 
    & GBM & \cmark & & \cmark & \cmark & \cmark & \xmark &  \cellcolor{blue!25}\xmark \\ 
    & FFNN & \cmark & \cmark &  & \cmark & \cmark & \cmark & \cellcolor{blue!25}\xmark \\ 
    & CANN GLM fixed & \cmark & \xmark & \xmark & & \xmark & \xmark & \cellcolor{blue!25}\xmark \\ 
    & CANN GLM flex & \cmark & \xmark & \xmark & \cmark &  & \xmark & \cellcolor{blue!25}\xmark \\ 
    & CANN GBM fixed & \cmark & \cmark & \cmark & \cmark & \cmark &  & \cellcolor{blue!25}\xmark \\ 
    & CANN GBM flex & \cmark & \cmark & \cmark & \cmark & \cmark & \cmark & \\ 
    \bottomrule
    \end{NiceTabular}
  \caption{French data set}
  \label{tab_DB_FR}
  \end{subtable} 
  \hfill
  \begin{subtable}[t]{.49\linewidth}
    \begin{NiceTabular}{cr*{7}{r}}[
    code-before = \rowcolor[HTML]{FFFFFF}{1,2,4,6,8}
              \rowcolor[HTML]{FAFAFF}{3,5,7,9}
    ]
    \toprule
    & & \Block{1-7}{\footnotesize \textbf{Model B}} \\[1mm]
    \RowStyle{\rotate}
    & & GLM & GBM & FFNN & CANN GLM fixed & CANN GLM flex & CANN GBM fixed & CANN GBM flex\\
    \noalign{\hrule height 0.3pt}
    \parbox[t]{2mm}{\multirow{7}{*}{\rotatebox[origin=c]{90}{\footnotesize \textbf{Model A}}}}
    & GLM &  & \cmark & \cmark & \cmark & \cmark & \cmark & \cmark \\ 
    & GBM & \cmark & & \cmark & \cmark & \cmark & \cmark &  \cmark \\ 
    & FFNN & \cellcolor{blue!25}\xmark & \cellcolor{blue!25}\xmark &  & \cellcolor{blue!25}\xmark & \cellcolor{blue!25}\xmark & \cellcolor{blue!25}\xmark & \cellcolor{blue!25}\xmark \\ 
    & CANN GLM fixed & \cmark & \cmark & \cmark & & \cmark & \cmark & \cmark \\ 
    & CANN GLM flex & \cmark & \cmark & \cmark & \cmark &  & \cmark & \cmark \\ 
    & CANN GBM fixed & \cmark & \cmark & \cmark & \cmark & \cmark &  & \cmark \\ 
    & CANN GBM flex & \cmark & \cmark & \cmark & \cmark & \cmark & \cmark & \\ 
    \bottomrule
    \end{NiceTabular}
  \caption{Norwegian data set}
  \label{tab_DB_NOR}
  \end{subtable} 
  \end{adjustwidth}
 \caption{Results of the Diebold-Mariano test for the frequency models on test set $\mathcal{D}_1$ for each data set. The table indicates whether we cannot reject (\cmark) the null hypothesis $H_0: \E\left[\mathscr{L}(f_A(\X),Y) - \mathscr{L}(f_B(\X),Y)\right] = 0$, or if we reject $H_0$ (\xmark) in favor of the alternative hypothesis $H_1: \E\left[\mathscr{L}(f_A(\X),Y) - \mathscr{L}(f_B(\X),Y)\right] > 0$. Highlighted cells indicate columns or rows filled exclusively with \xmark\,marks, signifying that the model is either consistently seen as statistically more accurate (column) or that, for this model, the null hypothesis is always rejected in favor of the other models (row).}
  \label{tab_DB}
\end{table}

\paragraph{Prediction structure analysis and calibration}

\begin{figure}[ht!]
\begin{adjustwidth}{-1.6cm}{-1.2cm}
\centering
  \includegraphics[width = \linewidth]{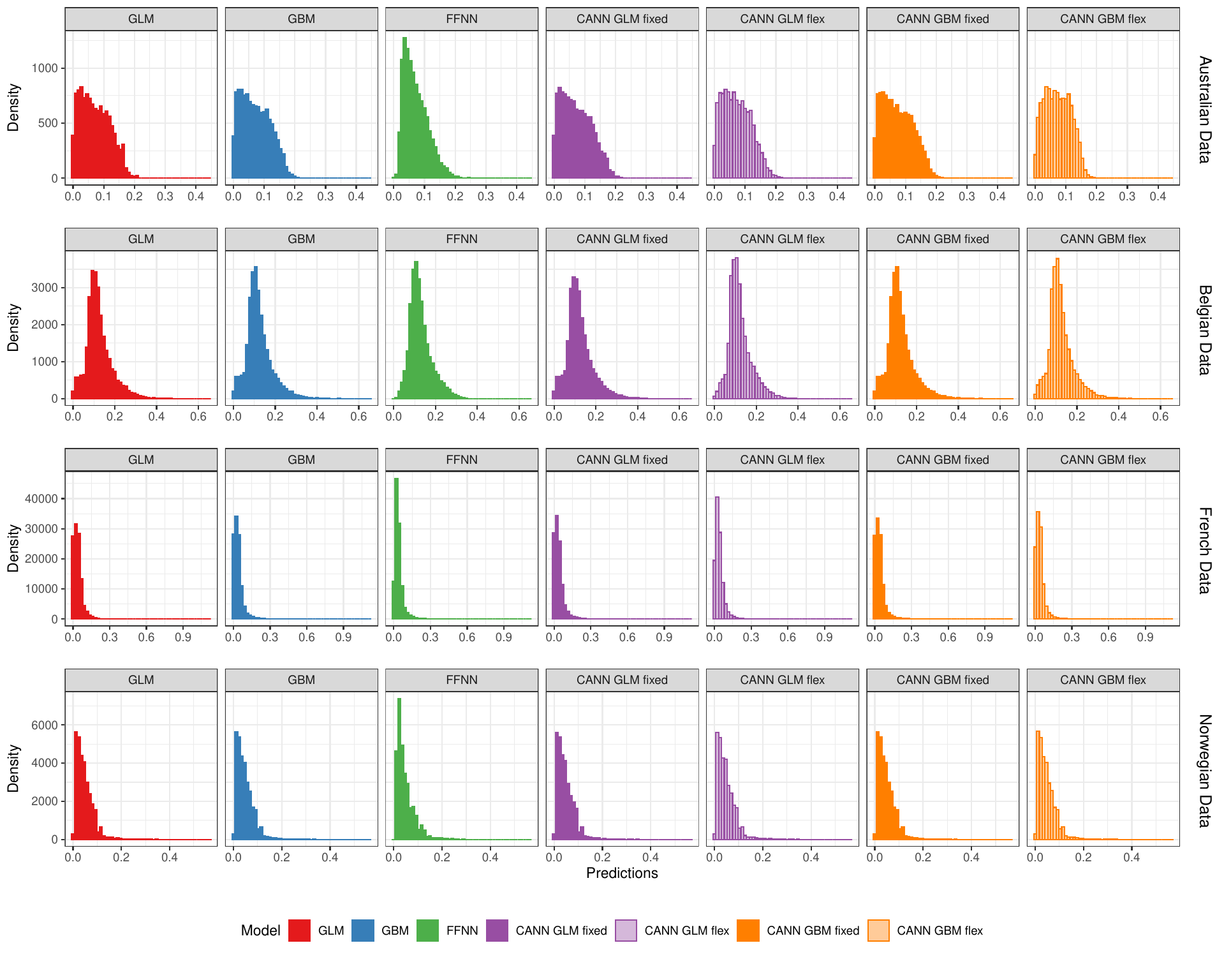}
\end{adjustwidth}
\caption{Histogram of predictions for the test set $\mathcal{D}_1$ obtained from each frequency model, illustrating the distribution and concentration of predictions for each model to identify differences in prediction patterns and potential outliers. }
\label{fig_predhisto_comp}
\end{figure}

\added{Following \citet{DENUIT2021485}, we examine the prediction structures to gain deeper insights into the model's behavior to facilitate comparisons and to identify outliers. Figure \ref{fig_predhisto_comp} shows the histograms of the predictions made on the out-of-sample test set $\mathcal{D}_1$ of each frequency model. Such histograms facilitate a visual inspection of the predictive distributions and their alignment across different models, for instance, with respect to dispersion and outliers. For the Australian, French, and Norwegian data, we observe that the dispersion of predictions is very similar between models, except for the neural network (in green), where the predictions are more concentrated. This behavior is not observed in the Belgian data, where for each model, the predictive distribution shows a similar dispersion.}

\begin{figure}[ht!]
\begin{adjustwidth}{-1.6cm}{-1.2cm}
\centering
  \includegraphics[width = \linewidth]{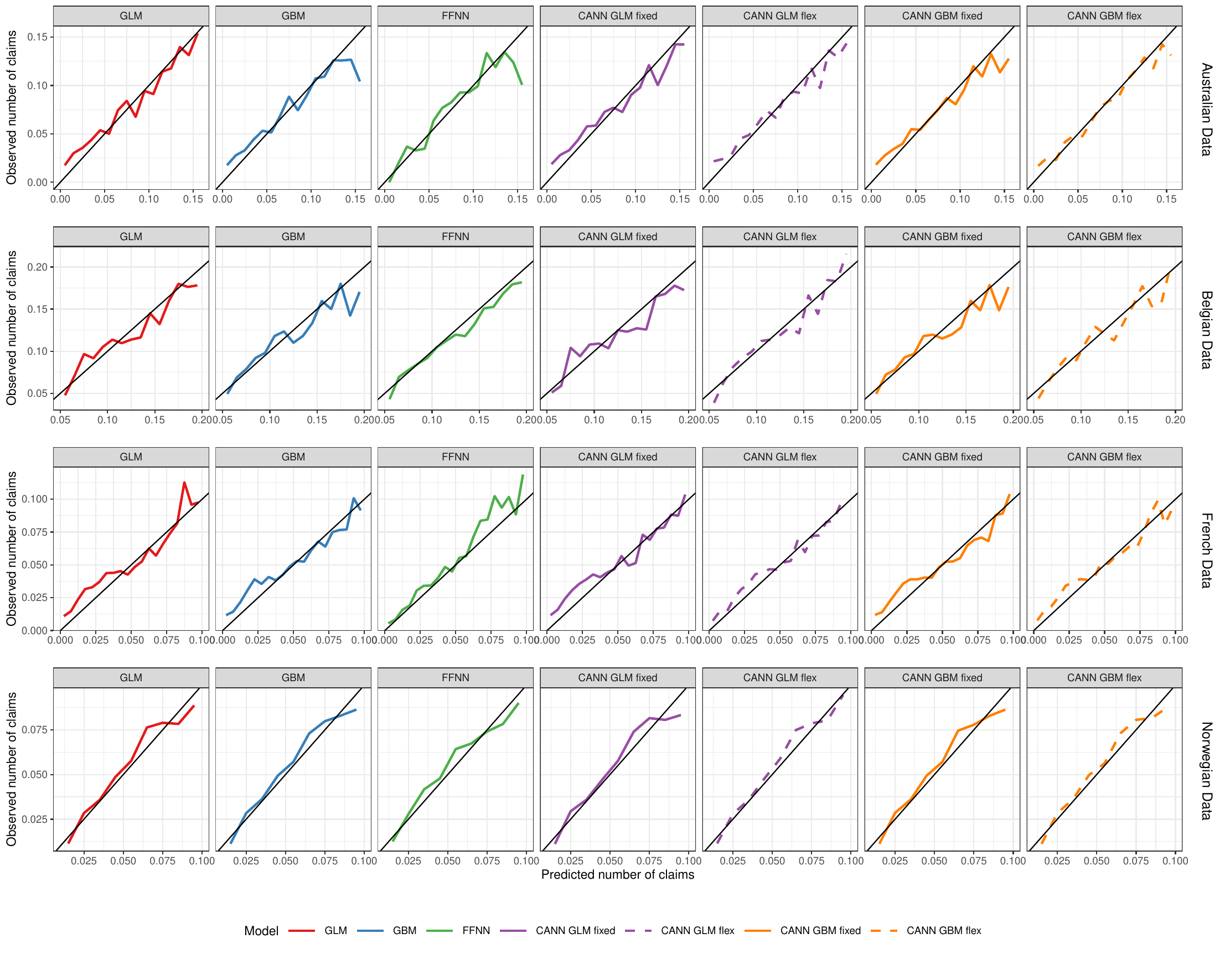}
\end{adjustwidth}
\caption{Plot of $\E \left[Y | f(\X) = s \right]$ over the binned range of predictions $s$ for each model type. Predictions are made on test set $\mathcal{D}_1$. A line above (below) the diagonal indicates that the model is underestimating (overestimating) the true number of claims in the data. Comparison between models allows us to assess how well the predicted values align with observed outcomes.}
\label{fig_exp_response}
\end{figure}

\added{In Figure \ref{fig_exp_response} we examine the calibration of the different frequency models by plotting $\E \left[Y | f(\x) = s \right]$ over the binned range of possible predictions $s$, following the approach in \citet{DENUIT2021485}. We detail the construction of these graphs in Appendix \ref{app_calibplot}. When $\E \left[Y | f(\x) = s \right] > s$, meaning the plot lies above the diagonal, the model underestimates the number of claims. Conversely, when $\E \left[Y | f(\x) = s \right] < s$, the model overestimates the number of claims. A model where $\E \left[Y | f(\x) = s \right] \sim s$ for all $s$ is considered well-calibrated, meaning the predictions accurately match the observed number of claims. Comparing the calibration between models can help to identify biases in the predictions that might not be evident from a comparison based on out-of-sample deviance. Both benchmark models seem similarly well calibrated, except for the slight overestimation for the higher prediction ranges in the Australian and Belgian data sets. With the exception of the Belgian data, the NN shows a worse calibration than the benchmark models. The calibration of the CANN GLM fixed and CANN GBM fixed is comparable to that of the GLM and GBM, respectively, but with an improved calibration for the high prediction range in the French data set. For all data sets, the CANN GLM flexible and CANN GBM flexible demonstrate the best calibration, exhibiting a better calibration than their counterparts with a fixed output layer.}

\paragraph{Murphy diagrams for predictive dominance}

\added{Following the methodology outlined in \citet{FisslerTobias2023MCaC}, we employ Murphy diagrams \citep{ehm2016quantiles} to evaluate whether the model rankings based on the out-of-sample deviances shown in Figure \ref{fig_oos_freq} are influenced by the choice of loss function. While Figure \ref{fig_oos_freq} indicates which model performs best according to the Poisson or gamma deviance, Murphy diagrams allow testing whether this ranking holds true across a variety of other scoring functions. By doing so, we confirm whether the observed performance differences are consistent and not merely artifacts of using a specific loss function. The Murphy diagram for a model $f$ displays $\left(\theta,S_\theta(f(\x),y)\right)$, for all $(\x_i,y_i)$ in the test set with $i=1,\ldots,n$, where $S_\theta(f(\x),y)$ is the elementary scoring function for parameter $\theta$, defined by \citet{ehm2016quantiles} as }
\[S_\theta(f(\x),y) = \frac{1}{n}\sum_{i=1}^{n}|\theta - y_i|\mathbbm{1}\{\min\left(f(\x_i),y_i\right)\leq\theta<\max\left(f(\x_i),y_i\right)\}.\]
\added{The predictions from model $f_A$ are said to dominate the predictions of model $f_B$, if and only if $S_\theta(f_A(\x),y) \leq S_\theta(f_B(\x),y)$ for all $\theta\in\mathbb{R}$, see \citet[Section 3.4]{ehm2016quantiles}. By examining the rankings between models according to the loss $S_\theta$ for a range of $\theta$ values, we can determine whether a model with a lower Poisson or gamma deviance consistently maintains this lower loss across all scoring functions $S_\theta$. Furthermore, if a model shows predictive dominance over another model as determined by the elementary scoring functions for all $\theta$, then the ranking of these models will be preserved across all consistent loss functions, see \citet[Proposition 2b]{ehm2016quantiles}.}

\begin{figure}[ht!]
 \begin{adjustwidth}{-1.4cm}{-1.4cm}
 \centering
 \includegraphics[width = 1\linewidth]{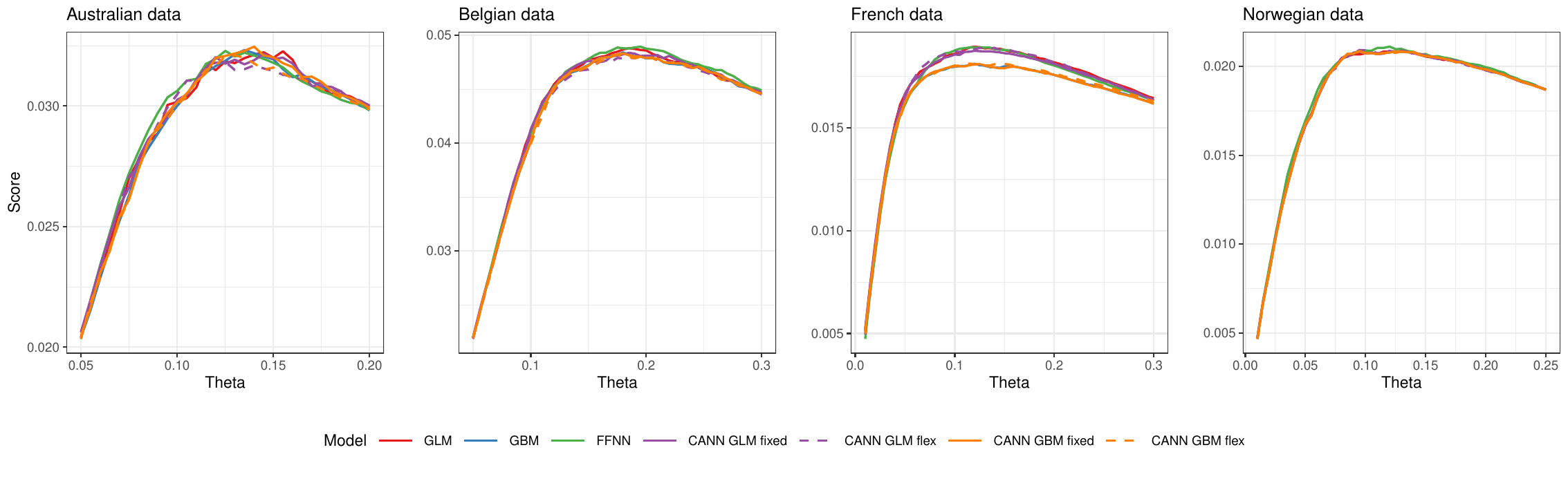}
 \end{adjustwidth}
\caption{Murphy diagram for the frequency models on test set $\mathcal{D}_1$ for each data set to compare the ranking of models across a range of scoring functions. We plot the losses for each model according to the elementary scoring functions $S_\theta$ over a range of values for the parameter $\theta$.}
\label{fig_murphy}
\end{figure}

\added{ Figure \ref{fig_murphy} shows the Murphy diagrams for the frequency models calibrated on the different data sets with the loss evaluated on the test set $\mathcal{D}_1$. For the French data, we see that the GBM and the CANN GBM fixed and flexible models have a lower value $S_\theta$ for all $\theta$, meaning they have predictive dominance over the GLM, NN, and CANN GLM models. Between these three models, however, none is consistently lower than the others, so we cannot say that one model has predictive dominance over all others according to the Murphy diagram. For the Norwegian data set, all models, except the neural network, have almost identical values of $S_\theta$ over all $\theta$. The values for the neural network are, however, consistently higher. Therefore, we conclude that all model types have predictive dominance over the neural network. The Murphy diagrams for the Australian and Belgian data show no model with clear predictive dominance. }

\paragraph{Model evaluation framework conclusions}

\added{We can draw comprehensive and well-supported conclusions about model performance and reliability for all four data sets by consolidating insights from the different techniques in the model evaluation framework. For the frequency models on the Australian data set, we see the CANN GBM flexible leads to the lowest out-of-sample Poisson deviance. The model is statistically more accurate as determined by the Diebold-Mariano test, although the Murphy diagram does not show predictive dominance for all values of $\theta$. Furthermore, the histogram of the predictions made by the CANN GBM flexible does not show unexpected outliers, and the model is well calibrated, leading to the conclusion that the CANN GBM flexible model is the most accurate for modeling claim frequency on the Australian data set. We draw the same conclusion for the French data set, where the Murphy diagram shows that the CANN GBM flexible has predictive dominance compared to the GLM, the FFNN, and the CANN GLM models. On the Belgian data set, the CANN GBM flexible also leads to the lowest out-of-sample Poisson deviance, but according to the Diebold-Mariano test, its accuracy is equal to that of the CANN GLM flexible. Neither model shows predictive dominance according to the Murphy diagram, and both are well-calibrated without unexpected outliers, showing both models are suitable choices when modeling claim frequency on the Belgian data set. The frequency models on the Norwegian data set have similar out-of-sample deviances, except for the neural network, for which the Diebold-Mariano test always rejects the null hypothesis when using the neural network as model A. The Murphy diagram shows no model has predictive dominance over the other models, although all models have (slightly) lower values for $S_\theta$ for all $\theta$, meaning they have predictive dominance over the neural network. To model claim frequency on the Norwegian data set, we conclude that the NN is the least suitable model, while other models perform comparably. Therefore, the more intuitive and explainable GLM seems sufficient.}

\section{Looking under the hood: interpretation tools and surrogate models} \label{sec_interpret}

We now consider two model interpretation tools to look under the hood of the constructed models. Then, we translate the model insights into a tariff structure by constructing GLM surrogates along the workflow presented in \citet{Henckaerts2022}. All results shown in this section are calculated using data fold one, meaning the models are trained using data subsets $\mathcal{D}_2,\ldots,\mathcal{D}_6$ and the shown results are calculated on the test set $\mathcal{D}_1$. \added{We show the results from these tools only for the GBM, FFNN, and CANN GBM flexible models. This illustrates how the interpretation techniques work and how we can extract insights from them while keeping the results easily readable.}

\subsection{Variable importance}

\begin{figure}[ht!]
\begin{adjustwidth}{-1.6cm}{-1.2cm}
\centering
  \includegraphics[width = \linewidth]{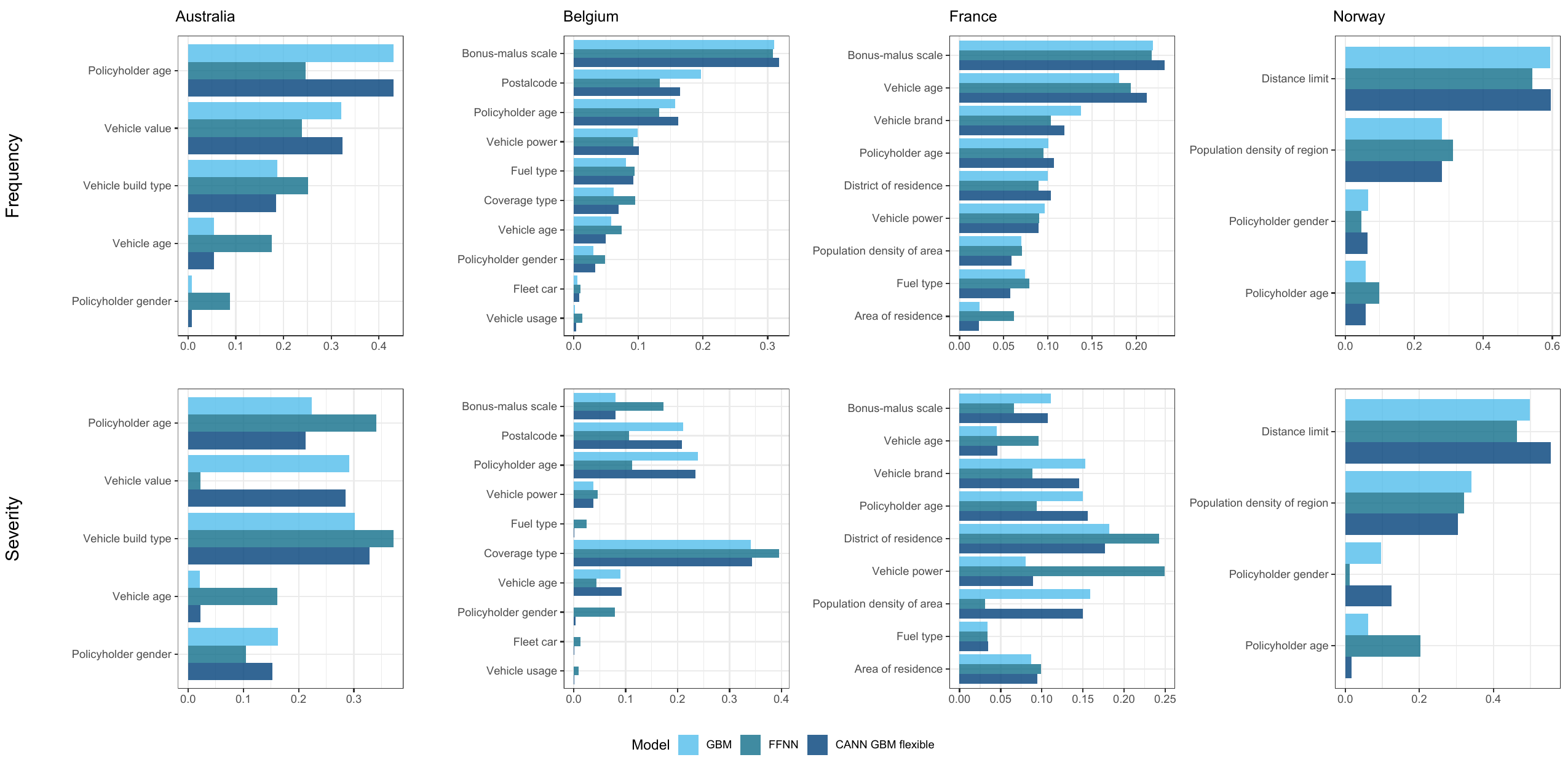}
\end{adjustwidth}
\caption{Relative variable importance in the GBM, the FFNN and the CANN GBM flexible. Top row shows the effects for the frequency models, the bottom row for the severity models.}
\label{fig_vip}
\end{figure}

We measure variable importance using the permutation method from \citet{Olden2004}. Hereby, we consider the average change in predictions when a variable is randomly permuted. For a trained model $f$, we measure the importance of a variable $x_j$ by calculating

\begin{equation}
    \text{VIP}_{x_j} = \sum_{i:\x_i\in\D} \text{abs}\big( f\left(x_{i,1},\ldots,x_{i,j},\ldots,x_{i,p}\right) - f\left(x_{i,1},\ldots,\tilde{x}_{i,j},\ldots,x_{i,p}\right)\big),
\end{equation}

where $\tilde{x}_{i,j}$ is a random permutation of the values observed for $x_j$ in the data set $\D$. A large value for $\text{VIP}_{x_j}$ indicates that the variable significantly influences the model output and is therefore considered important. Figure \ref{fig_vip} shows the variable importance of each variable in the four data sets for both frequency and severity modeling. For clarity, we show the relative VIP of each variable, calculated as 

\begin{equation}
    \overline{\text{VIP}}_{x_j} = \frac{\text{VIP}_{x_j}}{\sum_{x_j\in\mathcal{D}} \text{VIP}_{x_j}},\qquad\text{where the sum runs over all over all variables}\,\,x_j.
    \label{eq_vip}
\end{equation}

By comparing the variable importance of the GBM with the CANN model, we can evaluate the impact of the neural network adjustment component within the CANN. In general, most variables show similar importance in both the GBM and the CANN GBM flexible models, indicating that the adjustment calibrated by the neural network does not substantially alter the importance of the relationships between the input variables and the response variable. However, notable changes are observed for certain variables, such as the postal code in the frequency model for the Belgian data set and the vehicle age and brand in the frequency model for the French data set. When we compare the variable importance of the GBM and the CANN GBM with the FFNN, we observe more substantial changes, particularly in claim severity modeling. This shows that the FFNN captures a significantly different relationship between the input variables and the response variable when compared to the GBM and CANN GBM flexible model.  

\subsection{Partial dependence effects}\label{sec_pdp}

We consider partial dependence effects \citep{Freidmanetal2001, Henckaerts2021} to explore the relationship between an input variable and the model output. Let the variable space $X_j$ be the vector containing all possible values for variable $x_j$. In case the variable $x_j$ is continuous, we discretize $X_j$ by dividing the range between the smallest and largest possible values for $x_j$ into intervals using the smallest step size that occurs in the data. For a trained model $f$, the partial dependency effect of a variable $x_j$ is the vector calculated as

\begin{equation}
    \text{PD}_{x_j} = \left\{\frac{1}{|\mathcal{D}|}\sum_{\x_i\in\mathcal{D}}f\left(x_{i,1},\ldots,X_{o,j},\ldots,x_{i,p}\right)\,;\, \forall X_{o,j} \in X_j\right\},
    \label{eq_pdp}
\end{equation}

where $\left(x_{i,1},\ldots,X_{o,j},\ldots,x_{i,p}\right)$ is the data point $\x_i\in\mathcal{D}$ with element $x_{i,j}$ replaced by the value $X_{o,j}\in X_j$. The vector $\text{PD}_{x_j}$ can be seen as the average prediction on $\mathcal{D}$, while letting the variable $x_j$ range over all possible values in the variable space $X_j$. A partial dependence plot is the plotted effect between $X_j$ and $\text{PD}_{x_j}$. Equation \eqref{eq_pdp} can be extended to a two-way interaction partial dependence effect by letting two variables range over their respective variable spaces.

\begin{figure}[ht!]
\begin{adjustwidth}{-1.2cm}{-1.2cm}
\centering
  \includegraphics[width = \linewidth]{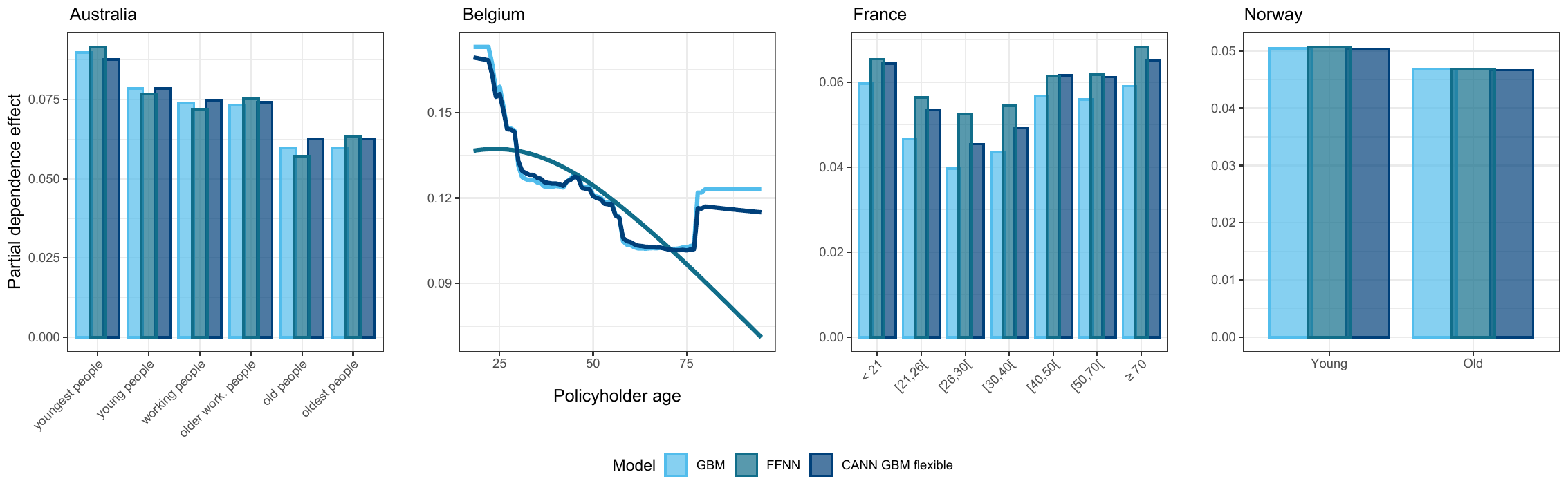}
\end{adjustwidth}
\caption{Partial dependence effect of the policyholder's age across the four data sets, claim frequency models. We compare the benchmark GBM, the FFNN and the CANN GBM flexible.}
\label{fig_pdp_ageph}
\end{figure}
 
Figure \ref{fig_pdp_ageph} shows the partial dependence effect between the policyholder's age and the predicted claim frequency across the four data sets in the benchmark study. We compare the effects of the benchmark GBM, the FFNN and the CANN GBM flexible. The effect in all three models is similar for the Australian, French and Norwegian data. However, for the Belgian data set, the GBM and CANN GBM flexible show a similar partial dependence effect, while the FFNN shows a very different pattern. The partial dependence effect of this FFNN shows a less complex, less nuanced relationship between age of the policyholder and claim frequency. Across the four data sets, the average predicted claim frequency decreases with age, which is an expected relationship between age and claim frequency. For the Belgian and French data sets, we observe an increasing effect for the older ages.

\begin{figure}[ht!]
\begin{adjustwidth}{-1.2cm}{-1.2cm}
\centering
  \includegraphics[width = \linewidth]{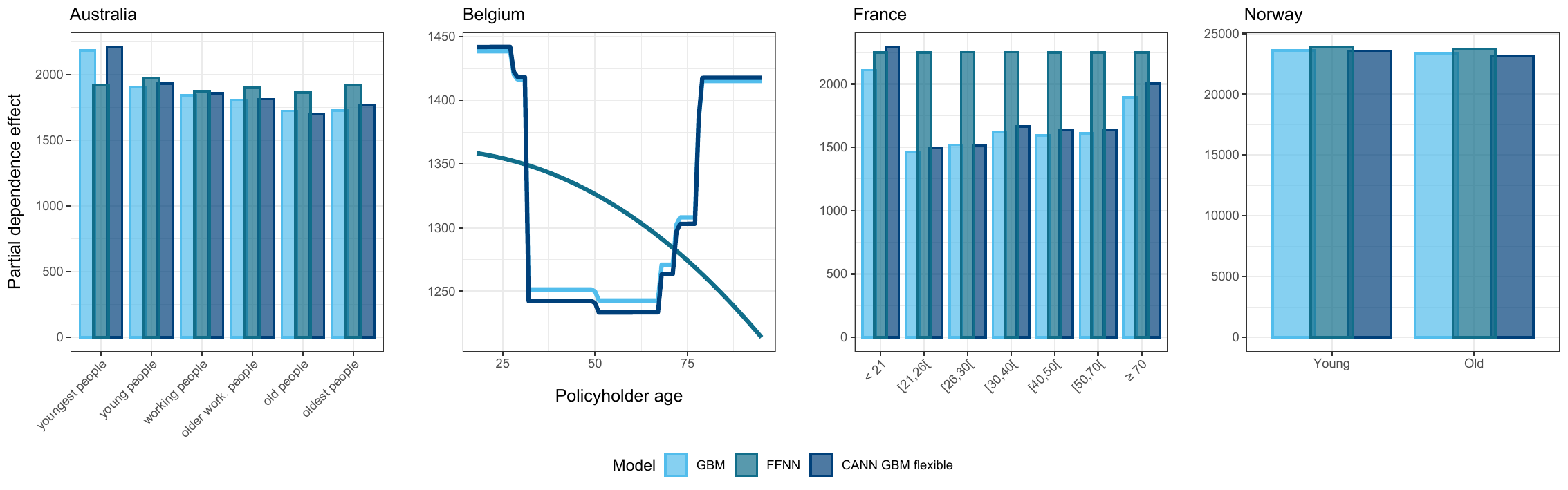}
\end{adjustwidth}
\caption{Partial dependence effect of the policyholder's age across the four data sets, claim severity models. We compare the benchmark GBM, the FFNN and the CANN GBM flexible.}
\label{fig_pdp_ageph_SEV}
\end{figure}

Figure \ref{fig_pdp_ageph_SEV} displays the partial dependence effect of the policyholder's age when calibrated on the claim severity data. Similar to the effects portrayed in Figure \ref{fig_pdp_ageph}, the three models applied to the Australian and Norwegian data sets exhibit a comparable effect. For the Belgian and French data sets, the FFNN showcases a notably distinct partial dependence effect. Specifically for the French data, the FFNN model reveals an almost flat effect across all age groups.

\begin{figure}[ht!]
\centering
  \includegraphics[width = 0.7\linewidth]{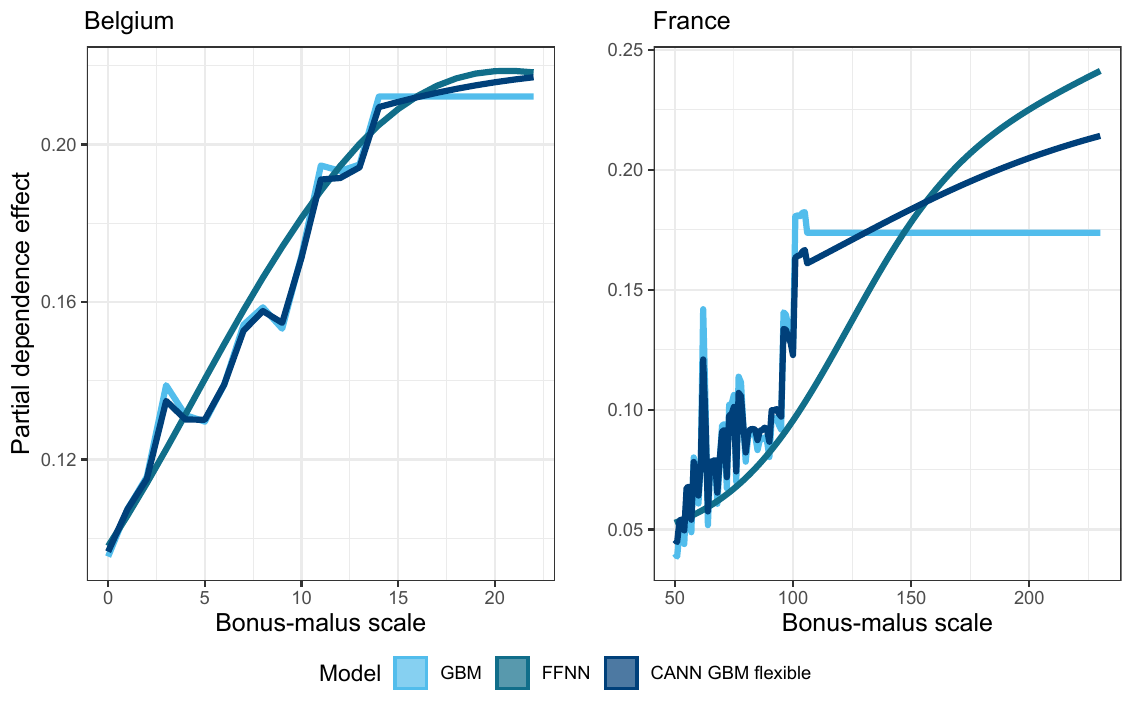}
\caption{Partial dependence effect of the bonus-malus score for the Belgian data set, claim frequency model (left) and claim severity model (right). We compare the benchmark GBM, the FFNN and the CANN GBM flexible.}
\label{fig_pdp_bm}
\end{figure}

Figure \ref{fig_pdp_bm} shows the partial dependence effect of the bonus-malus score for the Belgian and French frequency data sets. For both data sets, the three models show an increasing relation between the level occupied in the bonus-malus scale and the expected claim frequency. According to the FFNN, the partial dependence is a distinctly smoother effect compared to the effect calibrated by the GBM and CANN GBM flexible, showing again the less complex, less nuanced relationships captured by the FFNN. 

\begin{figure}[ht!]
\centering
  \includegraphics[width = 0.72\linewidth]{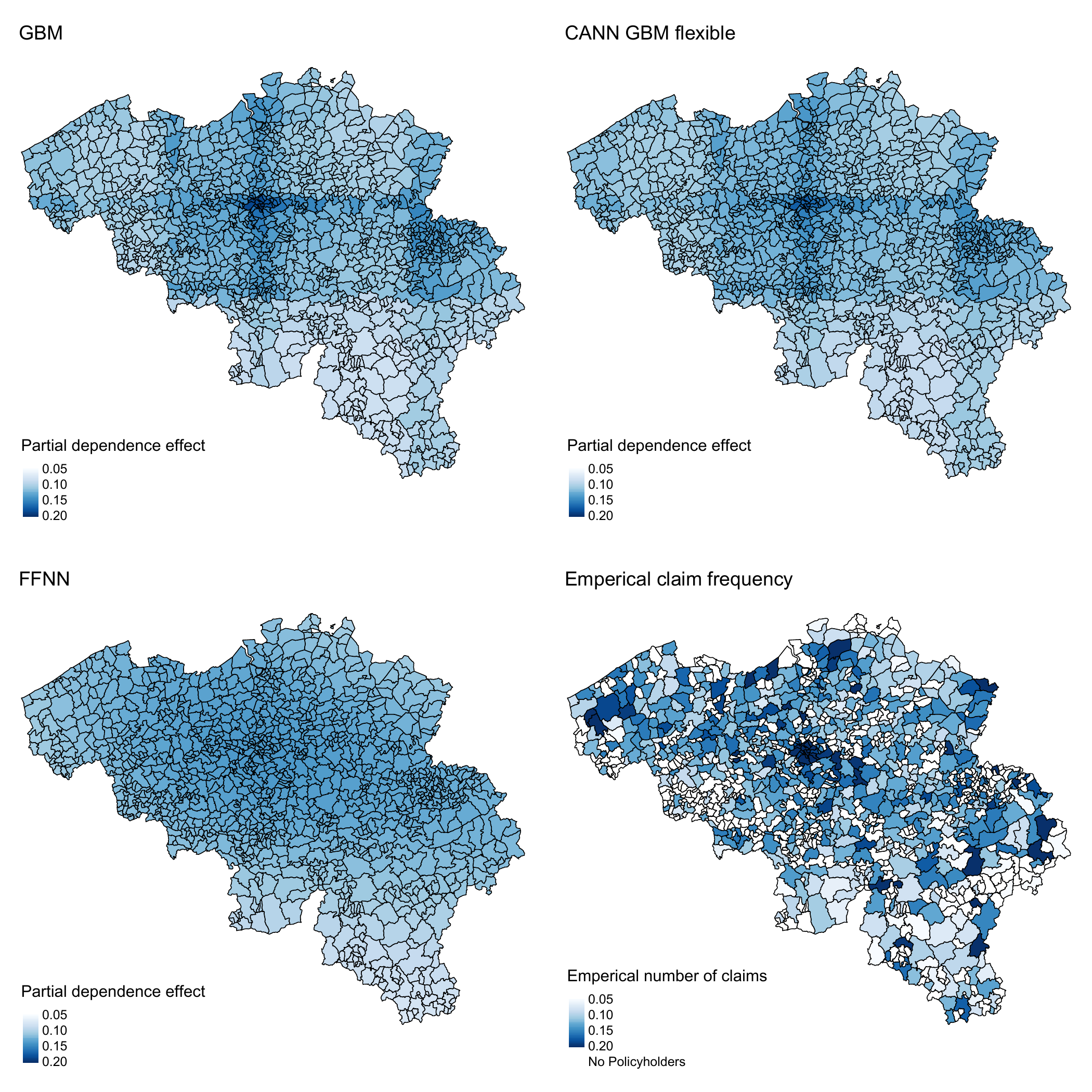}
\caption{Partial dependence relationship of the spatial variable and the expected number of claims in the GBM, FFNN and CANN GBM flexible for the Belgian data. We compare the modelled effects with the empirical claim frequency in the Belgian data set.}
\label{fig_pdp_spatial}
\end{figure}

We display the partial dependence effect of the postal code in the Belgian frequency data set in Figure \ref{fig_pdp_spatial}. We compare the partial dependence effect with the empirical claim frequency in the Belgian data, calculated as the number of claims per postal code divided by the sum of the exposure for that postal code. The effect in the GBM and CANN GBM flexible is very similar, with a higher expected claim frequency around the capital of Belgium. The effect in the FFNN also shows a higher expected number of claims in the capital but the calibrated spatial effect is much smoother. This aligns with the smoother partial dependence effects for the policyholder age and bonus-malus in the Belgian frequency FFNN model. Empirically, we see a higher concentration of claims per unit of exposure in and around the capital and for some postal codes in the west and east of Belgium. This effect is visible for the GBM and CANN model but not for the FFNN. 

\subsection{Surrogate models for practical applications}\label{sec_surrogates}

\paragraph{Surrogate model construction}
\cite{Henckaerts2022} present a workflow for constructing a surrogate GLM by leveraging insights obtained with a black box model. In our study, we apply this technique to the CANN model with GBM input, as discussed in Section \ref{sec_NNbased} and calibrated in Section \ref{sec_Bench}. To create the surrogates, we first calculate the partial dependence effect for each individual variable and for interactions between any two variables, as discussed in Section \ref{sec_pdp}. Next, we use the dynamic programming algorithm introduced by \cite{wang2011} to segment the input data into homogeneous groups based on these partial dependence effects. On the resulting binned dataset, we fit a generalized linear model for all combinations of segmented input variables and then select the optimal GLM based on BIC. Constructing the surrogate GLM on the segmented frequency and severity data leads to a tabular premium structure incorporating the insights captured by the CANN architectures.

\begin{figure}[ht!]
\begin{adjustwidth}{-1.6cm}{-1cm}
\centering
\begin{subfigure}[T]{0.60\linewidth}
  \includegraphics[height=6cm, keepaspectratio]{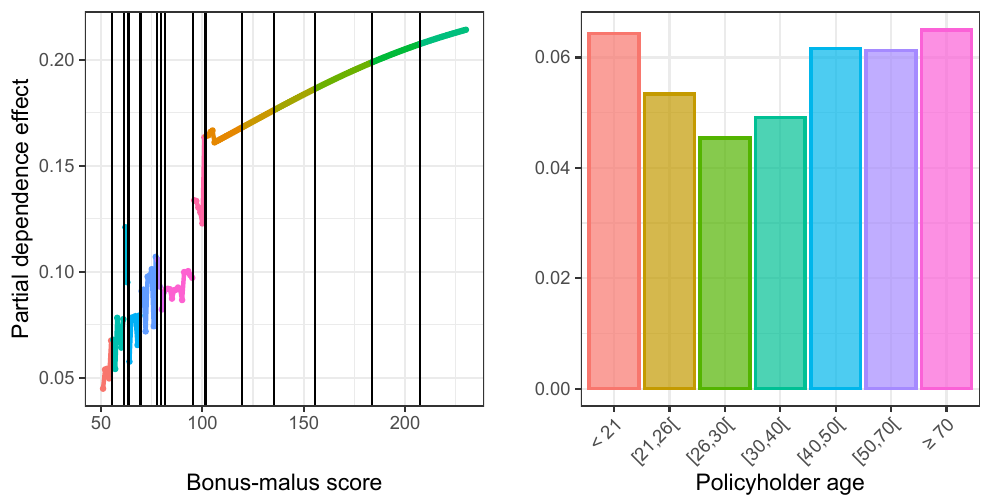}
\end{subfigure}%
\hfill
\begin{subfigure}[T]{0.30\linewidth}
  \includegraphics[height=6cm, keepaspectratio]{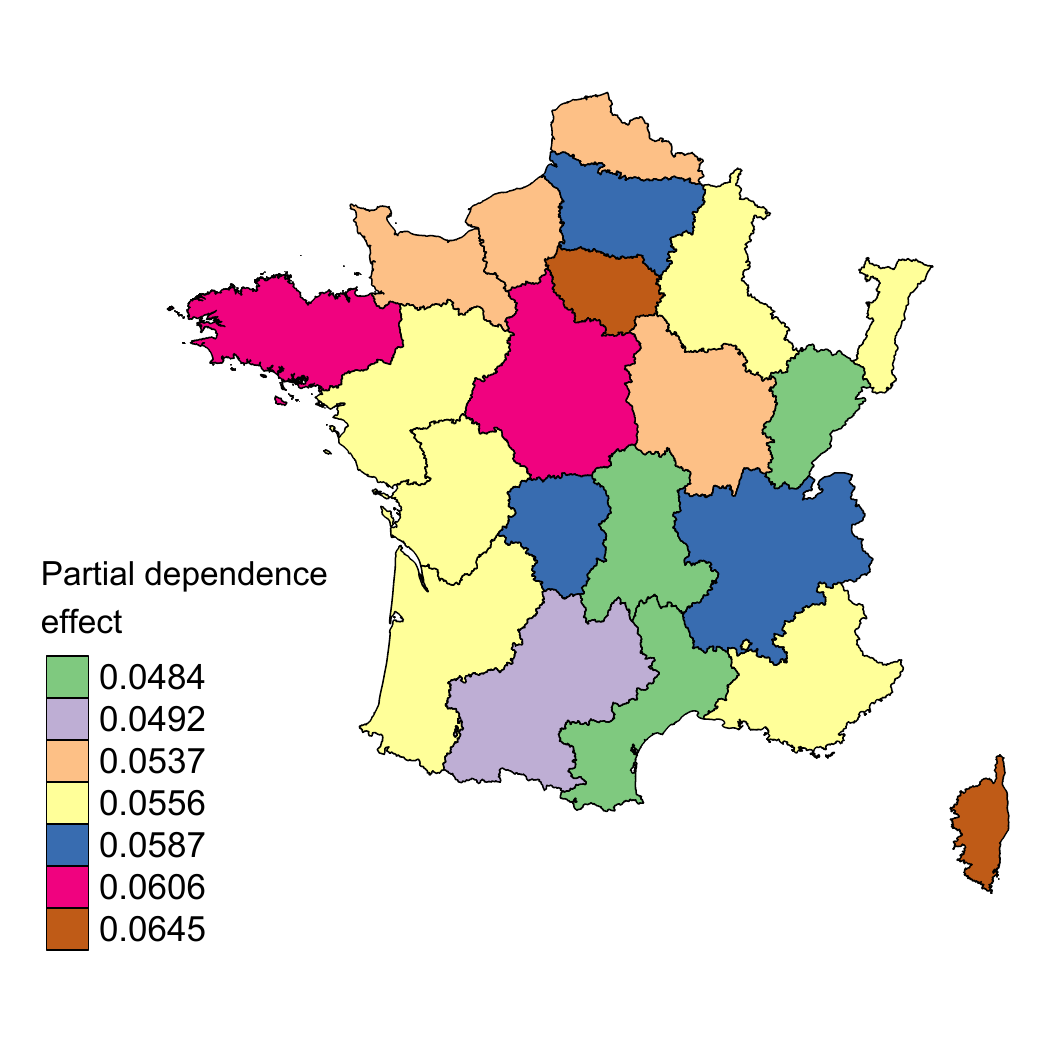}
\end{subfigure}%
\end{adjustwidth}
\caption{Partial dependence plots for three variables in the French data set; left to right: bonus-malus scale, policyholder age and region. In color, we show the binning of the input data used in the frequency surrogate GLM. Each color represents one bin of the input variable.}
\label{fig_surr_illustration}
\end{figure}

Figure \ref{fig_surr_illustration} shows the partial dependence effects of the CANN GBM flexible for the bonus-malus score, the policyholder age and the region variable from the French data set, with respect to frequency modeling. The color coding shows the obtained data segmentation. \added{Note that when segmenting a categorical variable encoded using an autoencoder, the segmentation is performed on the original dataset $\mathcal{D}$, not on the encoded values.} The so-called surrogate GLM is fitted on the segmented input data. The benchmark GLM used in the comparative analysis in Section \ref{sec_evalframe} is also fitted on binned data. However, as explained in Section \ref{sec_benchmodel}, this GLM was constructed by binning the smooth effects calibrated by a GAM. Therefore, it is insightful to compare both the predictive accuracy and the selected variables as obtained with both techniques. To avoid data leakage in the comparison between two models, we compare the predictive accuracy on the withheld test set $\mathcal{D}_1$. 

\begin{table}[ht!]
\scriptsize
\begin{adjustwidth}{-1.85cm}{-1.85cm}
\begin{subtable}[t]{.325\linewidth}
\begin{NiceTabular}[t]{ll}[
code-before = \rowcolor[HTML]{FFFFFF}{1,3,5,6}
              \rowcolor[HTML]{FAFAFF}{2,4}
]
\toprule
\textbf{Benchmark GLM} & \textbf{Surrogate GLM} \\
\noalign{\hrule height 0.3pt}
VehValue & VehValue \\
DrivAge & DrivAge \\
 & VehBody \\
 & VehValue $\times$ DrivAge \\
 \noalign{\hrule height 0.3pt}
 $\mathbf{0.3817}$ & $\mathbf{0.3805}$ \\
\bottomrule
\end{NiceTabular}
\caption{Australian data set}
\label{tab_surr_GLM_A}
\end{subtable} 
\hfill
\begin{subtable}[t]{.325\linewidth}
\begin{NiceTabular}[t]{ll}[
code-before = \rowcolor[HTML]{FFFFFF}{1,3,5,7,9,11,12}
              \rowcolor[HTML]{FAFAFF}{2,4,6,8,10}
]
\toprule
\textbf{Benchmark GLM} & \textbf{Surrogate GLM} \\
\noalign{\hrule height 0.3pt}
ageph & ageph \\
bm & bm \\
power & power \\
coverage &  coverage\\
 & agec \\
fuel & fuel \\
 & fleet \\
 & sex \\
postcode & postcode \\
ageph $\times$ power &  \\
 \noalign{\hrule height 0.3pt}
 $\mathbf{0.5314}$ & $\mathbf{0.5308}$ \\
\bottomrule
\end{NiceTabular}
\caption{Belgian data set}
\label{tab_surr_GLM_B}
\end{subtable} 
\hfill
\begin{subtable}[t]{.325\linewidth}
\begin{NiceTabular}[t]{ll}[
code-before = \rowcolor[HTML]{FFFFFF}{1,3,5,7,9,11,13,15,17,19}
              \rowcolor[HTML]{FAFAFF}{2,4,6,8,10,12,14,16,18}
]
\toprule
\textbf{Benchmark GLM} & \textbf{Surrogate GLM} \\
\noalign{\hrule height 0.3pt}
VehValue & VehPower \\
VehAge & VehAge \\
DrivAge & DrivAge \\
BonusMalus & BonusMalus \\
VehGas & VehGas \\
Density & Density \\
 & VehBrand \\
 & Region \\
 & VehBrand $\times$ Region \\
  & Density $\times$ Region \\
 & VehPower $\times$ Region \\
 & VehPower $\times$ VehBrand \\
 & DrivAge $\times$ Region \\
 & BonusMalus $\times$ Region \\
 & VehBrand $\times$ Density \\
 & BonusMalus $\times$ VehBrand \\
 & DrivAge $\times$ BonusMalus \\
 \noalign{\hrule height 0.3pt}
 $\mathbf{0.2761}$ & $\mathbf{0.2738}$ \\
\bottomrule
\end{NiceTabular}
\caption{French data set}
\label{tab_surr_GLM_C}
\end{subtable} 
\end{adjustwidth}
\caption{Comparison between the benchmark GLM and the surrogate GLM for frequency modeling on the Australian, Belgian and French data sets. The surrogate GLM is constructed from the CANN with GBM input and flexible output layer. The last row shows the Poisson deviance of both GLMs on the out-of-sample data set $\mathcal{D}_1$.}
\label{tab_surr_all_GLM}
\end{table}

Table \ref{tab_surr_all_GLM} shows the variables included in the benchmark GLMs and the surrogate GLMs for the Australian, Belgian and French data set. The out-of-sample performances of these models are evaluated on the withheld data set $\mathcal{D}_1$. We excluded the Norwegian data set from the surrogate fitting, as this data set only consists of categorical variables. The surrogate technique selects more variables and performs better on the out-of-sample test set than the benchmark GLM. This finding is consistent across all three data sets. Hence, the surrogate GLM benefits from the insights learned from the neural network adjustments in the CANN compared to the direct construction of the benchmark GLM.

\paragraph{Identification of risk profiles}

We estimate the expected number of claims and the expected claim severity using the surrogate models constructed for the frequency and severity CANN GBM flexible, respectively. 

\begin{table}[ht!]
%\begin{adjustwidth}{-0.3cm}{-0.3cm}
\centering
%\scriptsize
\begin{NiceTabular}{lccc}[
code-before = \rowcolor[HTML]{FFFFFF}{1,3,5,7,9,10,12}
              \rowcolor[HTML]{FAFAFF}{2,4,6,8,11}
]
\toprule
\textbf{Variables} & \textbf{Low risk} & \textbf{Medium risk} & \textbf{High risk}\\
\noalign{\hrule height 0.3pt}
Vehicle power & $4$ & $6$ & $9$ \\
Vehicle age & $3$ & $2$  & $1$ \\
Policyholder age & $21$ & $34$ & $72$ \\
Bonus-malus scale & $50$ & $70$ & $190$ \\
Vehicle brand & B12 & B5 & B11 \\
Fuel type & Regular & Regular & Diesel \\
Population density of area & $2.71$ & $665.14$ & $22\,026.47$ \\
District of residence & Midi-Pyrenees & Basse-Normandie & Corse \\
\noalign{\hrule height 0.3pt}
\noalign{\medskip}
\multicolumn{4}{l}{\textbf{Predicted number of claims}} \\
\noalign{\hrule height 0.3pt}
Surrogate GLM & $0.020$ & $0.106$ & $0.361$ \\
CANN GBM flexible & $0.021$ & $0.101$ & $0.519$ \\
\bottomrule
\end{NiceTabular}
\caption{Example of a low, medium and high risk profile for the French data set, using the surrogate GLM constructed for the CANN model with GBM input and flexible output layer, withholding test set one. We compare the predicted number of claims for each profile.}
\label{tab_surr_example}
%\end{adjustwidth}
\end{table}

\begin{figure}[ht!]
\begin{adjustwidth}{-1cm}{-1cm}
\centering
  \includegraphics[width = \linewidth]{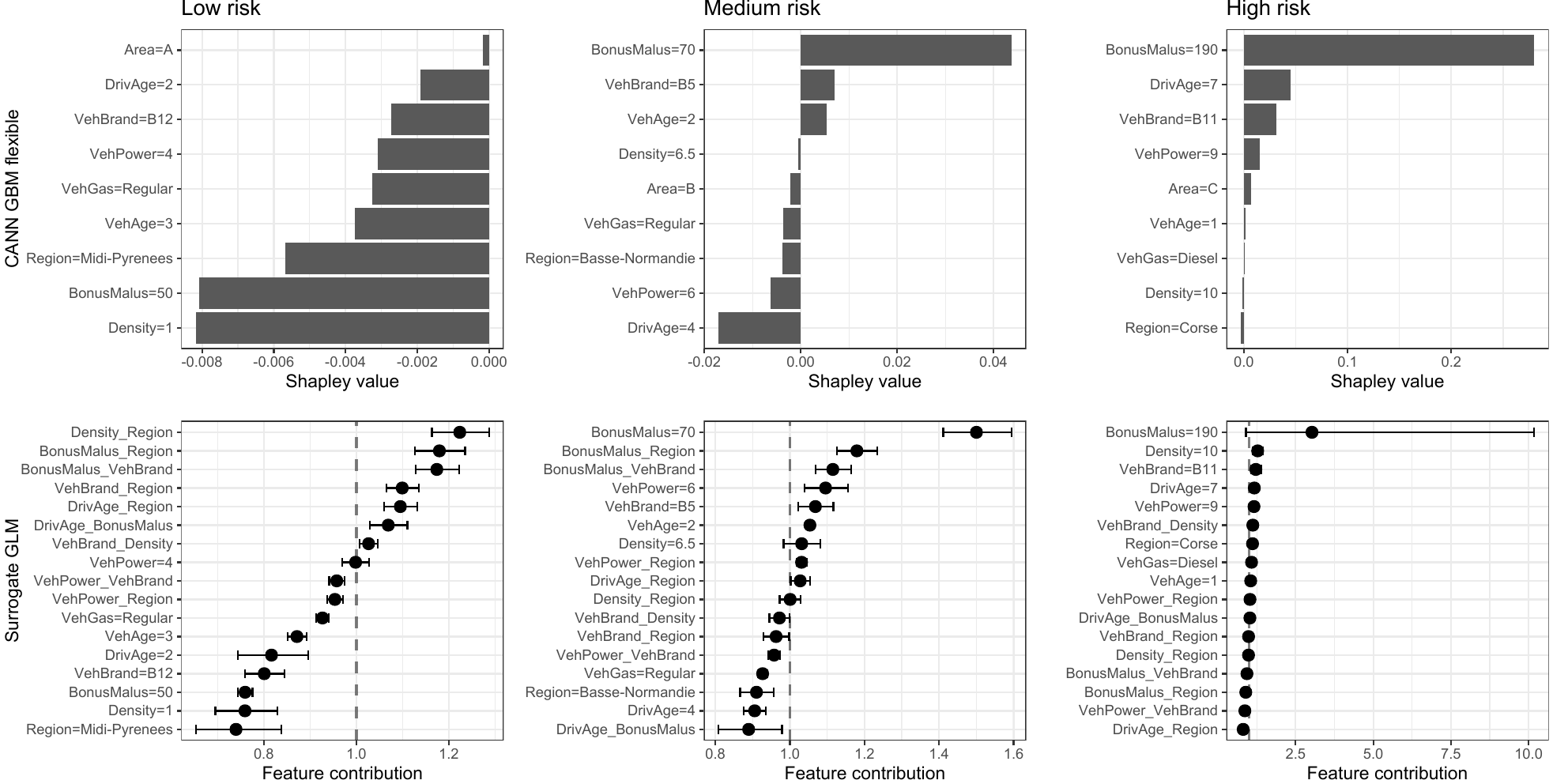}
\end{adjustwidth}
\caption{Comparison between the low, medium and high risk profiles in the frequency models on the French data set according to the Shapley values of the CANN GBM flexible model (top row) and the fitted coefficients in the surrogate GLM (bottom row).}
\label{fig_shapley}
\end{figure}

We select a low, medium, and high-risk profile from the French data set based on the frequency surrogate GLM. Table \ref{tab_surr_example} compares these profiles via their expected claim frequency according to the surrogate GLM and the CANN GBM flexible model. We compare the influence of each variable on the assessed risk using two local interpretation tools in Figure \ref{fig_shapley}. For the GLM, we show the fitted coefficients on the response scale. A value lower (higher) than one means the feature's value leads to a lower (higher) prediction than the baseline prediction obtained with the intercept of the GLM. The uncertainty of each contribution is shown with the $95\%$ confidence interval. Shapley values \citep{shapley1953value} are used to compare the feature contributions in the GLM to the influences in the CANN model. A positive (negative) Shapley value indicates that this feature's value leads to a higher (lower) than average prediction. The effects in the GLM and CANN models mostly align. We see a strong impact of the variables region, driver age and bonus-malus score on the predicted number of claims. The variable area was not selected in the surrogate GLM construction, and its Shapley value is negligible in all three risk profiles.

\paragraph{Model evaluation framework}

\added{We use the model evaluation framework from Section \ref{sec_evalframe} to compare the predictions' structure and to evaluate the predictive dominance and predictive accuracy of the surrogate GLM when compared to the benchmark GLM. This type of comparison is valuable as it directly measures whether a GLM with integrated deep-learning insights can compete with an industry-standard GLM, offering insights into its practical applicability and potential for adoption. We restrict this section to the comparison between the frequency surrogate GLM and refer to Appendix \ref{app_sev_surr} for the results on the severity surrogate GLM and to Appendix \ref{app_surr_cann} for the comparison between the surrogate GLM and the CANN GBM flexible. The out-of-sample deviances are shown in Table \ref{tab_surr_all_GLM}, where we see that the surrogate GLM leads to lower deviance on the test set when compared to the benchmark GLM. Table \ref{tab_DB_surr} shows the results of the Diebold-Mariano test, comparing the predictive accuracy of the benchmark GLM and the surrogate model. For both the Australian and Belgian data, we cannot reject the null hypothesis for any model. For the French data, when looking at the benchmark GLM as model A, the Diebold-Mariano test rejects the null hypothesis in favor of model B, meaning the surrogate has higher predictive accuracy than the benchmark GLM.}

\setlength{\extrarowheight}{3pt} % a bit of extra white space between lines in tables
\begin{table}[ht!]
 \begin{adjustwidth}{-0.3cm}{-0.3cm}
  \centering
  \scriptsize
  \begin{subtable}[t]{.3\linewidth}
    \begin{NiceTabular}{cr*{2}{r}}[
    code-before = \rowcolor[HTML]{FFFFFF}{1,2,4}
              \rowcolor[HTML]{FAFAFF}{3}
    ]
    \toprule
    \multirow{14}{*}{\rotatebox[origin=c]{90}{\footnotesize \textbf{Model A}}} & \multicolumn{3}{r}{\footnotesize \textbf{Model B}} \\[1mm]
    \RowStyle{\rotate}
    & & Benchmark GLM & Surrogate GLM \\\cline{2-4}
    & Benchmark GLM &  & \cmark \\ 
    & Surrogate GLM & \cmark &  \\ 
    \bottomrule
    \end{NiceTabular}
  \caption{Australian data set}
  \label{tab_DB_surr_AUS}
  \end{subtable} 
  \hfill
  \begin{subtable}[t]{.3\linewidth}
    \begin{NiceTabular}{cr*{2}{r}}[
    code-before = \rowcolor[HTML]{FFFFFF}{1,2,4}
              \rowcolor[HTML]{FAFAFF}{3}
    ]
    \toprule
    \multirow{14}{*}{\rotatebox[origin=c]{90}{\footnotesize \textbf{Model A}}} & \multicolumn{3}{r}{\footnotesize \textbf{Model B}} \\[1mm]
    \RowStyle{\rotate}
    & & Benchmark GLM & Surrogate GLM \\\cline{2-4}
    & Benchmark GLM &  & \cmark \\ 
    & Surrogate GLM & \cmark &  \\ 
    \bottomrule
    \end{NiceTabular}
  \caption{Belgian data set}
  \label{tab_DB_surr_BE}
  \end{subtable} 
  \hfill
  \begin{subtable}[t]{.3\linewidth}
    \begin{NiceTabular}{cr*{2}{r}}[
    code-before = \rowcolor[HTML]{FFFFFF}{1,2,4}
              \rowcolor[HTML]{FAFAFF}{3}
    ]
    \toprule
    \multirow{14}{*}{\rotatebox[origin=c]{90}{\footnotesize \textbf{Model A}}} & \multicolumn{3}{r}{\footnotesize \textbf{Model B}} \\[1mm]
    \RowStyle{\rotate}
    & & Benchmark GLM & Surrogate GLM \\\cline{2-4}
    & Benchmark GLM &  & \cellcolor{blue!25}\xmark \\ 
    & Surrogate GLM & \cmark &  \\ 
    \bottomrule
    \end{NiceTabular}
  \caption{French data set}
  \label{tab_DB_surr_FR}
  \end{subtable} 
  \end{adjustwidth}
  \caption{Results of the Diebold-Mariano test comparing the predictive accuracy between the benchmark GLM and the surrogate GLM. The table indicates whether we cannot reject (\cmark) the null hypothesis $H_0: \text{accuracy}(\text{Model A}) = \text{accuracy}(\text{Model B})$, or if we reject $H_0$ (\xmark) in favor of the alternative hypothesis $H_1: \text{accuracy}(\text{Model B}) > \text{accuracy}(\text{Model A})$. Highlighted cells indicate when the null hypothesis is rejected.}
  \label{tab_DB_surr}
\end{table}

\begin{figure}[ht!]
\begin{adjustwidth}{-1.6cm}{-1cm}
\centering
  \includegraphics[width = \linewidth]{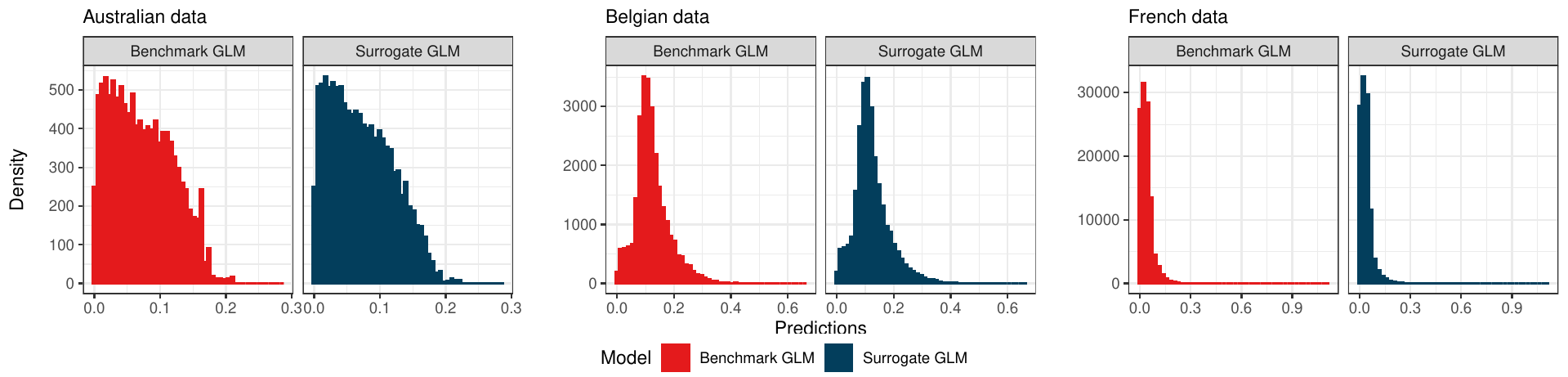}
\end{adjustwidth}
\caption{Dispersion of predictions made by the GLM compared to the surrogate model based on the CANN GBM flexible.}
\label{fig_histo_surr}
\end{figure}

\added{Figure \ref{fig_histo_surr} shows the histogram of the predictions made by the benchmark GLM compared to those from the surrogate model. These prediction structures are very comparable between the two models for all three data sets, meaning the surrogate does not lead to a significantly different dispersion in the predictions.}

\begin{figure}[ht!]
\begin{adjustwidth}{-1.6cm}{-1cm}
\centering
  \includegraphics[width = \linewidth]{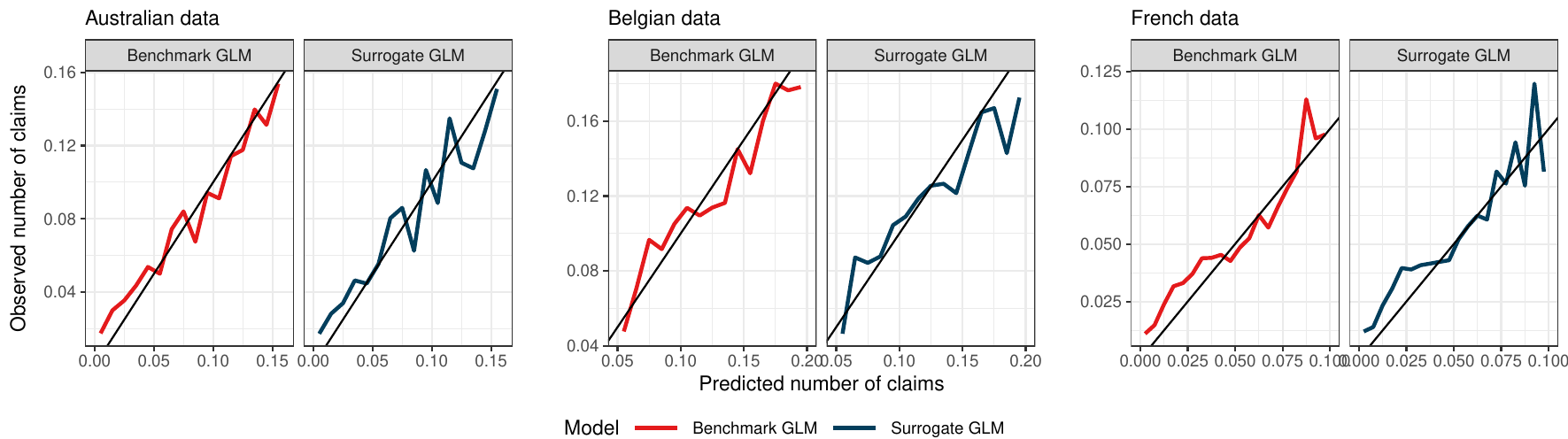}
\end{adjustwidth}
\caption{Plot of $\E \left[Y | f(\X) = s \right]$ over the range of predictions $s$ for the benchmark GLM and the surrogate GLM. Predictions are made on test set $\mathcal{D}_1$. A line above (below) the diagonal indicates that the model is underestimating (overestimating) the true number of claims in the data.}
\label{fig_calib_surr}
\end{figure}

\added{Figure \ref{fig_calib_surr} shows the plot of $\E \left[Y | f(\X) = s \right]$ over all $s$ for both the benchmark GLM and the surrogate GLM. For the Australian data, the plot of the benchmark GLM is close to the diagonal for all $s$, while the plot for the surrogate is slightly more volatile. On the Belgian data, both GLMs seem well calibrated, with underestimation only for high values of $s$ in the surrogate GLM. This also holds for the French data set with both GLMs being well calibrated but the surrogate GLM slightly underestimating for high values of $s$. As the plots are constructed on a rather small test set, the slight differences between the calibration of the benchmark GLM and the surrogate GLM do not lead to a conclusion on whether one model is preferable over the other.}

\begin{figure}[ht!]
\centering
  \includegraphics[width = \linewidth]{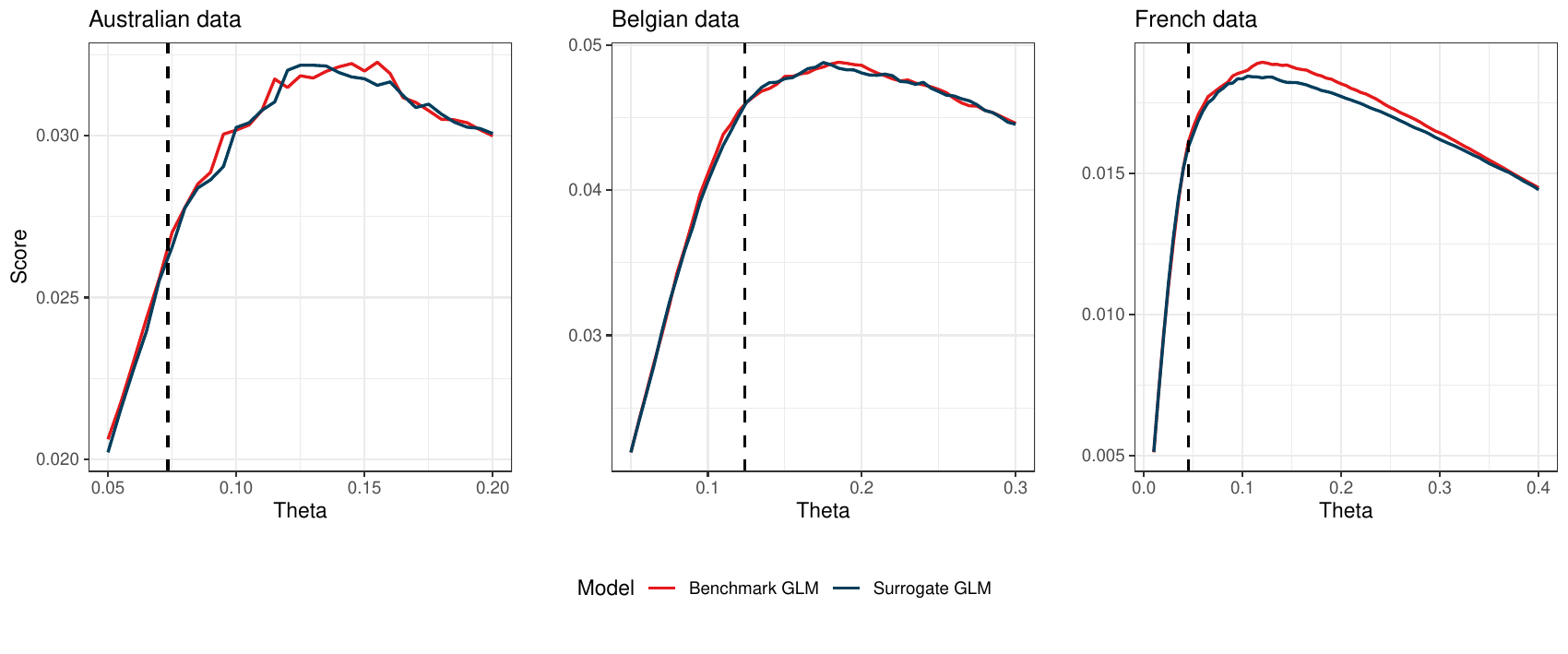}
\caption{Murphy diagrams to compare predictive dominance between the benchmark GLM model and the surrogate GLM based on the CANN GBM flexible.}
\label{fig_murphy_surr}
\end{figure}

\added{Figure \ref{fig_murphy_surr} shows the Murphy diagram for the benchmark GLM compared with the surrogate model. For both the Australian and Belgian data, the score of the surrogate model is not lower than that of the GLM for all $\theta$, so we cannot conclude the surrogate model has predictive dominance over the benchmark GLM, nor that the benchmark GLM has predictive dominance over the surrogate. For the French data, however, the surrogate model leads to a lower value of $S_\theta$ for all $\theta$, so the surrogate does have predictive dominance over the benchmark GLM. This is in line with the results of the Diebold-Mariano test in Table \ref{tab_DB_surr}.}

\added{Combining the insights of the model evaluation techniques, we conclude that for frequency modeling on the Australian and Belgian data set, the surrogate GLM is statistically as accurate as the benchmark GLM according to the Diebold-Mariano test, even though the surrogate GLM has a lower out-of-sample deviance. The prediction structure shows that the surrogate GLM is less well-calibrated when compared to the benchmark GLM, and the Murphy diagram shows that no model has predictive dominance over the other. On the French data set, the Diebold-Mariano test shows the surrogate model has a higher predictive accuracy than the benchmark GLM and has predictive dominance according to the Murphy diagram. Hence, for this data set, there is added value in leveraging the construction of the GLM with insights captured with more advanced deep learning architectures.}

\section{Managerial insights: a comparison between technical tariff structures}

We now combine the predictions for claim frequency and severity into a technical tariff. For each data set and each fold, we evaluate the predictions for all observations in the test set $\mathcal{D}_\ell$, using the model trained on the data subsets $\mathcal{D}\setminus\mathcal{D}_\ell$. As such, we obtain out-of-sample predictions for the expected number of claims and the expected claim severities for each policyholder in the data set. The predicted loss, or technical tariff, for each policyholder is the expected number of claims times the expected claim severity. \added{This section focuses on comparing the tariff structures of the binned GLM with the surrogate GLM and evaluating whether the CANN model with GBM input offers added value over the original GBM.}

\added{Table \ref{tab_totallosses} shows the total predicted loss next to the total loss observed in each data set; we also show the ratio of predicted losses over the observed losses.} A ratio of one means the model has a perfect balance at portfolio level. For the Norwegian data set, the predicted losses are very close to the observed losses for all models. For the Australian and Belgian data, both tariffs constructed from GLM models are close to balance, meaning the predicted losses are close to the observed losses. Although a canonical link GLM satisfies the balance property \citep{Nelder1972}, our severity models use a gamma distribution with non-canonical log-link, and the tariff structures shown here are based on out-of-sample predictions. The tariff obtained with GBM and the CANN model deviates slightly from perfect balance.

\setlength{\extrarowheight}{3pt} % a bit of extra whitespaces between lines in tables
\begin{table}[!h]
%\begin{adjustwidth}{-1cm}{-1cm}
\scriptsize
\centering
\begin{NiceTabular}{>{\raggedleft\arraybackslash}m{0.05cm}>{\raggedleft\arraybackslash}m{2.2cm}>{\raggedleft\arraybackslash}m{2cm}>{\raggedleft\arraybackslash}m{1.8cm}>{\raggedleft\arraybackslash}m{1.8cm}>{\raggedleft\arraybackslash}m{1.8cm}>{\raggedleft\arraybackslash}m{1.8cm}}[
code-before = \rowcolor[HTML]{FFFFFF}{1,2,4,5,7,9,11}
              \rowcolor[HTML]{FAFAFF}{3,6,8,10}
]
\toprule

 & & \textbf{Observed} & \textbf{Benchmark GLM} & \textbf{GBM} & \textbf{CANN GBM flex} & \textbf{Surrogate GLM} \\
\noalign{\hrule height 0.3pt}
\noalign{\medskip}
\multicolumn{7}{l}{\textbf{Observed and predicted losses}} \\
\noalign{\hrule height 0.3pt}
& Australia (AU\$) & $9\,314\,604$ & $9\,345\,113$ & $9\,136\,324$ & $9\,154\,467$ & $9\,355\,718$\\
& Belgium (\euro) & $26\,464\,970$ & $26\,399\,027$ & $26\,079\,709$ & $25\,720\,143$ & $26\,345\,969$ \\
& France (\euro) & $58\,872\,147$ & $56\,053\,341$ & $56\,207\,993$ & $58\,629\,584$ & $57\,048\,375$ \\
& Norway (NOK) & $206\,649\,080$ & $206\,634\,401$ & $206\,475\,980$ & $206\,494\,683$ & - \\
\noalign{\hrule height 0.3pt}
\noalign{\medskip}
\multicolumn{7}{l}{\textbf{Ratio of predicted losses over observed losses}} \\
\noalign{\hrule height 0.3pt}
& Australia & - & $1.00$ & $0.98$ & $0.98$ & $1.00$ \\
& Belgium & - & $1.00$ & $0.99$ & $0.97$ & $1.00$ \\
& France & - & $0.95$ & $0.95$ & $1.00$ & $0.97$\\
& Norway & - & $1.00$ & $1.00$ & $1.00$ & - \\
\bottomrule
\end{NiceTabular}
\caption{Comparison between the total observed losses for each data set and the total predicted losses for the benchmark GLM, GBM, CANN GBM flex and the surrogate GLM. Each prediction is made out-of-sample. We also show the ratio of total predicted losses over observed losses.}
\label{tab_totallosses}
%\end{adjustwidth}
\end{table}

To compare tariff structures, we follow the methodology from \citet{henckaerts2022added} using risk scores. For a model $f$, let $F_n$ be the empirical cumulative distribution function of the predictions made by the model $f$. For each policyholder $i$, the risk score $r^f_i$ is the evaluation of $F_n$ in $f(\x_i)$. The risk score is calculated as 

\begin{equation}
    r^f_i = F_n\left(f(\x_i)\right).
    \label{eq_riskscore}
\end{equation}

For frequency-severity modeling, with a frequency model $f^{\text{freq}}$ and a severity model $f^{\text{sev}}$, we replace $f(\x_i)$ with $f^{\text{freq}}(\x_i)\times f^{\text{sev}}(\x_i)$.

We compare the risk scores of multiple models using Lorenz curves \citep{lorenz1905methods}. For a model $f$, the Lorenz curve evaluated in $s\in[0,1]$ is 

\begin{equation}
    LC^f(s) = \frac{\sum_{i=1}^{n} L_i\,\mathbbm{1}\{r_i^f\le s\}}{\sum_{i=1}^{n} L_i},  
    \label{eq_lorenz}
\end{equation}

with $L_i$ the observed loss for policyholder $i$. We visualize the Lorenz curve by plotting the pairs $\left(s,LC^f(s)\right)$, for $s\in[0,1]$. The Lorenz curves shows the accumulation of losses, ordered by risk score $r^f_i$ obtained with model $f$. A model with a better risk classification accumulates losses slower for low risk scores and faster for high risk scores.  A Lorenz curve further away from the equality line at $45\%$, represents a better risk classification.

\begin{figure}[ht!]
\begin{adjustwidth}{-1.2cm}{-1.2cm}
\centering
  \includegraphics[width = \linewidth]{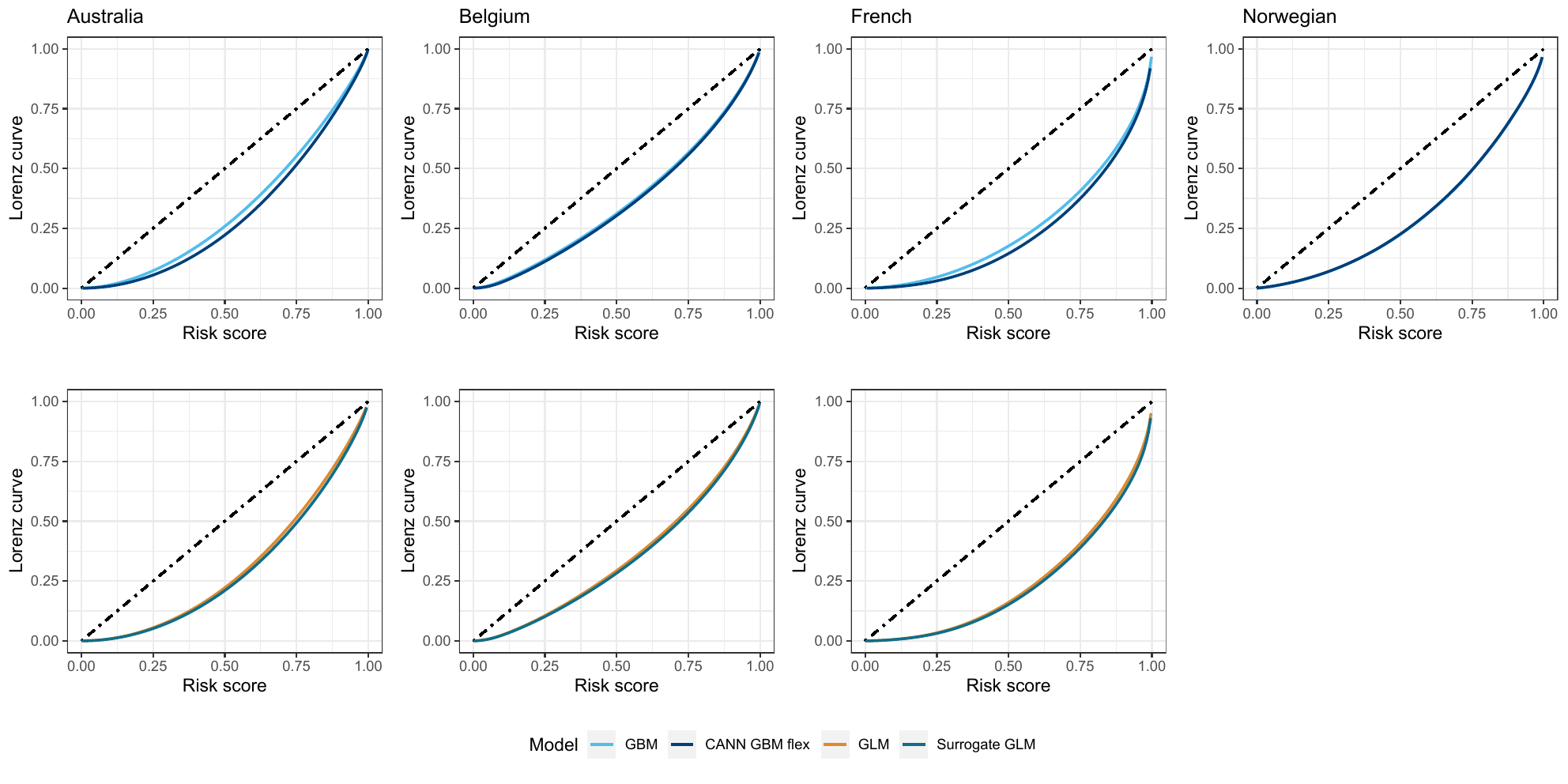}
\end{adjustwidth}
\caption{Lorenz curve comparison between the GBM benchmark model and the CANN GBM flexible model in the top row. Bottom row compares the GLM benchmark with the surrogate GLM. A dashed line is added to show the line of equality.}
\label{fig_lorenzcurve}
\end{figure}

\added{The top row in Figure \ref{fig_lorenzcurve} shows the Lorenz curves of the GBM versus the CANN GBM flexible, while the bottom row shows the benchmark GLM versus the surrogate GLM.} For the Australian and French data sets, the tariff structure from the CANN model is (slightly) preferred to that of the benchmark GBM according to the Lorenz curve. For the Belgian and Norwegian data sets, the CANN model is also preferred, but the two curves are very similar. For the Australian, Belgian and French data sets, the Lorenz curves of the benchmark GLM and the surrogate GLM show a very similar pattern, with the surrogate model having a preferable curve over the benchmark GLM, showing that the higher predictive accuracy of the surrogate model also results in a slightly better risk classification in the tariff structure.

\added{Next to the visual comparison on risk classification using the Lorenz curves, we use the ordered Lorenz curve, introduced by \citet{freesgini}, to directly compare the tariff structure under one model relative to another. The ordered Lorenz curve is constructed by looking at the accumulation of losses ordered by relativities instead of risk scores. For two models $f_A$ and $f_B$, we define the relativities of model B over model A as} 

\begin{equation}\text{rel}^{(A,B)}_i = \frac{f_B(\x_i)}{f_A(\x_i)}.\label{eq_relativities}\end{equation}

\added{With $F_n$ the empirical distribution function of the relativities $\text{rel}^{(A,B)}_i$ over all $i\in1,\ldots,n$, the ordered Lorenz curve is defined as}

\begin{equation}
    \left(
    \frac{\sum_{i=1}^{n} f_A(\x_i)\,\mathbbm{1}\left\{F_n\left(\text{rel}^{(A,B)}_i\right)\le s\right\}}{\sum_{i=1}^{n} f_A(\x_i)},
    \frac{\sum_{i=1}^{n} L_i\,\mathbbm{1}\left\{F_n\left(\text{rel}^{(A,B)}_i\right)\le s\right\}}{\sum_{i=1}^{n} L_i}
    \right)\,,
    \label{eq_ord_lorenz}
\end{equation}

\added{over $s\in[0,1]$. When working with frequency-severity models, we replace the predictions $f(\x_i)$ with $f^{\text{freq}}(\x_i)\times f^{\text{sev}}(\x_i)$. Intuitively, the ordered Lorenz curve illustrates the potential improvement in risk classification an insurer can achieve by switching from a model $f_A$ to a model $f_B$. An ordered Lorenz curve below the diagonal line indicates that the new tariff based on model $f_B$ improves risk classification. Conversely, an ordered Lorenz curve above the diagonal line suggests that switching to a tariff based on model $f_B$ results in worse risk classification compared to using model $f_A$.}

\added{\citet{freesgini} extend the Gini-index \citep{gini1912} to ordered Lorenz curves as twice the area between the diagonal line of equality and the ordered Lorenz curve. A large positive Gini index indicates that choosing model $f_B$ over model $f_A$ results in a tariff structure with a better risk classification. A negative Gini index means the ordered Lorenz curve lies above the diagonal, meaning the tariff under model $f_B$ leads to a worse risk classification. Table \ref{tab_gini} presents the Gini indices, comparing the benchmark GLM, GBM, CANN GBM flexible, and the surrogate model. For each row, we highlight the maximum Gini index, signifying that switching from the tariff under model A to the model B with the highest Gini index results in the most significant improvement in risk classification. For instance, the Gini indices for the Australian data in Table \ref{tab_gini_AUS} show that with the benchmark GLM as model A, each model shows an improvement in risk classification, but the improvement is greatest when switching to the tariff under the CANN GBM flexible.}

\added{We apply the min-max strategy from \citet{freesgini}, selecting model A, which yields the lowest maximum Gini index, indicating the least potential for improvement in risk classification. For the Australian, Belgian, and French data sets, the min-max strategy selects the tariff structure under the CANN GBM flexible model. For the Norwegian data, the benchmark GLM is chosen. When comparing the benchmark GLM to the surrogate GLM, the Australian tariff structure under the benchmark GLM is preferred, whereas, for the Belgian and French data, the surrogate GLM is favored. This preference aligns with the insights from the Murphy diagrams in Figure \ref{fig_murphy_surr} and the Diebold-Mariano test results in Table \ref{tab_DB_surr} for the Australian and French data. However, for the Belgian data, while Murphy diagram and Diebold-Mariano tests show none of the GLMs has predictive dominance over the other, combining the frequency and severity models in a tariff structure leads to a more favorable outcome under the surrogate GLMs according to the Gini index.}

\setlength{\extrarowheight}{3pt} % a bit of extra white space between lines in tables
\begin{table}[ht!]
 \begin{adjustwidth}{-1cm}{-1cm}
  \centering
  \scriptsize
  \begin{subtable}[t]{.49\linewidth}
    \begin{NiceTabular}{cr*{4}{r}}[
    code-before = \rowcolor[HTML]{FFFFFF}{1,2,4,6}
              \rowcolor[HTML]{FAFAFF}{3,5}
    ]
    \toprule
    & & \Block{1-4}{\footnotesize \textbf{Model B}} \\[1mm]
    \RowStyle{\rotate}
    & & Benchmark GLM & GBM & CANN GBM flex & Surrogate GLM \\
    \noalign{\hrule height 0.3pt}
    \parbox[t]{2mm}{\multirow{4}{*}{\rotatebox[origin=c]{90}{\footnotesize \textbf{Model A}}}}
    & Benchmark GLM & \phantom{$-00.00$} & $5.30$ & $\mathbf{16.59}$ & $1.90$ \\ 
    & GBM & $3.35$ & \phantom{$-00.00$} & $\mathbf{16.48}$ & $-3.18$ \\ 
    & \ulcolor[blue!25]{CANN GBM flex} & \cellcolor{blue!25} $\mathbf{-6.41}$ & $-10.68$ & \phantom{$-00.00$} & $-10.19$  \\ 
    & Surrogate GLM & $10.82$ & $12.48$ & $\mathbf{21.50}$ & \phantom{$-00.00$} \\ 
    \bottomrule
    \end{NiceTabular}
  \caption{Australian data set}
  \label{tab_gini_AUS}
  \end{subtable} 
  \hfill
  \begin{subtable}[t]{.49\linewidth}
    \begin{NiceTabular}{cr*{4}{r}}[
    code-before = \rowcolor[HTML]{FFFFFF}{1,2,4,6}
              \rowcolor[HTML]{FAFAFF}{3,5}
    ]
    \toprule
    & & \Block{1-4}{\footnotesize \textbf{Model B}} \\[1mm]
    \RowStyle{\rotate}
    & & Benchmark GLM & GBM & CANN GBM flex & Surrogate GLM \\
    \noalign{\hrule height 0.3pt}
    \parbox[t]{2mm}{\multirow{4}{*}{\rotatebox[origin=c]{90}{\footnotesize \textbf{Model A}}}}
    & Benchmark GLM & \phantom{$-00.00$} & $6.59$ & $\mathbf{10.49}$ & $6.37$ \\ 
    & GBM & $3.86$ & \phantom{$-00.00$} & $\mathbf{9.71}$ & $2.69$ \\ 
    & \ulcolor[blue!25]{CANN GBM flex} & \cellcolor{blue!25} $\mathbf{0.14}$ & $-6.48$ & \phantom{$-00.00$} & $-1.85$  \\ 
    & Surrogate GLM & $5.68$ & $6.83$ & $\mathbf{11.63}$ & \phantom{$-00.00$} \\ 
    \bottomrule
    \end{NiceTabular}
  \caption{Belgian data set}
  \label{tab_gini_BE}
  \end{subtable} 
\newline
\vspace*{1 cm}
\newline
  \begin{subtable}[t]{.49\linewidth}
    \begin{NiceTabular}{cr*{4}{r}}[
    code-before = \rowcolor[HTML]{FFFFFF}{1,2,4,6}
              \rowcolor[HTML]{FAFAFF}{3,5}
    ]
    \toprule
    & & \Block{1-4}{\footnotesize \textbf{Model B}} \\[1mm]
    \RowStyle{\rotate}
    & & Benchmark GLM & GBM & CANN GBM flex & Surrogate GLM \\
    \noalign{\hrule height 0.3pt}
    \parbox[t]{2mm}{\multirow{4}{*}{\rotatebox[origin=c]{90}{\footnotesize \textbf{Model A}}}}
    & Benchmark GLM & \phantom{$-00.00$} & $26.82$ & $\mathbf{28.08}$ & $18.35$ \\ 
    & GBM & $-2.48$ & \phantom{$-00.00$} & $\mathbf{6.34}$ & $-0.23$ \\ 
    & \ulcolor[blue!25]{CANN GBM flex} & $-4.60$ & \cellcolor{blue!25} $\mathbf{5.00}$ & \phantom{$-00.00$} & $-2.90$  \\ 
    & Surrogate GLM & $4.55$ & $19.57$ & $\mathbf{21.04}$ & \phantom{$-00.00$} \\ 
    \bottomrule
    \end{NiceTabular}
  \caption{French data set}
  \label{tab_gini_FR}
  \end{subtable} 
  \hfill
  \begin{subtable}[t]{.49\linewidth}
    \begin{NiceTabular}{cr*{3}{r}c}[
    code-before = \rowcolor[HTML]{FFFFFF}{1,2,4,6}
              \rowcolor[HTML]{FAFAFF}{3,5}
    ]
    \toprule
    & & \Block{1-4}{\footnotesize \textbf{Model B}} \\[1mm]
    \RowStyle{\rotate}
    & & Benchmark GLM & GBM & CANN GBM flex & \textcolor{lgray}{Surrogate GLM} \\
    \noalign{\hrule height 0.3pt}
    \parbox[t]{2mm}{\multirow{4}{*}{\rotatebox[origin=c]{90}{\footnotesize \textbf{Model A}}}}
    & \ulcolor[blue!25]{Benchmark GLM} & \phantom{$-00.00$} & \cellcolor{blue!25}$\mathbf{1.12}$ & $0.45$ & \textcolor{lgray}{-} \\ 
    & GBM & $\mathbf{1.98}$ & \phantom{$-00.00$} & $-1.02$ & \textcolor{lgray}{-} \\ 
    & CANN GBM flex & $\mathbf{3.10}$ & $2.65$ &  \phantom{$-00.00$} & \textcolor{lgray}{-} \\ 
    & \textcolor{lgray}{Surrogate GLM} & \textcolor{lgray}{-} & \textcolor{lgray}{-} & \textcolor{lgray}{-} & \textcolor{lgray}{\phantom{$0.0$}-\phantom{$0.0$}} \\ 
    \bottomrule
    \end{NiceTabular}
  \caption{Norwegian data set}
  \label{tab_gini_NOR}
  \end{subtable} 
  \end{adjustwidth}
 \caption{Gini indexes comparing the tariff structures between the benchmark GLM, GBM, CANN GBM flexible and the surrogate GLM. The Gini index is calculated as twice the area under the ordered Lorenz curve for model B relative to model A, showing the possible improvements in tariff structure when switching from pricing under model A to model B. A negative Gini index means the ordered Lorenz curve lies above the diagonal, showing that the tariff structure under model B leads to a worse risk classification than the tariff under model A. For each row, the maximum Gini index is given in bold. The min-max strategy selects model A with the lowest maximum Gini-index, shown by the highlighted cells and underlined model.}
  \label{tab_gini}
\end{table}

\section{Conclusion}\label{conclusion}

This paper explores the potential of deep learning models for the analysis of tabular frequency and severity data in non-life insurance pricing. We detail a benchmark study using extensive cross-validation on multiple model architectures. Categorical input features are embedded by use of an autoencoder of which the encoder part is integrated into the deep learning structures. Our results demonstrate the performance gains achieved using the autoencoder embedding technique for categorical input variables, especially in modeling claim severities. Interpretation tools are applied to both frequency and severity modeling. \added{Next to comparing the performance of models based on out-of-sample deviance, we analyze the prediction dispersion and calibration, and we use Murphy diagrams and Diebold-Mariano tests to see whether a model with lower deviance has predictive dominance and/or is a significant improvement. Together, these techniques provide a robust and reliable model evaluation framework, allowing assessment of whether certain modeling techniques have an added benefit compared to industry-standard techniques.}

The literature often questions the value created when analyzing tabular data with deep learning models. Indeed, our feed-forward neural network does not improve upon a carefully designed \added{GLM} or GBM. Combining gradient-boosted trees and neural networks leads to a lower out-of-sample deviance when modeling claim frequencies. This aligns with what we see in other fields, where GBM and neural network combinations outperform the corresponding stand-alone models on tabular data. \added{However, using Murphy diagrams and the Diebold-Mariano tests, we see the (slightly) lower out-of-sample deviance is not always significant and does not always lead to predictive dominance. These results show the importance of a robust model evaluation framework, going beyond the comparison on out-of-sample deviance.} In modeling the severity data, out-of-sample deviances are relatively similar across benchmark models and deep learning architectures. This suggests that the added value created by using the deep learning approach is limited when applied to these datasets. Using interpretation tools applied to the deep learning models, we create a GLM that is carefully designed as a global surrogate for the deep learning model. Hence, this surrogate GLM leverages the insights from the deep learning models. The workflow to construct a GLM as a global surrogate for a deep learning model can potentially be of interest to insurance companies aiming to harvest refined insights from their available data while aiming for an interpretable and explainable model. The latter consideration is of much interest in light of the GDPR algorithmic accountability clause.

\added{The end result of our study is a comparison of the technical tariffs constructed from our benchmark and deep learning models based on (ordered) Lorenz curves and Gini indices. Based on the model evaluation framework for the frequency and severity models and the comparison of the technical tariffs, the CANN GBM flexible model is most often the preferred model architecture, except \addedtwo{on the Norwegian data set, which only contains a few input variables.} When comparing the benchmark GLM with the surrogate GLM, the surrogate leads to the highest Gini index, meaning it has an improved risk classification compared to the benchmark GLM. These findings are particularly valuable for actuaries, as they show that not only can machine learning models offer higher predictive accuracy, but they can also be adapted into interpretable and explainable surrogate models, making them suitable for practical implementation.}

Data sets with a high dimensional set of input variables and/or complex input features might benefit more from using deep learning models to model claim frequency and severity data. \addedtwo{This aligns with our observation that deep learning models show limited improvement on the Norwegian dataset, which consists of only four input variables.} \addedtwo{\citet{GAO2022185}, for instance, use deep learning models to analyze high-frequency time series of telematics data, and \citet{Blier-Wong2024-wj} use convolutional neural networks to embed images for pricing with GLMs.} Further research could look into data sets with high dimensional feature sets, including features with images, text or times series, and explore the value of deep learning architectures via a carefully designed benchmark study, \added{with an extensive model evaluation framework}, as outlined in this paper.  

\section*{Acknowledgements}

Katrien Antonio gratefully acknowledges funding from the FWO and Fonds De La Recherche Scientifique - FNRS (F.R.S.-FNRS) under the Excellence of Science (EOS) program, project ASTeRISK Research Foundation Flanders [grant number $40007517$]. The authors gratefully acknowledge support from the Ageas research chair on insurance analytics at KU Leuven, from the Chaire DIALog sponsored by CNP Assurances and the FWO network W$001021$N. The authors thank Simon Gielis for his contributions (as MSc student) in the early stage of the research. We also express our gratitude to the editor and reviewers for their valuable feedback, which has greatly enhanced the quality of this article.
%\emph{ANONYMOUS}

\section*{Declaration of interest statement}

The authors declare no potential conflicts of interest. 

\section*{Supplemental material}

The results in this paper were obtained using R. All code is available through Github: \url{https://github.com/freekholvoet/NNforFreqSevPricing}. 
%\emph{ANONYMOUS}
An R Markdown demonstration is available on that GitHub page, called NNforFreqSevPricing.nb.html.

\bibliographystyle{plainnat}
\small
\bibliography{20220715_Bibliography}

\newpage
\appendix
\section{Data sets}\label{app_datasets}

Tables for each of the four data sets with the variable names and their explanations.

\setlength{\extrarowheight}{3pt} % a bit of extra white space between lines in tables
\begin{table}[ht!]
	\begin{adjustwidth}{-1cm}{-1cm}
		\centering
		\scriptsize
		\begin{subtable}[t]{.49\linewidth}
			\begin{NiceTabular}[t]{m{1.4cm}m{5cm}m{1.3cm}}[
					code-before = \rowcolor[HTML]{FFFFFF}{1,3,5}
					\rowcolor[HTML]{FAFAFF}{2,4,6}
				]
				\toprule
				\textbf{Variable} & \textbf{Description} & \textbf{Type}\\
				\noalign{\hrule height 0.3pt}
				VehValue & The vehicle value in thousands of AUD. & Cont. \\
				VehAge & The vehicle age group. & Cat. (4)(4)  \\
				VehBody & The vehicle body group. & Cat. (13)(13)  \\
				Gender & The gender of the policyholder. & Cat. (2)(2)  \\
                    DrivAge & The age of the policyholder. & Cat. (6)(6)  \\
				\bottomrule
			\end{NiceTabular}
			\caption{Australian data set, as defined in \texttt{CASdatasets} \citep{casdatasets}.}
			\label{tab_vars_AUS}
		\end{subtable} 
		\hfill
		\begin{subtable}[t]{.49\linewidth}
			\begin{NiceTabular}[t]{m{1.4cm}m{5cm}m{1.3cm}}[
					code-before = \rowcolor[HTML]{FFFFFF}{1,3,5,7,9}
					\rowcolor[HTML]{FAFAFF}{2,4,6,8,10}
				]
				\toprule
				\textbf{Variable} & \textbf{Description} & \textbf{Type}\\
				\noalign{\hrule height 0.3pt}
                    ageph & Policyholder age in years. & Cont. \\
				power & Horsepower of the vehicle in kilowatts. & Cont. \\
				agec & Vehicle age in years. & Cont. \\
				bm & Belgian bonus-malus score of the policyholder. & Cont. \\
				coverage & Type of coverage. & Cat. $(3)$  \\
				fuel & Fuel type of the vehicle. & Cat. $(2)$  \\
				sex & Policyholder gender. & Cat. $(2)$  \\
				use & Main use of the vehicle, private or work. & Cat. $(2)$  \\
                    fleet & Whether or not the vehicle is part of a fleet. & Cat. $(2)$  \\
                    postalcode & Postal code of the municipality of residence of the policyholder. Replaced by latitude and longitude coordinate of the center of the municipality. & Cont. \\
				\bottomrule
			\end{NiceTabular}
			\caption{Belgian data set, as defined in \citet{Henckaerts2018}.}
			\label{tab_vars_BE}
		\end{subtable} 
		\newline
		\vspace*{1 cm}
		\newline
		\begin{subtable}[t]{.49\linewidth}
			\begin{NiceTabular}[t]{m{1.4cm}m{5cm}m{1.3cm}}[
					code-before = \rowcolor[HTML]{FFFFFF}{1,3,5,7,9}
					\rowcolor[HTML]{FAFAFF}{2,4,6,8,10}
				]
				\toprule
				\textbf{Variable} & \textbf{Description} & \textbf{Type}\\
				\noalign{\hrule height 0.3pt}
				VehPower & The power of the car. & Cat. (12)(12)  \\
				VehAge & The vehicle age. & Cat. $(3)$  \\
				DrivAge & The driver age, grouped. & Cat. $(7)$  \\
				BonusMalus & French bonus-malus score of the policyholder. & Cont. \\
                    VehBrand  & Car brand grouped in unknown categories. & Cat. $(11)$  \\
                    VehGas  & The car gas, Diesel or regular. & Cat. $(2)$  \\
                    Area  & Area type of the city where the policyholder lives in six groups, from `A' for rural areas to `F' for urban centers. & Cat. $(6)$  \\
                    Density  & The numerical population density per square kilometer of inhabitants of the city where the policyholder lives. & Cont. \\
                    Region  & The policy region in France (based on the 1970-2015 classification). & Cat. $(21)$  \\
				\bottomrule
			\end{NiceTabular}
			\caption{French data set, from \citet{Schelldorfer2019}.}
			\label{tab_vars_FR}
		\end{subtable} 
		\hfill
		\begin{subtable}[t]{.49\linewidth}
			\begin{NiceTabular}[t]{m{1.4cm}m{5cm}m{1.3cm}}[
					code-before = \rowcolor[HTML]{FFFFFF}{1,3,5}
					\rowcolor[HTML]{FAFAFF}{2,4}
				]
				\toprule
				\textbf{Variable} & \textbf{Description} & \textbf{Type}\\
				\noalign{\hrule height 0.3pt}
				Male & Binary variable with value one if the policyholder is male and zero otherwise. & Cat. $(2)$  \\
				Young & Binary variable with value one if the policyholder is less than $26$ years of age and zero otherwise. & Cat. $(2)$  \\
				DistLimit & Maximum number of kilometers per year allowed as stated in the insurance contract. There are six different limits possible. & Cat. $(6)$  \\
				GeoRegion & Density of the geographical region binned in six groups. & Cat. $(6)$  \\
				\bottomrule
			\end{NiceTabular}
			\caption{Norwegian data set, as defined in \texttt{CASdatasets} \citep{casdatasets}.}
			\label{tab_vars_NOR}
		\end{subtable} 		
	\end{adjustwidth}
	\caption{Overview of the variables and their meaning in the four data sets used in the benchmark study. \addedtwo{The column \emph{Type} shows whether the variable is continuous (Cont.) or categorical (Cat.). For categorical variables, we note the number of levels between brackets.}}
	\label{tab_allvariables}
\end{table}

\section{Numerical out-of-sample Poisson and gamma deviances}\label{app_oostable}

We show the numerical values behind the out-of-sample deviances from Figure \ref{fig_oos_freq} in Table \ref{tab_oosnum_freq} for the claim frequency models and Table \ref{tab_oosnum_sev} for the claim severity models.

\setlength{\extrarowheight}{3pt} % a bit of extra white space between lines in tables
\begin{table}[ht!]
 \begin{adjustwidth}{-1cm}{-1cm}
  \centering
  \footnotesize
    \begin{NiceTabular}{cr*{6}{r}}[
    code-before = \rowcolor[HTML]{FFFFFF}{1,2,4,6,8,10,12,14,16,18,20,22,24,26,28,30}
              \rowcolor[HTML]{FAFAFF}{3,5,7,9,11,13,15,17,19,21,23,25,27,29}
    ]
    \toprule
    & & \Block{1-6}{\footnotesize \textbf{Test fold}} \\[1mm]
    & & $\mathcal{D}_1$ & $\mathcal{D}_2$ & $\mathcal{D}_3$ & $\mathcal{D}_4$ & $\mathcal{D}_5$ & $\mathcal{D}_6$ \\
    \noalign{\hrule height 0.3pt}
    \parbox[t]{2mm}{\multirow{7}{*}{\rotatebox[origin=c]{90}{\footnotesize \textbf{Australia}}}}
 & GLM & $0.3816$ & $0.3797$ & $0.3648$ & $0.3704$ & $0.3735$ & $0.3770$ \\  
 & GBM & $0.3804$ & $0.3800$ & $0.3644$ & $0.3701$ & $0.3732$ & $0.3761$ \\          
 & FFNN & $0.3816$ & $0.3794$ & $0.3657$ & $0.3748$ & $0.3748$ & $0.3769$ \\          
 & CANN GLM fixed & $0.3820$ & $0.3800$ & $0.3647$ & $0.3709$ & $0.3730$ & $0.3769$ \\
 & CANN GLM flex & $0.3793$ & $0.3786$ & $0.3635$ & $0.3700$ & $0.3717$ & $0.3750$ \\
 & CANN GBM fixed & $0.3805$ & $0.3800$ & $0.3644$ & $0.3701$ & $0.3731$ & $0.3761$ \\
 & CANN GBM flex & $0.3782$ & $0.3781$ & $0.3639$ & $0.3700$ & $0.3719$ & $0.3737$ \\
     \noalign{\hrule height 0.3pt}
    \parbox[t]{2mm}{\multirow{7}{*}{\rotatebox[origin=c]{90}{\footnotesize \textbf{Belgium}}}}
 & GLM & $0.5314$ & $0.5341$ & $0.5332$ & $0.5323$ & $0.5292$ & $0.5313$ \\   
 & GBM & $0.5295$ & $0.5316$ & $0.5308$ & $0.5299$ & $0.5284$ & $0.5295$ \\          
 & FFNN & $0.5319$ & $0.5351$ & $0.5322$ & $0.5327$ & $0.5325$ & $0.5317$ \\           
 & CANN GLM fixed & $0.5307$ & $0.5343$ & $0.5324$ & $0.5321$ & $0.5294$ & $0.5307$ \\
 & CANN GLM flex & $0.5283$ & $0.5324$ & $0.5293$ & $0.5297$ & $0.5277$ & $0.5288$ \\ 
 & CANN GBM fixed & $0.5295$ & $0.5316$ & $0.5308$ & $0.5299$ & $0.5284$ & $0.5295$ \\
 & CANN GBM flex & $0.5279$ & $0.5306$ & $0.5291$ & $0.5287$ & $0.5276$ & $0.5291$ \\ 
    \noalign{\hrule height 0.3pt}
    \parbox[t]{2mm}{\multirow{7}{*}{\rotatebox[origin=c]{90}{\footnotesize \textbf{France}}}}
 & GLM & $0.2762$ & $0.2386$ & $0.2450$ & $0.2455$ & $0.2421$ & $0.2186$ \\    
 & GBM & $0.2714$ & $0.2351$ & $0.2417$ & $0.2415$ & $0.2384$ & $0.2158$ \\           
 & FFNN & $0.2706$ & $0.2355$ & $0.2413$ & $0.2415$ & $0.2399$ & $0.2156$ \\            
 & CANN GLM fixed & $0.2765$ & $0.2383$ & $0.2437$ & $0.2447$ & $0.2417$ & $0.2184$ \\
 & CANN GLM flex & $0.2743$ & $0.2362$ & $0.2429$ & $0.2434$ & $0.2396$ & $0.2164$ \\ 
 & CANN GBM fixed & $0.2711$ & $0.2353$ & $0.2417$ & $0.2415$ & $0.2385$ & $0.2160$ \\
 & CANN GBM flex & $0.2695$ & $0.2341$ & $0.2406$ & $0.2401$ & $0.2373$ & $0.2148$ \\ 
    \noalign{\hrule height 0.3pt}
    \parbox[t]{2mm}{\multirow{7}{*}{\rotatebox[origin=c]{90}{\footnotesize \textbf{Norway}}}}
 & GLM & $0.2779$ & $0.2723$ & $0.2700$ & $0.2672$ & $0.2708$ & $0.2624$ \\    
 & GBM & $0.2778$ & $0.2724$ & $0.2699$ & $0.2671$ & $0.2708$ & $0.2624$ \\           
 & FFNN & $0.2799$ & $0.2742$ & $0.2709$ & $0.2688$ & $0.2728$ & $0.2640$ \\            
 & CANN GLM fixed & $0.2778$ & $0.2723$ & $0.2698$ & $0.2672$ & $0.2708$ & $0.2624$ \\
 & CANN GLM flex & $0.2779$ & $0.2723$ & $0.2698$ & $0.2672$ & $0.2708$ & $0.2624$ \\ 
 & CANN GBM fixed & $0.2777$ & $0.2724$ & $0.2699$ & $0.2671$ & $0.2708$ & $0.2624$ \\
 & CANN GBM flex & $0.2778$ & $0.2724$ & $0.2699$ & $0.2671$ & $0.2708$ & $0.2624$ \\ 
    \bottomrule
    \end{NiceTabular}
  \end{adjustwidth}
 \caption{Numerical values behind the left-hand side of Figure \ref{fig_oos_freq}, showing the out-of-sample deviances for the claim frequency models on all four data sets.}
  \label{tab_oosnum_freq}
\end{table}

\setlength{\extrarowheight}{3pt} % a bit of extra white space between lines in tables
\begin{table}[ht!]
 \begin{adjustwidth}{-1cm}{-1cm}
  \centering
  \footnotesize
    \begin{NiceTabular}{cr*{6}{r}}[
    code-before = \rowcolor[HTML]{FFFFFF}{1,2,4,6,8,10,12,14,16,18,20,22,24,26,28,30}
              \rowcolor[HTML]{FAFAFF}{3,5,7,9,11,13,15,17,19,21,23,25,27,29}
    ]
    \toprule
    & & \Block{1-6}{\footnotesize \textbf{Test fold}} \\[1mm]
    & & $\mathcal{D}_1$ & $\mathcal{D}_2$ & $\mathcal{D}_3$ & $\mathcal{D}_4$ & $\mathcal{D}_5$ & $\mathcal{D}_6$ \\
    \noalign{\hrule height 0.3pt}
    \parbox[t]{2mm}{\multirow{7}{*}{\rotatebox[origin=c]{90}{\footnotesize \textbf{Australia}}}}
 & GLM & $1.5562$ & $1.7005$ & $1.7145$ & $1.6531$ & $1.6745$ & $1.5749$ \\    
 & GBM & $1.5359$ & $1.6864$ & $1.7396$ & $1.6589$ & $1.6355$ & $1.5565$ \\           
 & FFNN & $1.5752$ & $1.7059$ & $1.8833$ & $1.6734$ & $1.6574$ & $1.5877$ \\            
 & CANN GLM fixed & $1.5414$ & $1.7008$ & $1.7101$ & $1.6635$ & $1.6557$ & $1.5772$ \\
 & CANN GLM flex & $1.5508$ & $1.7099$ & $1.7117$ & $1.6561$ & $1.6643$ & $1.5751$ \\ 
 & CANN GBM fixed & $1.5357$ & $1.6905$ & $1.7336$ & $1.6636$ & $1.6356$ & $1.5749$ \\
 & CANN GBM flex & $1.5395$ & $1.6854$ & $1.7329$ & $1.6624$ & $1.6390$ & $1.5573$ \\ 
     \noalign{\hrule height 0.3pt}
    \parbox[t]{2mm}{\multirow{7}{*}{\rotatebox[origin=c]{90}{\footnotesize \textbf{Belgium}}}}
 & GLM & $2.2280$ & $2.2631$ & $2.2937$ & $2.2802$ & $2.2709$ & $2.2508$ \\    
 & GBM & $2.2365$ & $2.2475$ & $2.2845$ & $2.2806$ & $2.2789$ & $2.2290$ \\           
 & FFNN & $2.2436$ & $2.2690$ & $2.2946$ & $2.3010$ & $2.2955$ & $2.2330$ \\            
 & CANN GLM fixed & $2.2284$ & $2.2634$ & $2.2933$ & $2.2813$ & $2.2709$ & $2.2652$ \\
 & CANN GLM flex & $2.2284$ & $2.2637$ & $2.2956$ & $2.2859$ & $2.2711$ & $2.2513$ \\ 
 & CANN GBM fixed & $2.2364$ & $2.2486$ & $2.2852$ & $2.2827$ & $2.2793$ & $2.2290$ \\
 & CANN GBM flex & $2.2365$ & $2.2472$ & $2.2848$ & $2.2824$ & $2.2801$ & $2.2262$ \\ 
    \noalign{\hrule height 0.3pt}
    \parbox[t]{2mm}{\multirow{7}{*}{\rotatebox[origin=c]{90}{\footnotesize \textbf{France}}}}
 & GLM & $1.7083$ & $1.6943$ & $1.8824$ & $2.0947$ & $2.0557$ & $1.4111$ \\    
 & GBM & $1.6471$ & $1.6294$ & $1.8053$ & $2.1109$ & $1.9616$ & $1.3800$ \\           
 & FFNN & $1.6104$ & $1.8199$ & $1.8131$ & $2.5473$ & $2.0442$ & $1.4390$ \\            
 & CANN GLM fixed & $1.7132$ & $1.6973$ & $1.8813$ & $2.0854$ & $2.0334$ & $1.4119$ \\
 & CANN GLM flex & $1.7124$ & $1.6956$ & $1.8877$ & $2.0965$ & $2.0325$ & $1.4143$ \\ 
 & CANN GBM fixed & $1.6472$ & $1.6317$ & $1.8084$ & $2.1101$ & $1.9562$ & $1.3837$ \\
 & CANN GBM flex & $1.7153$ & $1.6784$ & $1.8457$ & $2.2282$ & $1.9652$ & $1.3854$ \\ 
    \noalign{\hrule height 0.3pt}
    \parbox[t]{2mm}{\multirow{7}{*}{\rotatebox[origin=c]{90}{\footnotesize \textbf{Norway}}}}
 & GLM & $1.1355$ & $0.9900$ & $1.0667$ & $1.0727$ & $1.0076$ & $1.0285$ \\    
 & GBM & $1.1370$ & $0.9932$ & $1.0694$ & $1.0736$ & $1.0077$ & $1.0277$ \\           
 & FFNN & $1.1353$ & $0.9898$ & $1.0657$ & $1.0739$ & $1.0080$ & $1.0417$ \\            
 & CANN GLM fixed & $1.1373$ & $0.9888$ & $1.0695$ & $1.0731$ & $1.0082$ & $1.0284$ \\
 & CANN GLM flex & $1.1358$ & $0.9897$ & $1.0717$ & $1.0722$ & $1.0085$ & $1.0284$ \\ 
 & CANN GBM fixed & $1.1374$ & $0.9918$ & $1.0695$ & $1.0734$ & $1.0074$ & $1.0281$ \\
 & CANN GBM flex & $1.1378$ & $0.9923$ & $1.0693$ & $1.0731$ & $1.0086$ & $1.0283$ \\ 
    \bottomrule
    \end{NiceTabular}
  \end{adjustwidth}
 \caption{Numerical values behind the right-hand side of Figure \ref{fig_oos_freq}, showing the out-of-sample deviances for the claim severity models on all four data sets.}
  \label{tab_oosnum_sev}
\end{table}

\section{Model evaluation framework applied on the claim severity models}\label{app_sev_evalframe}

\begin{figure}[ht!]
\begin{adjustwidth}{-1.6cm}{-1.2cm}
\centering
  \includegraphics[width = \linewidth]{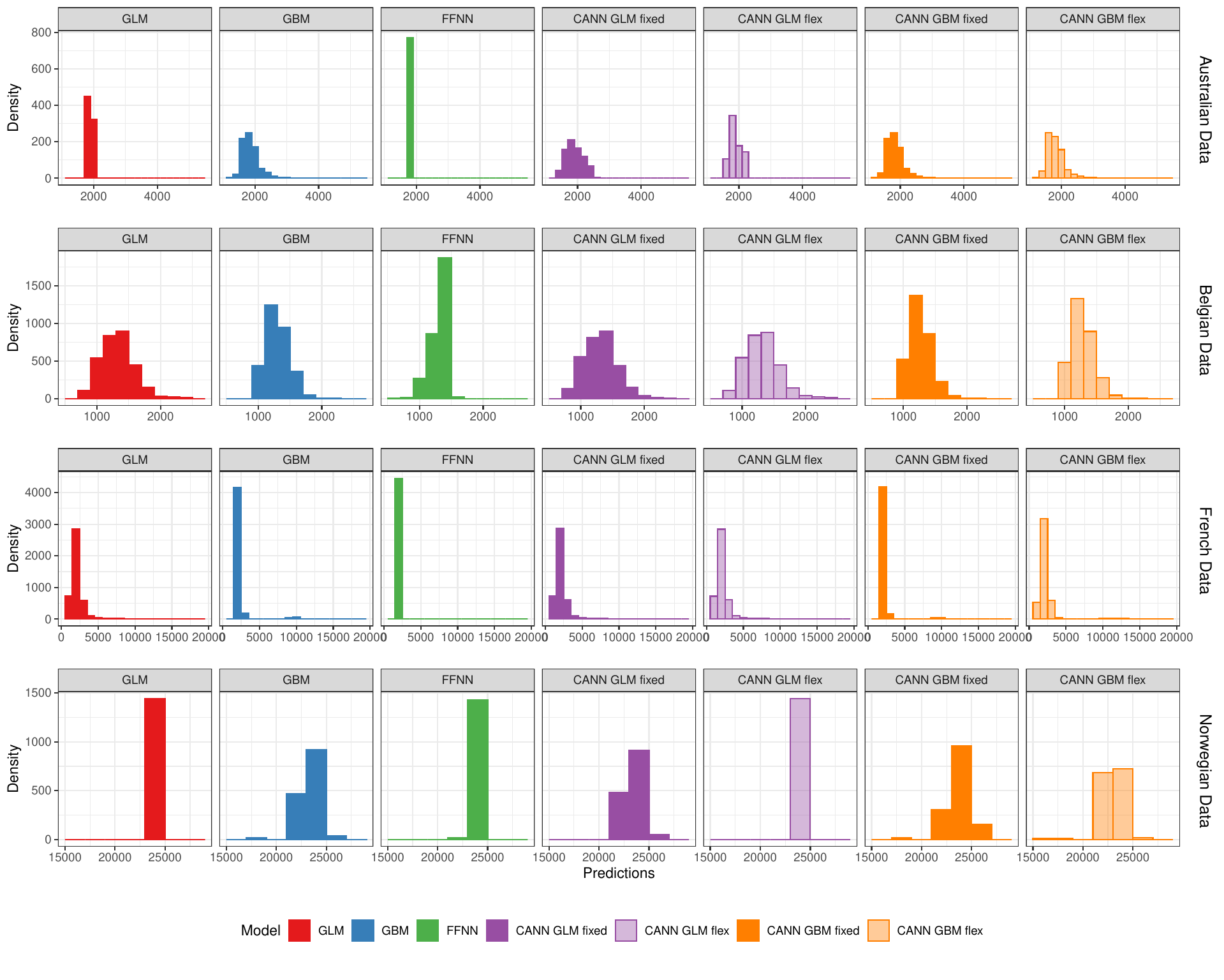}
\end{adjustwidth}
\caption{Histogram of predictions for the test set $\mathcal{D}_1$ obtained from each severity model, illustrating the distribution and concentration of predictions for each model to identify differences in prediction patterns and potential outliers.}
\label{fig_predhisto_comp_sev}
\end{figure}

\begin{figure}[ht!]
\begin{adjustwidth}{-1.6cm}{-1.2cm}
\centering
  \includegraphics[width = \linewidth]{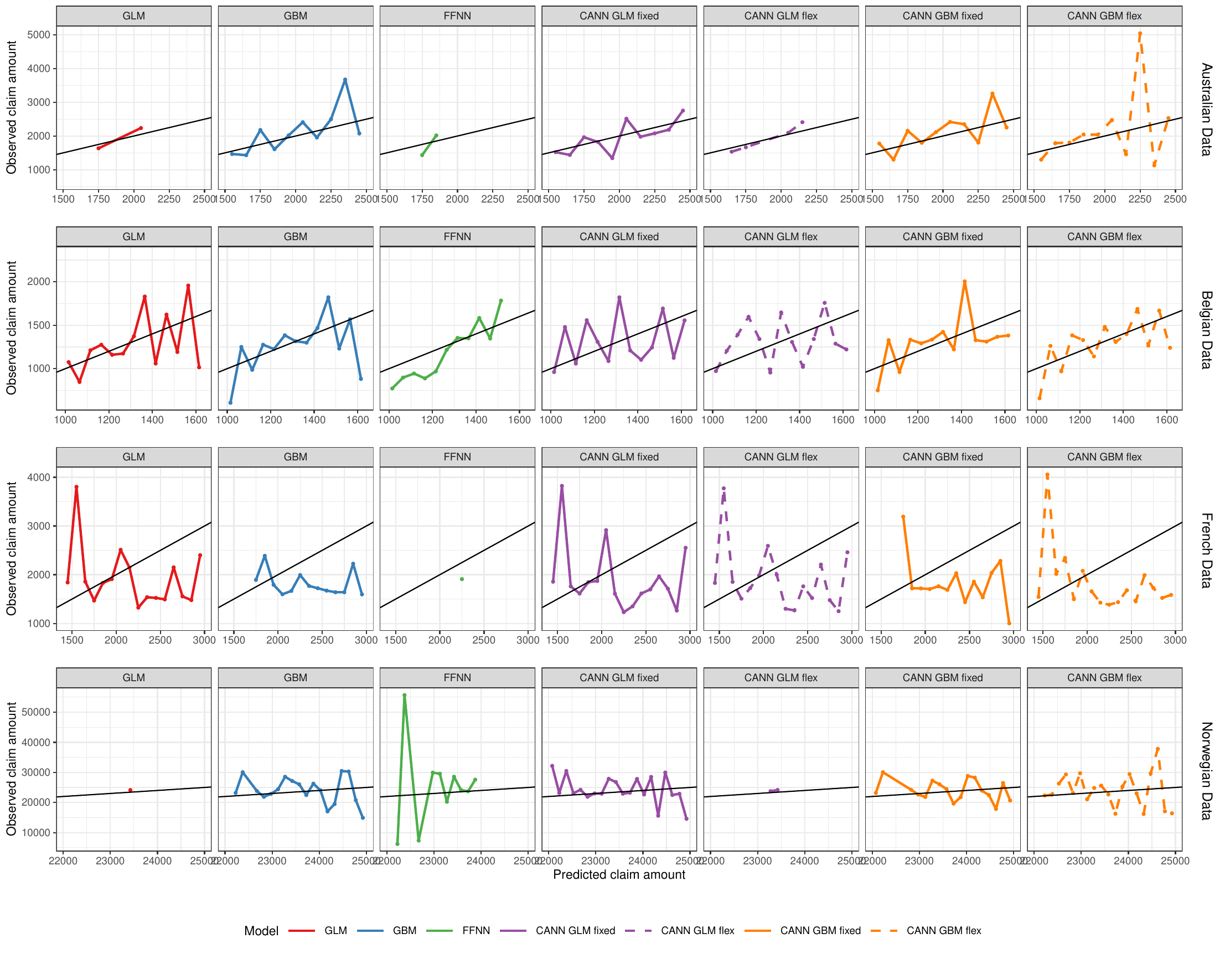}
\end{adjustwidth}
\caption{Plot of $\E \left[y_i | f(\x_i) = s \right]$ over the binned range of predictions $s$ for each severity model type. Predictions are made on test set $\mathcal{D}_1$. A line above (below) the diagonal indicates that the model is underestimating (overestimating) the true number of claims in the data. Comparison between models allows us to assess how well the predicted values align with observed outcomes.}
\label{fig_exp_response_sev}
\end{figure}

\begin{figure}[ht!]
 \begin{adjustwidth}{-1.4cm}{-1.4cm}
 \centering
 \includegraphics[width = 1\linewidth]{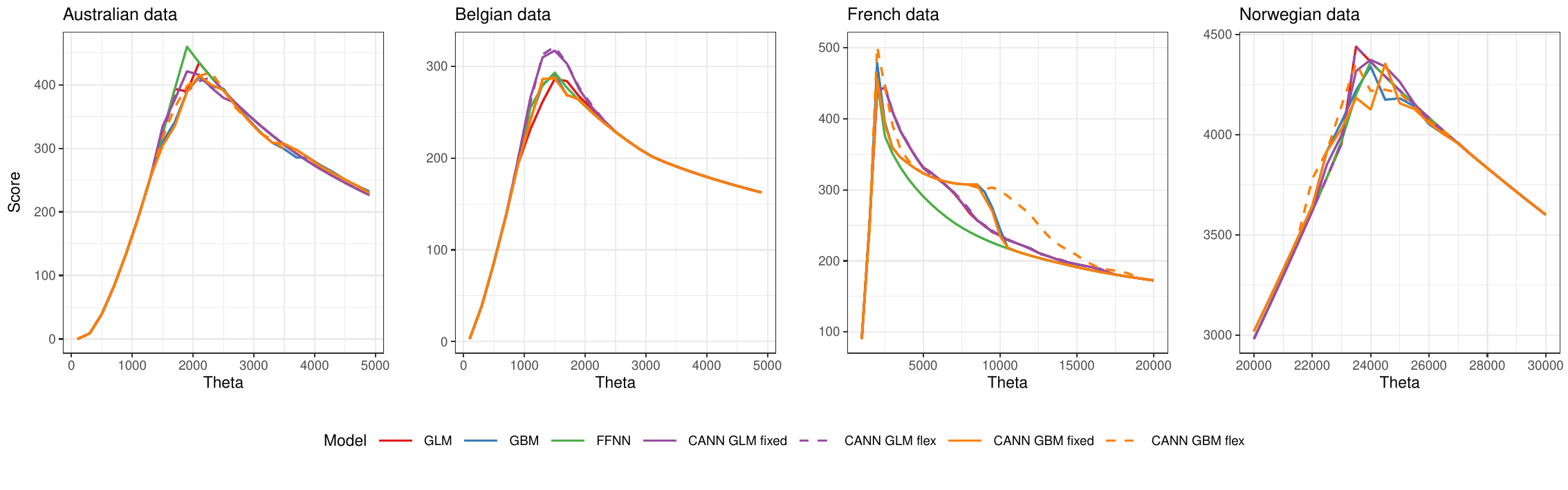}
 \end{adjustwidth}
\caption{Murphy diagram for the severity models on test set $\mathcal{D}_1$ for each data set to compare the ranking of models across a range of scoring functions. We plot the losses for each model according to the elementary scoring functions $S_\theta$ over a range of values for the parameter $\theta$.}
\label{fig_murphy_sev}
\end{figure}

Next to the out-of-sample gamma deviances in Figure \ref{fig_oos_freq}, we apply the other model evaluation techniques from Section \ref{sec_evalframe} to the claim severity models. For each of the four data sets, the results of the Diebold-Mariano tests are given in Table \ref{tab_DB_sev}. For the Australian data, the null hypothesis is rejected in favor of the CANN GBM models when compared to the CANN GLM fixed and FFNN models but not when compared to the benchmark GLM and GBM. For the Belgian data, the null hypothesis is always rejected when comparing either CANN GLM model with other model architectures, meaning the other model architects are seen as statistically more accurate than the CANN GLM models when modeling claim severity. For the French data set, the null hypothesis is most often rejected in favor of the GBM and CANN GBM fixed but cannot be rejected when comparing those to each other, meaning both seem suitable choices for modeling claim severity on the French data set. On the Norwegian data set, we see that the null hypothesis is never rejected, meaning each model is seen as having statistically similar accuracy. Figure \ref{fig_predhisto_comp_sev} shows the prediction histograms for each severity model, and Figure \ref{fig_exp_response_sev} shows the graph of $\E \left[Y | f(\X) = s \right]$ over the binned range of predictions $s$. We note that these techniques are more difficult to draw conclusions from, as the smaller severity data sets lead to more concentrated dispersions and more volatile calibration plots. The Murphy diagrams are shown in Figure \ref{fig_murphy_sev}. We see on the Belgian data set, that each model has predictive dominance over both CANN GLM models, which aligns with the findings of the Diebold-Mariano test. For the other data sets, no model seems to have clear predictive dominance over the others.

\setlength{\extrarowheight}{3pt} % a bit of extra white space between lines in tables
\begin{table}[ht!]
 \begin{adjustwidth}{-0.6cm}{-0.6cm}
  \centering
  \scriptsize
  \begin{subtable}[t]{.49\linewidth}
    \begin{NiceTabular}{cr*{7}{r}}[
    code-before = \rowcolor[HTML]{FFFFFF}{1,2,4,6,8}
              \rowcolor[HTML]{FAFAFF}{3,5,7,9}
    ]
    \toprule
    & & \Block{1-7}{\footnotesize \textbf{Model B}} \\[1mm]
    \RowStyle{\rotate}
    & & GLM & GBM & FFNN & CANN GLM fixed & CANN GLM flex & CANN GBM fixed & CANN GBM flex\\
    \noalign{\hrule height 0.3pt}
    \parbox[t]{2mm}{\multirow{7}{*}{\rotatebox[origin=c]{90}{\footnotesize \textbf{Model A}}}}
    & GLM &  & \cmark & \cmark & \cmark & \xmark & \cmark & \cmark \\ 
    & GBM & \cmark & & \cmark & \cmark & \cmark & \cmark &  \cmark \\ 
    & FFNN & \cmark & \xmark &  & \cmark & \cmark & \xmark & \cellcolor{blue!25}\xmark \\ 
    & CANN GLM fixed & \cmark & \cmark & \cmark & & \xmark & \xmark & \cellcolor{blue!25}\xmark \\ 
    & CANN GLM flex & \cmark & \cmark & \cmark & \cmark &  & \cmark & \cmark \\ 
    & CANN GBM fixed & \cmark & \cmark & \cmark & \cmark & \cmark &  & \cellcolor{blue!25}\xmark \\ 
    & CANN GBM flex & \cmark & \cmark & \cmark & \cmark & \cmark & \cmark & \\ 
    \bottomrule
    \end{NiceTabular}
  \caption{Australian data set}
  \label{tab_DB_sev_AUS}
  \end{subtable} 
  \hfill
  \begin{subtable}[t]{.49\linewidth}
    \begin{NiceTabular}{cr*{7}{r}}[
    code-before = \rowcolor[HTML]{FFFFFF}{1,2,4,6,8}
              \rowcolor[HTML]{FAFAFF}{3,5,7,9}
    ]
    \toprule
    & & \Block{1-7}{\footnotesize \textbf{Model B}} \\[1mm]
    \RowStyle{\rotate}
    & & GLM & GBM & FFNN & CANN GLM fixed & CANN GLM flex & CANN GBM fixed & CANN GBM flex\\
    \noalign{\hrule height 0.3pt}
    \parbox[t]{2mm}{\multirow{7}{*}{\rotatebox[origin=c]{90}{\footnotesize \textbf{Model A}}}}
    & GLM &  & \cmark & \cmark & \cmark & \cmark & \cmark & \cmark \\ 
    & GBM & \cmark & & \cmark & \cmark & \cmark & \cmark &  \cmark \\ 
    & FFNN & \cmark & \cmark &  & \cmark & \cmark & \cmark & \cmark \\ 
    & CANN GLM fixed & \cellcolor{blue!25}\xmark & \cellcolor{blue!25}\xmark & \cellcolor{blue!25}\xmark & & \cmark & \cellcolor{blue!25}\xmark & \cellcolor{blue!25}\xmark \\ 
    & CANN GLM flex & \cellcolor{blue!25}\xmark & \cellcolor{blue!25}\xmark & \cellcolor{blue!25}\xmark & \cmark &  & \cellcolor{blue!25}\xmark & \cellcolor{blue!25}\xmark \\ 
    & CANN GBM fixed & \cmark & \cmark & \cmark & \cmark & \cmark &  & \cmark \\ 
    & CANN GBM flex & \cmark & \cmark & \cmark & \cmark & \cmark & \cmark & \\ 
    \bottomrule
    \end{NiceTabular}
  \caption{Belgian data set}
  \label{tab_DB_sev_BE}
  \end{subtable} 
\newline
\vspace*{1 cm}
\newline
  \begin{subtable}[t]{.49\linewidth}
    \begin{NiceTabular}{cr*{7}{r}}[
    code-before = \rowcolor[HTML]{FFFFFF}{1,2,4,6,8}
              \rowcolor[HTML]{FAFAFF}{3,5,7,9}
    ]
    \toprule
    & & \Block{1-7}{\footnotesize \textbf{Model B}} \\[1mm]
    \RowStyle{\rotate}
    & & GLM & GBM & FFNN & CANN GLM fixed & CANN GLM flex & CANN GBM fixed & CANN GBM flex\\
    \noalign{\hrule height 0.3pt}
    \parbox[t]{2mm}{\multirow{7}{*}{\rotatebox[origin=c]{90}{\footnotesize \textbf{Model A}}}}
    & GLM &  & \cellcolor{blue!25}\xmark & \cmark & \cmark & \cmark & \cellcolor{blue!25}\xmark & \cmark \\ 
    & GBM & \cmark & & \cmark & \cmark & \cmark & \cmark &  \cmark \\ 
    & FFNN & \cmark & \cmark &  & \cmark & \cmark & \cmark & \cmark \\ 
    & CANN GLM fixed & \cmark & \cellcolor{blue!25}\xmark & \cmark & & \cmark & \cellcolor{blue!25}\xmark & \cmark \\ 
    & CANN GLM flex & \xmark & \cellcolor{blue!25}\xmark & \cmark & \cmark &  & \cellcolor{blue!25}\xmark & \cmark \\ 
    & CANN GBM fixed & \cmark & \cmark & \cmark & \cmark & \cmark &  & \cmark \\ 
    & CANN GBM flex & \xmark & \cellcolor{blue!25}\xmark & \xmark & \cmark & \cmark & \cellcolor{blue!25}\xmark & \\ 
    \bottomrule
    \end{NiceTabular}
  \caption{French data set}
  \label{tab_DB_sev_FR}
  \end{subtable} 
  \hfill
  \begin{subtable}[t]{.49\linewidth}
    \begin{NiceTabular}{cr*{7}{r}}[
    code-before = \rowcolor[HTML]{FFFFFF}{1,2,4,6,8}
              \rowcolor[HTML]{FAFAFF}{3,5,7,9}
    ]
    \toprule
    & & \Block{1-7}{\footnotesize \textbf{Model B}} \\[1mm]
    \RowStyle{\rotate}
    & & GLM & GBM & FFNN & CANN GLM fixed & CANN GLM flex & CANN GBM fixed & CANN GBM flex\\
    \noalign{\hrule height 0.3pt}
    \parbox[t]{2mm}{\multirow{7}{*}{\rotatebox[origin=c]{90}{\footnotesize \textbf{Model A}}}}
    & GLM &  & \cmark & \cmark & \cmark & \cmark & \cmark & \cmark \\ 
    & GBM & \cmark & & \cmark & \cmark & \cmark & \cmark &  \cmark \\ 
    & FFNN & \cmark & \cmark &  & \cmark & \cmark & \cmark & \cmark \\ 
    & CANN GLM fixed & \cmark & \cmark & \cmark & & \cmark & \cmark & \cmark \\ 
    & CANN GLM flex & \cmark & \cmark & \cmark & \cmark &  & \cmark & \cmark \\ 
    & CANN GBM fixed & \cmark & \cmark & \cmark & \cmark & \cmark &  & \cmark \\ 
    & CANN GBM flex & \cmark & \cmark & \cmark & \cmark & \cmark & \cmark & \\ 
    \bottomrule
    \end{NiceTabular}
  \caption{Norwegian data set}
  \label{tab_DB_sev_NOR}
  \end{subtable} 
  \end{adjustwidth}
 \caption{Results of the Diebold-Mariano test for the severity models on test set $\mathcal{D}_1$ for each data set. The table indicates whether we cannot reject (\cmark) the null hypothesis $H_0: \E\left[\mathscr{L}(f_A(\X),Y) - \mathscr{L}(f_B(\X),Y)\right] = 0$, or if we reject $H_0$ (\xmark) in favor of the alternative hypothesis $H_1: \E\left[\mathscr{L}(f_A(\X),Y) - \mathscr{L}(f_B(\X),Y)\right] > 0$. Highlighted cells indicate columns or rows filled almost exclusively with \xmark\,marks, signifying that the model is either consistently seen as statistically more accurate (column) or that, for this model, the null hypothesis is always rejected in favor of the other models (row).}
  \label{tab_DB_sev}
\end{table}

\section{Comparison between benchmark GLM and surrogate GLM for claim severity modeling}\label{app_sev_surr}

We compare the surrogate GLM based on the severity CANN GBM flexible with the benchmark GLM. Table \ref{tab_sev_surr_all_GLM} shows the selected variables for the benchmark GLM and the surrogate GLM together with the out-of-sample gamma deviance on test set $\mathcal{D}_1$. Only on the surrogate GLM on Australian data has a lower out-of-sample deviance compared to the benchmark GLM. The Diebold-Mariano test cannot reject the null hypothesis in favor of either the benchmark GLM or the surrogate GLM on the Australian and Belgian data. On the French data set, even though the surrogate GLM has a higher out-of-sample deviance, the Diebold-Mariano test shows the surrogate is statistically more accurate than the benchmark GLM. The prediction histogram in Figure \ref{fig_histo_surr} shows that the benchmark GLM and the surrogate GLM have a similar prediction dispersion for the Belgian and French data set, but on the Australian data, the predictions of the benchmark GLM are more concentrated than those of the surrogate GLM. Figure \ref{fig_calib_surr} shows no model is significantly better calibrated than the other on each of the four data sets. However, the calibration plots are more volatile due to the limited severity data sets. The Murhpy diagram in Figure \ref{fig_murphy_surr} shows that neither model has predictive dominance over the other for all $\theta$. 

\begin{table}[ht!]
\scriptsize
\begin{adjustwidth}{-1.5cm}{-1.5cm}
\begin{subtable}[t]{.31\linewidth}
\begin{NiceTabular}[t]{ll}[
code-before = \rowcolor[HTML]{FFFFFF}{1,3,5,7}
              \rowcolor[HTML]{FAFAFF}{2,4,6}
]
\toprule
\textbf{Benchmark GLM} & \textbf{Surrogate GLM} \\
\noalign{\hrule height 0.3pt}
Gender & Gender \\
 & VehAge \\
 & VehBody \\
 & DrivAge \\
 & VehValue \\
 \noalign{\hrule height 0.3pt}
 $\mathbf{1.5562}$ & $\mathbf{1.5433}$ \\
\bottomrule
\end{NiceTabular}
\caption{Australian data set}
\label{tab_sev_surr_GLM_AUS}
\end{subtable} 
\hfill
\begin{subtable}[t]{.31\linewidth}
\begin{NiceTabular}[t]{ll}[
code-before = \rowcolor[HTML]{FFFFFF}{1,3,5,7,8}
              \rowcolor[HTML]{FAFAFF}{2,4,6}
]
\toprule
\textbf{Benchmark GLM} & \textbf{Surrogate GLM} \\
\noalign{\hrule height 0.3pt}
ageph & ageph \\
agec & agec \\
coverage & coverage \\
 & bm\\
 & power \\
 & postcode \\
 \noalign{\hrule height 0.3pt}
 $\mathbf{2.2280}$ & $\mathbf{2.2437}$ \\
\bottomrule
\end{NiceTabular}
\caption{Belgian data set}
\label{tab_sev_surr_GLM_BE}
\end{subtable} 
\hfill
\begin{subtable}[t]{.31\linewidth}
\begin{NiceTabular}[t]{ll}[
code-before = \rowcolor[HTML]{FFFFFF}{1,3,5,7,9,11}
              \rowcolor[HTML]{FAFAFF}{2,4,6,8,10}
]
\toprule
\textbf{Benchmark GLM} & \textbf{Surrogate GLM} \\
\noalign{\hrule height 0.3pt}
VehPower & VehPower \\
DrivAge & DrivAge \\
BonusMalus & BonusMalus \\
VehBrand & VehBrand \\
Area & Area \\
Density & Density \\
Region & Region \\
 & VehAge \\
 & VehGas \\
 \noalign{\hrule height 0.3pt}
 $\mathbf{1.7083}$ & $\mathbf{1.7796}$ \\
\bottomrule
\end{NiceTabular}
\caption{French data set}
\label{tab_sev_surr_GLM_FR}
\end{subtable} 
\end{adjustwidth}
\caption{Comparison between the benchmark GLM and the surrogate GLM for claim severity modeling on the Australian, Belgian, and French data sets. The surrogate GLM is constructed from the CANN with GBM input and flexible output layer. The last row shows the gamma deviance of both GLMs on the out-of-sample data set $\mathcal{D}_1$.}
\label{tab_sev_surr_all_GLM}
\end{table}

\setlength{\extrarowheight}{3pt} % a bit of extra white space between lines in tables
\begin{table}[ht!]
 \begin{adjustwidth}{-0.3cm}{-0.3cm}
  \centering
  \scriptsize
  \begin{subtable}[t]{.3\linewidth}
    \begin{NiceTabular}{cr*{2}{r}}[
    code-before = \rowcolor[HTML]{FFFFFF}{1,2,4}
              \rowcolor[HTML]{FAFAFF}{3}
    ]
    \toprule
    \multirow{14}{*}{\rotatebox[origin=c]{90}{\footnotesize \textbf{Model A}}} & \multicolumn{3}{r}{\footnotesize \textbf{Model B}} \\[1mm]
    \RowStyle{\rotate}
    & & Benchmark GLM & Surrogate GLM \\\cline{2-4}
    & Benchmark GLM &  & \cmark \\ 
    & Surrogate GLM & \cmark &  \\ 
    \bottomrule
    \end{NiceTabular}
  \caption{Australian data set}
  \label{tab_DB_sev_surr_AUS}
  \end{subtable} 
  \hfill
  \begin{subtable}[t]{.3\linewidth}
    \begin{NiceTabular}{cr*{2}{r}}[
    code-before = \rowcolor[HTML]{FFFFFF}{1,2,4}
              \rowcolor[HTML]{FAFAFF}{3}
    ]
    \toprule
    \multirow{14}{*}{\rotatebox[origin=c]{90}{\footnotesize \textbf{Model A}}} & \multicolumn{3}{r}{\footnotesize \textbf{Model B}} \\[1mm]
    \RowStyle{\rotate}
    & & Benchmark GLM & Surrogate GLM \\\cline{2-4}
    & Benchmark GLM &  & \cmark \\ 
    & Surrogate GLM & \cmark &  \\ 
    \bottomrule
    \end{NiceTabular}
  \caption{Belgian data set}
  \label{tab_DB_sev_surr_BE}
  \end{subtable} 
  \hfill
  \begin{subtable}[t]{.3\linewidth}
    \begin{NiceTabular}{cr*{2}{r}}[
    code-before = \rowcolor[HTML]{FFFFFF}{1,2,4}
              \rowcolor[HTML]{FAFAFF}{3}
    ]
    \toprule
    \multirow{14}{*}{\rotatebox[origin=c]{90}{\footnotesize \textbf{Model A}}} & \multicolumn{3}{r}{\footnotesize \textbf{Model B}} \\[1mm]
    \RowStyle{\rotate}
    & & Benchmark GLM & Surrogate GLM \\\cline{2-4}
    & Benchmark GLM &  & \cellcolor{blue!25}\xmark \\ 
    & Surrogate GLM & \cmark &  \\ 
    \bottomrule
    \end{NiceTabular}
  \caption{French data set}
  \label{tab_DB_sev_surr_FR}
  \end{subtable} 
  \end{adjustwidth}
  \caption{Results of the Diebold-Mariano test comparing the predictive accuracy between the benchmark GLM and the surrogate GLM for claim severity modeling. The table indicates whether we cannot reject (\cmark) the null hypothesis $H_0: \E\left[\mathscr{L}(f_A(\X),Y) - \mathscr{L}(f_B(\X),Y)\right] = 0$, or if we reject $H_0$ (\xmark) in favor of the alternative hypothesis $H_1: \E\left[\mathscr{L}(f_A(\X),Y) - \mathscr{L}(f_B(\X),Y)\right] > 0$. Highlighted cells indicate when the null hypothesis is rejected.}
  \label{tab_DB_sev_surr}
\end{table}

\begin{figure}[ht!]
\begin{adjustwidth}{-1.6cm}{-1cm}
\centering
  \includegraphics[width = \linewidth]{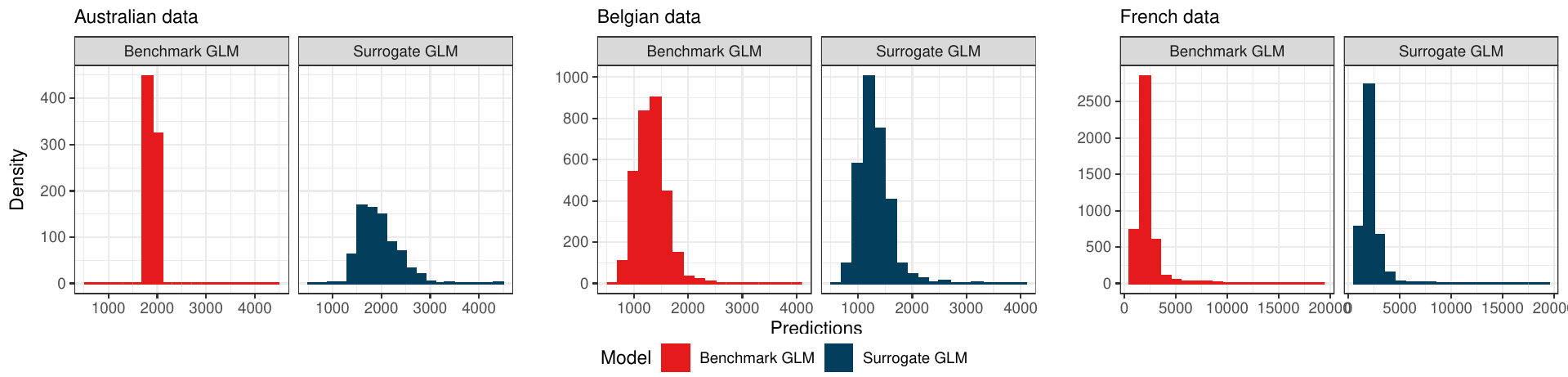}
\end{adjustwidth}
\caption{Dispersion of predictions made by the GLM compared to the surrogate model based on the CANN GBM flexible.}
\label{fig_histo_sev_surr}
\end{figure}

\begin{figure}[ht!]
\begin{adjustwidth}{-1.6cm}{-1cm}
\centering
  \includegraphics[width = \linewidth]{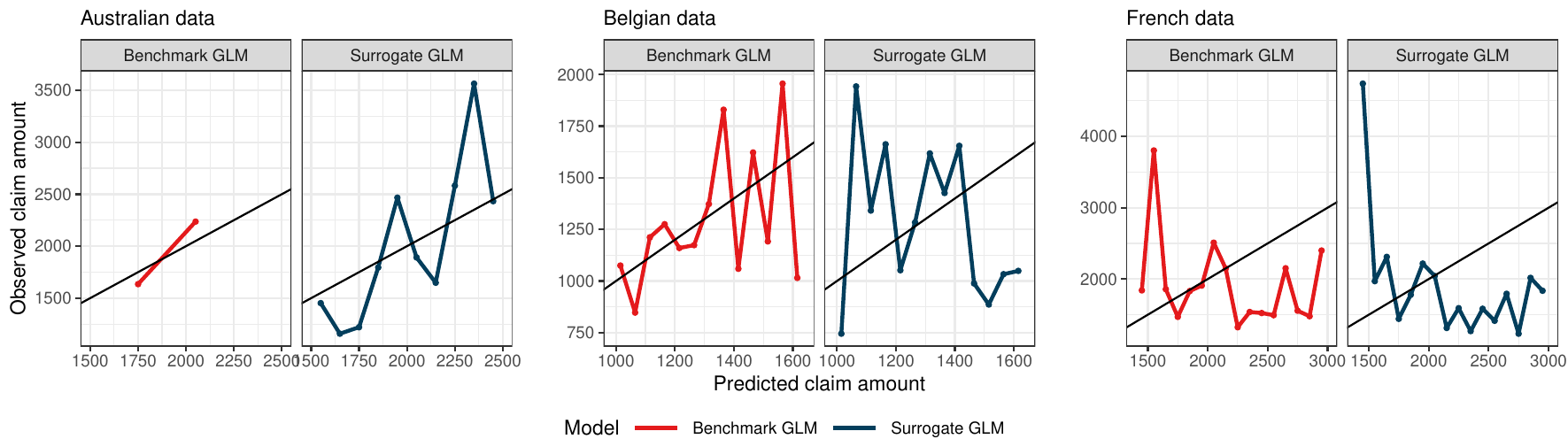}
\end{adjustwidth}
\caption{Plot of $\E \left[Y | f(\X) = s \right]$ over the range of predictions $s$ made by the benchmark GLM and the surrogate GLM for claim severity modeling. Predictions are made on test set $\mathcal{D}_1$. A line above (below) the diagonal indicates that the model is underestimating (overestimating) the true number of claims in the data.}
\label{fig_calib_sev_surr}
\end{figure}

\begin{figure}[ht!]
\centering
  \includegraphics[width = \linewidth]{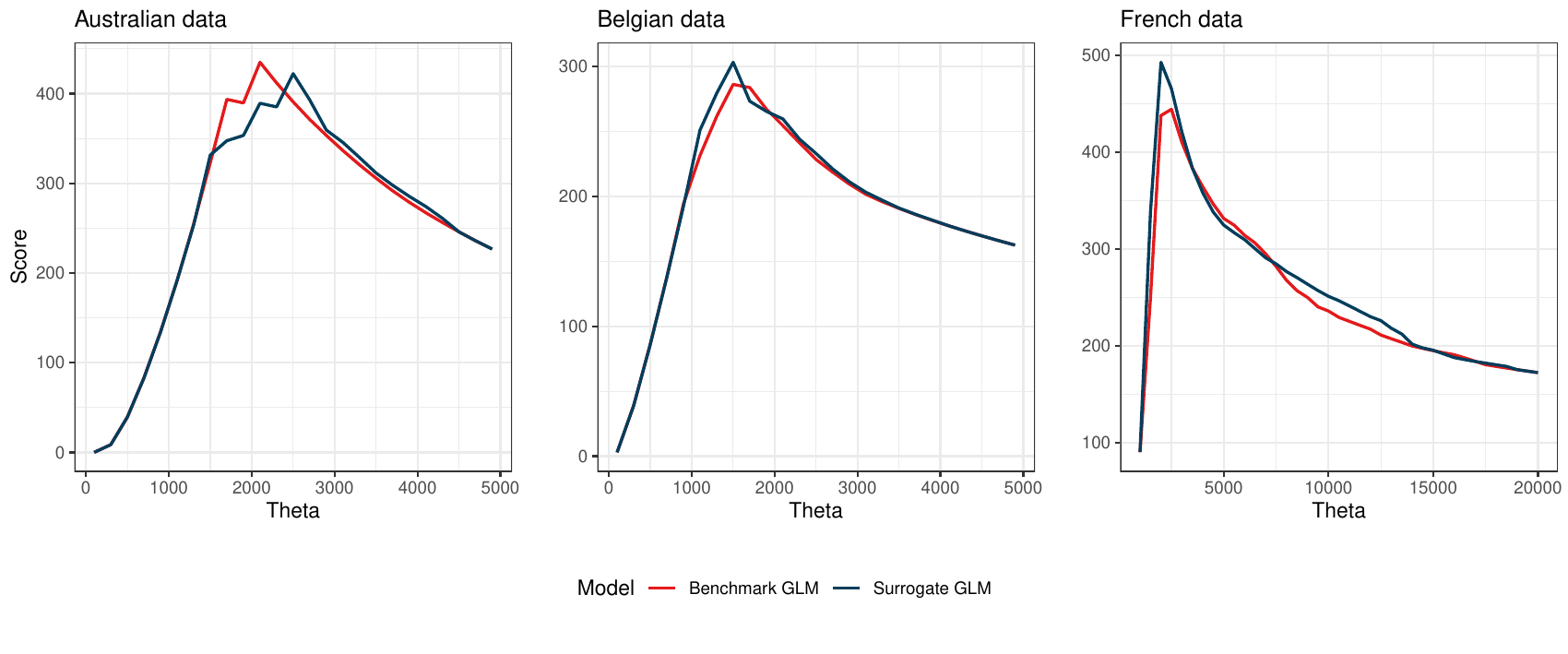}
\caption{Murphy diagram to compare predictive dominance between the benchmark GLM severity model and the surrogate GLM based on the CANN GBM flexible severity model.}
\label{fig_murphy_sev_surr}
\end{figure}

\section{Comparison between surrogate GLM and the CANN GBM flexible}\label{app_surr_cann}

\begin{figure}[ht!]
\begin{adjustwidth}{-1.6cm}{-1cm}
\centering
  \includegraphics[width = \linewidth]{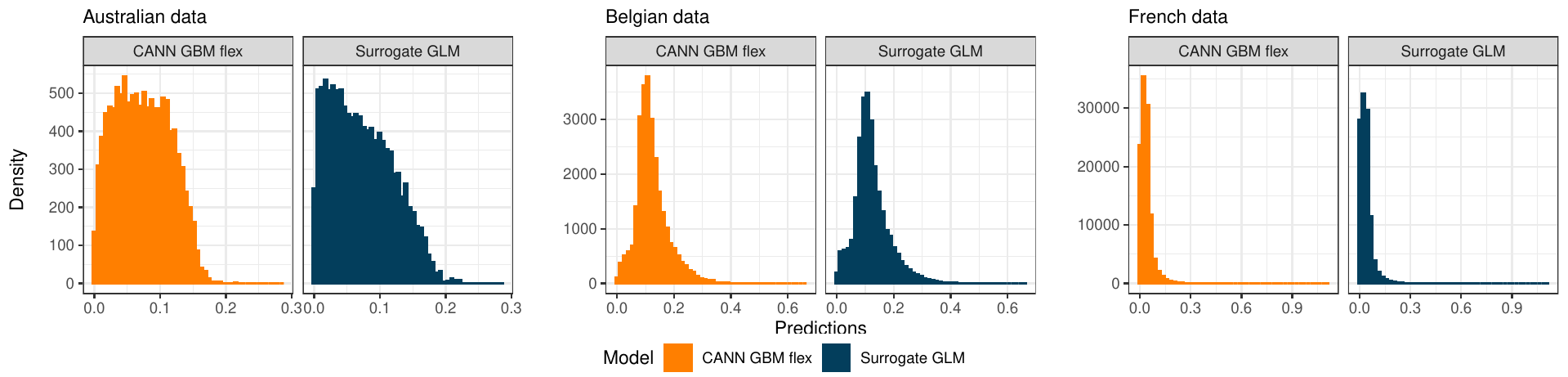}
\end{adjustwidth}
\caption{Dispersion of predictions made by the GLM compared to the surrogate model based on the CANN GBM flexible.}
\label{fig_histo_surr_CANN}
\end{figure}

\begin{figure}[ht!]
\begin{adjustwidth}{-1.6cm}{-1cm}
\centering
  \includegraphics[width = \linewidth]{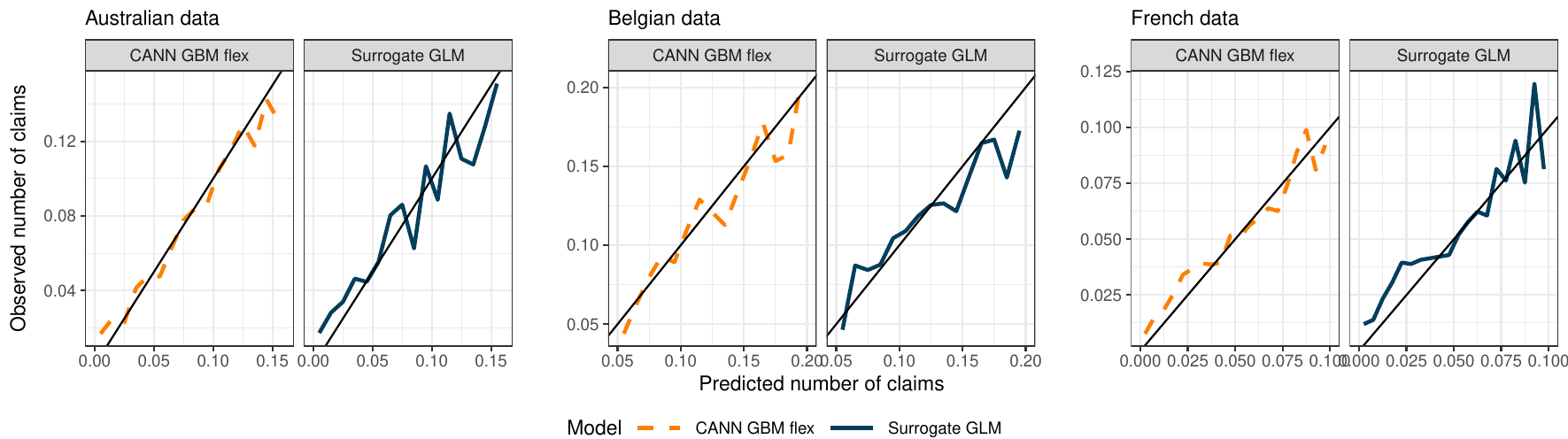}
\end{adjustwidth}
\caption{Plot of $\E \left[Y | f(\X) = s \right]$ over the range of predictions $s$ made by the CANN GBM flexible for frequency modeling and the surrogate GLM based on the CANN GBM flexible. Predictions are made on test set $\mathcal{D}_1$. A line above (below) the diagonal indicates that the model is underestimating (overestimating) the true number of claims in the data.}
\label{fig_calib_surr_CANN}
\end{figure}

\begin{figure}[ht!]
\centering
  \includegraphics[width = \linewidth]{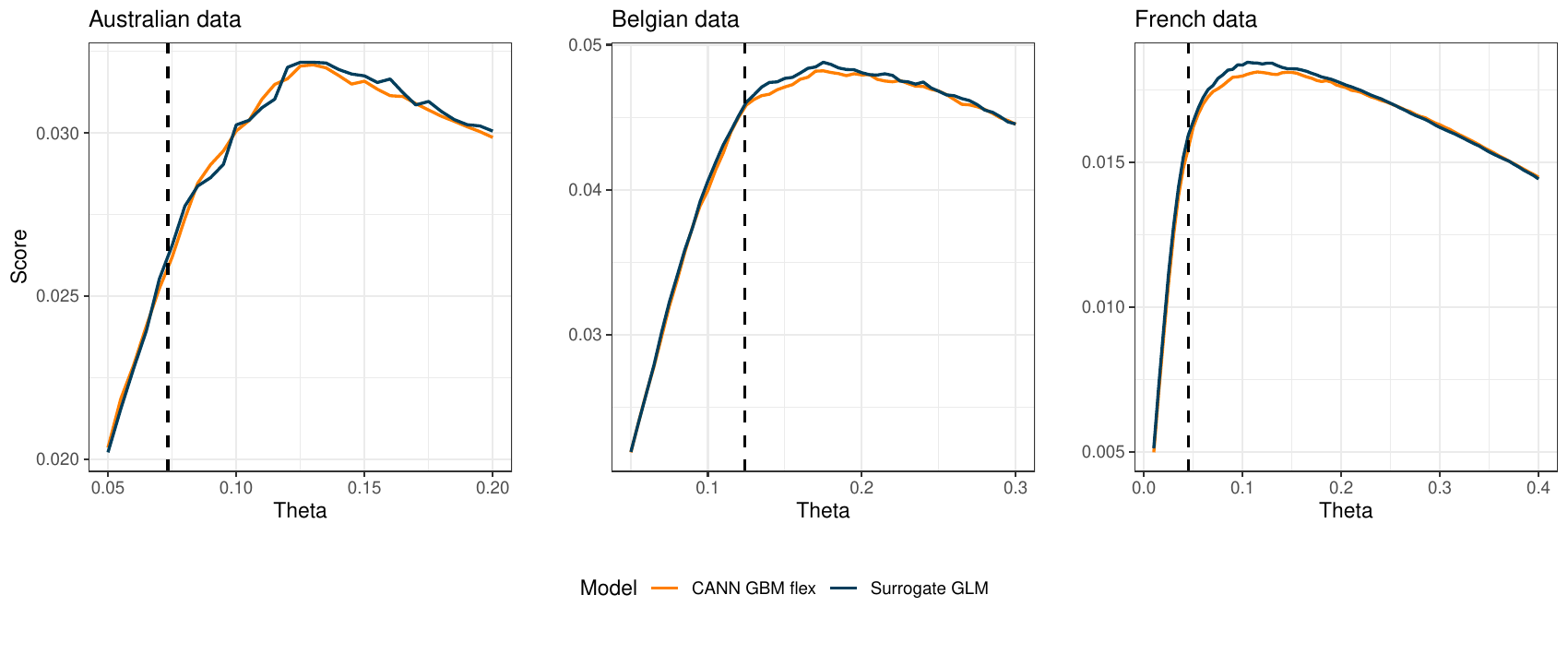}
\caption{Murphy diagram to compare predictive dominance between the CANN GBM flexible for claim frequency modeling and the surrogate GLM based on the CANN GBM flexible.}
\label{fig_murphy_surr_CANN}
\end{figure}

We compare the performance of the claim frequency surrogate GLM based on the insights from the CANN GBM flexible, as constructed in Section \ref{sec_surrogates}, with the performance of the CANN GBM flexible. Table \ref{tab_DB_surr_CANN} shows the results of the Diebold-Mariano test, as well as the out-of-sample Poisson deviance of both models on test set $\mathcal{D}_1$. Figure \ref{fig_histo_surr_CANN} shows the comparison of prediction dispersion, Figure \ref{fig_calib_surr_CANN} compares the calibration, and  Figure \ref{fig_murphy_surr_CANN} shows the Murphy diagrams for both models.

\setlength{\extrarowheight}{3pt} % a bit of extra white space between lines in tables
\begin{table}[ht!]
 \begin{adjustwidth}{-1cm}{-1cm}
  \centering
  \scriptsize
  \begin{subtable}[t]{.3\linewidth}
    \begin{NiceTabular}{cr*{2}{r}}[
    code-before = \rowcolor[HTML]{FFFFFF}{1,2,4,5}
              \rowcolor[HTML]{FAFAFF}{3}
    ]
    \toprule
    \multirow{15.4}{*}{\rotatebox[origin=c]{90}{\footnotesize \textbf{Model A}}} & \multicolumn{3}{r}{\footnotesize \textbf{Model B}} \\[1mm]
    \RowStyle{\rotate}
    & & CANN GBM flex & Surrogate GLM \\\cline{2-4}
    & CANN GBM flex &  & \cmark \\ 
    & Surrogate GLM & \cellcolor{blue!25}\xmark &  \\\cline{1-4}
    & Poisson deviance & $0.3782$ & $0.3805$ \\
    \bottomrule
    \end{NiceTabular}
  \caption{Australian data set}
  \label{tab_DB_surr_CANN_AUS}
  \end{subtable} 
  \hfill
  \begin{subtable}[t]{.3\linewidth}
    \begin{NiceTabular}{cr*{2}{r}}[
    code-before = \rowcolor[HTML]{FFFFFF}{1,2,4,5}
              \rowcolor[HTML]{FAFAFF}{3}
    ]
    \toprule
    \multirow{15.4}{*}{\rotatebox[origin=c]{90}{\footnotesize \textbf{Model A}}} & \multicolumn{3}{r}{\footnotesize \textbf{Model B}} \\[1mm]
    \RowStyle{\rotate}
    & & CANN GBM flex & Surrogate GLM \\\cline{2-4}
    & CANN GBM flex &  & \cmark \\ 
    & Surrogate GLM & \cellcolor{blue!25}\xmark &  \\\cline{1-4}
    & Poisson deviance & $0.5279$ & $0.5308$ \\
    \bottomrule
    \end{NiceTabular}
  \caption{Belgian data set}
  \label{tab_DB_surr_CANN_BE}
  \end{subtable} 
  \hfill
  \begin{subtable}[t]{.3\linewidth}
    \begin{NiceTabular}{cr*{2}{r}}[
    code-before = \rowcolor[HTML]{FFFFFF}{1,2,4,5}
              \rowcolor[HTML]{FAFAFF}{3}
    ]
    \toprule
    \multirow{15.4}{*}{\rotatebox[origin=c]{90}{\footnotesize \textbf{Model A}}} & \multicolumn{3}{r}{\footnotesize \textbf{Model B}} \\[1mm]
    \RowStyle{\rotate}
    & & CANN GBM flex & Surrogate GLM \\\cline{2-4}
    & CANN GBM flex &  & \cellcolor{blue!25}\xmark \\ 
    & Surrogate GLM & \cellcolor{blue!25}\xmark &  \\\cline{1-4}
    & Poisson deviance & $0.2695$ & $0.2738$ \\
    \bottomrule
    \end{NiceTabular}
  \caption{French data set}
  \label{tab_DB_surr_CANN_FR}
  \end{subtable} 
  \end{adjustwidth}
  \caption{Results of the Diebold-Mariano test comparing the predictive accuracy between the frequency CANN GBM flexible and the surrogate GLM based on the insights of the CANN GBM flexible. The table indicates whether we cannot reject (\cmark) the null hypothesis $H_0: \E\left[\mathscr{L}(f_A(\X),Y) - \mathscr{L}(f_B(\X),Y)\right] = 0$, or if we reject $H_0$ (\xmark) in favor of the alternative hypothesis $H_1: \E\left[\mathscr{L}(f_A(\X),Y) - \mathscr{L}(f_B(\X),Y)\right] > 0$. Highlighted cells indicate when the null hypothesis is rejected.}
  \label{tab_DB_surr_CANN}
\end{table}

\section{Construction of the prediction calibration plots}\label{app_calibplot}

In Figure \ref{fig_exp_response}, \ref{fig_calib_surr}, \ref{fig_calib_sev_surr}, and \ref{fig_calib_surr_CANN} we examine the graph of $\E \left[Y | f(\x) = s \right]$, which shows the relationship between predicted values and the corresponding average response variable. To construct this graph, we divide the range of predictions $s$ into splitpoints $s_1,\ldots,s_m$. The first interval boundary, $s_1$, is set at the 10th percentile and the last, $s_m$, at the 90th percentile. The remaining split points are determined to ensure a sufficient number of observations in each bin $[s_\gamma,s_{\gamma+1}]$, for $\gamma$ in $1,\ldots,m-1$. The Murphy diagram is constructed by looking at the average response for each bin:
\[\E \left[Y | f(\x) \in [s_\gamma,s_{\gamma+1}] \right] = \sum_{i: f(\x_i)\in[s_\gamma,s_{\gamma+1}]} \frac{y_i}{n_{\gamma,\gamma+1}},\]
where $n_{\gamma,\gamma+1}$ denotes the number of elements in the set $\{f(\x_i) | f(\x_i)\in[s_\gamma,s_{\gamma+1}]\}$. This approach provides a smoothed representation of the average response as a function of the predicted value. Table \ref{tab_calibbins} gives the used bins for each data set.

\setlength{\extrarowheight}{3pt} % a bit of extra white space between lines in tables
\begin{table}[ht!]
 \begin{adjustwidth}{-1cm}{-1cm}
  \centering
  \footnotesize
    \begin{NiceTabular}{lr}[
    code-before = \rowcolor[HTML]{FFFFFF}{1,2,4,6}
              \rowcolor[HTML]{FAFAFF}{3,5}
    ]
    \toprule
    Data set & \multicolumn{1}{r}{Frequency modeling}\\
    \noalign{\hrule height 0.3pt}
    Australian & $s = [<0.01, 0.01 - 0.02, \ldots,0.15 - 0.16,>0.16]$ \\
    Belgian & $s = [<0.06, 0.06 - 0.07, \ldots,0.19 - 0.20,>0.20]$ \\
    French & $s = [<0.005, 0.005 - 0.01, \ldots,0.095 - 0.1,>0.1]$ \\
    Norwegian & $s = [<0.02, 0.02 - 0.03, \ldots,0.09 - 0.1,>0.1]$ \\
    \noalign{\hrule height 0.3pt}
    \noalign{\medskip}
    \multicolumn{2}{r}{Severity modeling} \\
    \noalign{\hrule height 0.3pt}
    Australian & $s = [<1\,500, 1\,500 - 1\,600, \ldots,2\,400 - 2\,500,>2\,500]$ \\
    Belgian & $s = [<990, 990 - 1\,040, \ldots,1\,590 - 1\,640,>1\,640]$ \\
    French & $s = [<1\,400, 1\,400 - 1\,500, \ldots,2\,900 - 3\,000,>3\,000]$ \\
    Norwegian & $s = [<22\,000, 22\,000 - 22\,150, \ldots,24\,850 - 25\,000,>25\,000]$ \\
    \bottomrule
    \end{NiceTabular}
  \end{adjustwidth}
 \caption{Binned range of predictions used in the construction of Figure \ref{fig_exp_response}, \ref{fig_calib_surr}, \ref{fig_calib_sev_surr}, and \ref{fig_calib_surr_CANN}.}
  \label{tab_calibbins}
\end{table}

\end{document}